\documentclass[10pt,twocolumn,letterpaper]{article}

\usepackage[pagenumbers]{wacv} 

\usepackage[utf8]{inputenc} 
\usepackage{multirow}
\usepackage{graphicx}
\usepackage{subcaption}
\usepackage{makecell}
\usepackage{enumitem}
\usepackage{amsmath}
\usepackage{arydshln}
\usepackage{pifont}
\newcommand{\xmark}{\ding{55}}%

\definecolor{wacvblue}{rgb}{0.21,0.49,0.74}
\usepackage[pagebackref,breaklinks,colorlinks,allcolors=wacvblue]{hyperref}

\def\wacvPaperID{116} 
\def\confName{WACV}
\def\confYear{2026}

\title{SynPlay: Large-Scale Synthetic Human Data with Real-World Diversity for Aerial-View Perception}

\author{
\textbf{Jinsub Yim$^{1}$}\thanks{Authors contributed equally} \quad \textbf{Hyungtae Lee$^{*}$} \quad \textbf{Sungmin Eum$^{2*}$} \\ 
\textbf{Yi-Ting Shen}$^1$ \quad \textbf{Yan Zhang}$^1$ \quad \textbf{Heesung Kwon}$^2$ \quad \textbf{Shuvra S. Bhattacharyya}$^1$ \\
  $^1$University of Maryland College Park \quad $^2$DEVCOM Army Research Laboratory\\
  {\small $^*$Authors contributed equally}
}

\begin{document}

\twocolumn[{%
\renewcommand\twocolumn[1][]{#1}%
\maketitle
\begin{center}
    \centering
\setlength{\tabcolsep}{0.5pt}
\begin{tabular}{ccc}
\includegraphics[width=.32\linewidth]{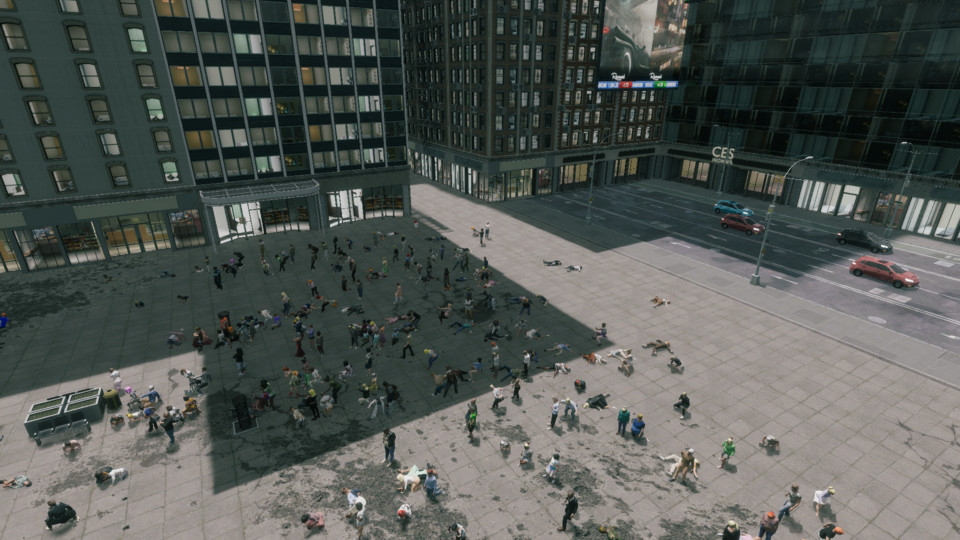} ~&~
\includegraphics[width=.32\linewidth]{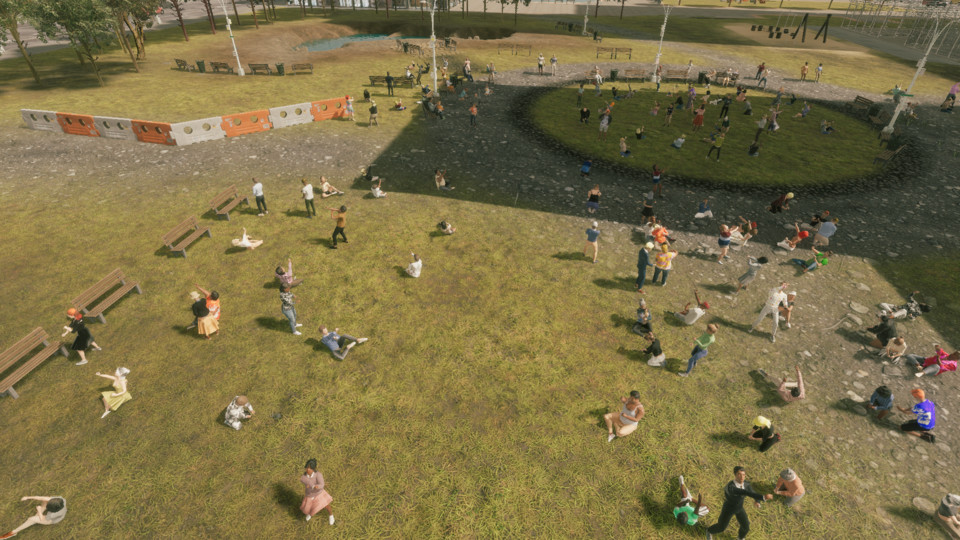} ~&~
\includegraphics[width=.32\linewidth]{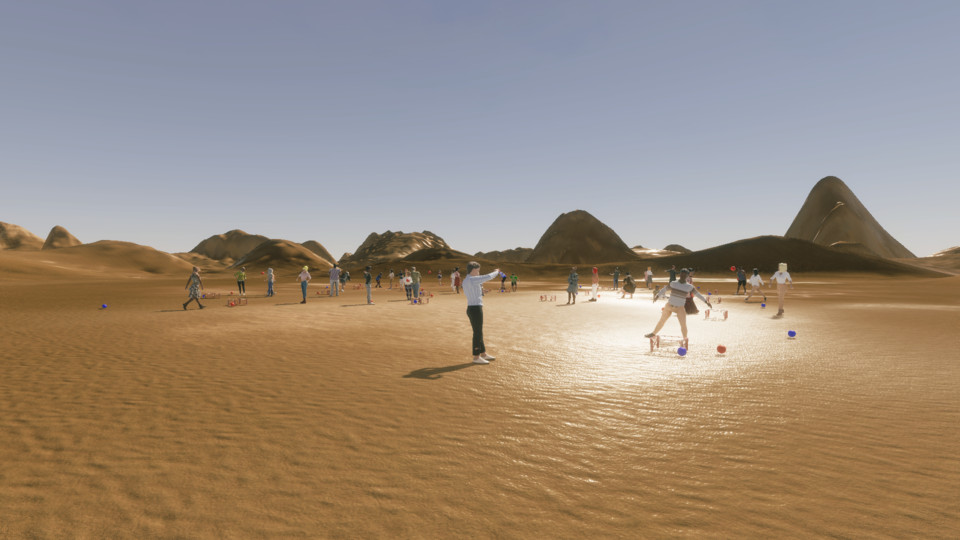} \\\vspace{-0.25cm}\\
\includegraphics[width=.32\linewidth]{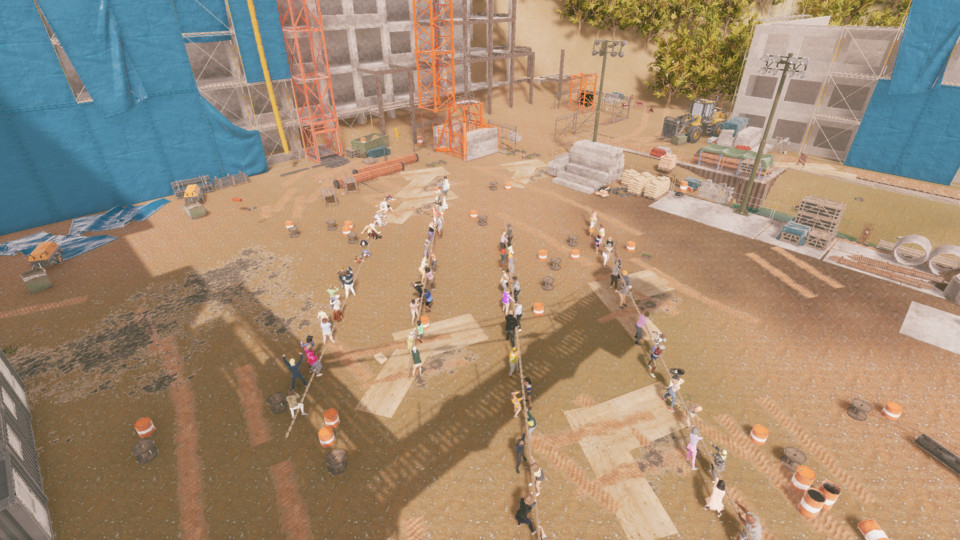} ~&~
\includegraphics[width=.32\linewidth]{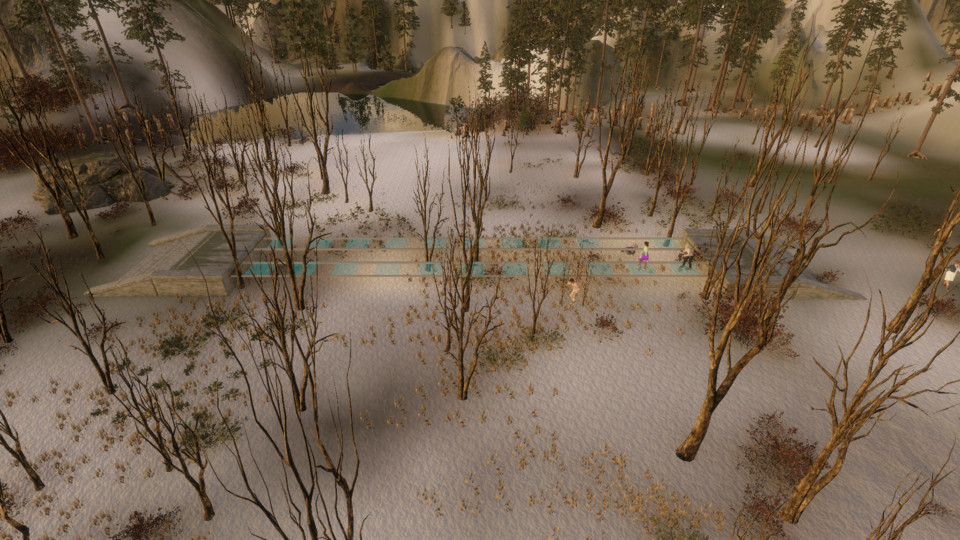} ~&~
\includegraphics[width=.32\linewidth]{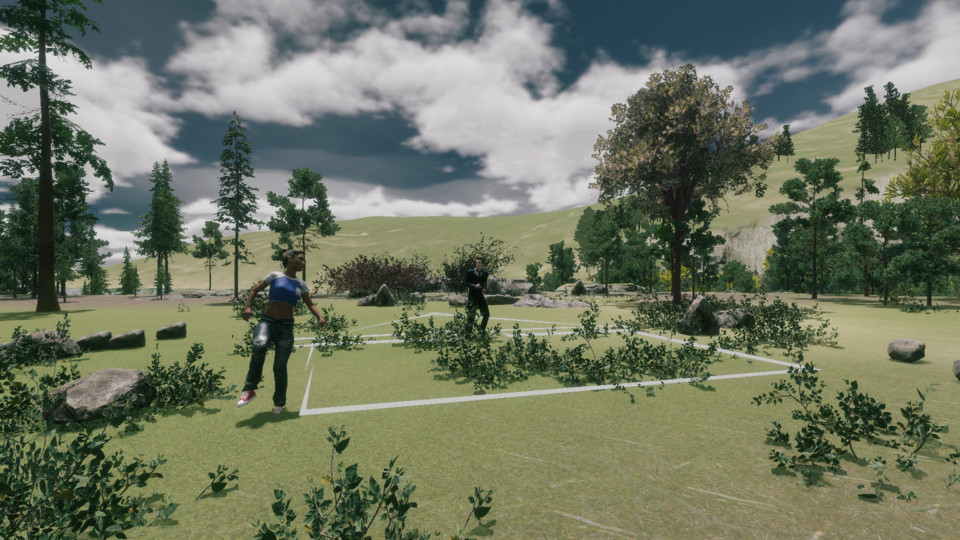} \\
\end{tabular}
\vspace{-0.3cm}
\captionof{figure}{{\bf SynPlay} captures players performing six traditional games in a virtual playground, also featured in the Netflix series ``Squid Game''~\citep{SquidGameNetflix2021}. It promotes motion diversity via \textbf{rule-guided motion generation}, where actions evolve dynamically through game mechanics. A \textbf{multi-perspective setup} enables near-to-far viewpoint coverage while capturing diverse behaviors and appearances across angles.
}
\label{fig:synplay}
\end{center}%
}]

\maketitle

\begin{abstract}

We introduce \textbf{SynPlay}, a large-scale synthetic human dataset purpose-built for advancing multi-perspective human localization, with a predominant focus on aerial-view perception. SynPlay departs from traditional synthetic datasets by addressing a critical but underexplored challenge: localizing humans in aerial scenes where subjects often occupy only \textbf{tens of pixels} in the image. In such scenarios, fine-grained details like facial features or textures become irrelevant, shifting the burden of recognition to human motion, behavior, and interactions. To meet this need, SynPlay implements a novel \textbf{rule-guided motion generation framework} that combines real-world motion capture with \textbf{motion evolution graphs}. This design enables human actions to evolve dynamically through high-level game rules rather than predefined scripts, resulting in \textbf{effectively uncountable motion variations}. Unlike existing synthetic datasets—which either focus on static visual traits or reuse a limited set of mocap-driven actions—SynPlay captures a wide spectrum of spontaneous behaviors, including complex interactions that naturally emerge from unscripted gameplay scenarios. SynPlay also introduces an extensive multi-camera setup that spans UAVs at random altitudes, CCTVs, and a freely roaming UGV, achieving \textbf{true near-to-far perspective coverage} in a single dataset. The majority of instances are captured from aerial viewpoints at varying scales, directly supporting the development of models for long-range human analysis—a setting where existing datasets fall short. Our data contains over 73k images and 6.5M human instances, with detailed annotations for detection, segmentation, and keypoint tasks. Extensive experiments demonstrate that training with SynPlay significantly improves human localization performance, especially in few-shot and data-scarce scenarios. 

\end{abstract}    
\section{Introduction}

Large-scale synthetic datasets have become a cornerstone in modern computer vision, providing scalable solutions for training large-capacity models \cite{Unity,Unreal}. Advances in rendering engines such as Unity \cite{Unity}, Unreal Engine \cite{Unreal}, and human modeling tools like MakeHuman \cite{MakeHuman} and Character Creator \cite{Charactercreator} have enabled increasingly realistic representations of virtual humans in diverse environments. These tools have expanded the use of synthetic data across applications such as human pose estimation \cite{PPatelCVPR2021,JLiaoCVPR2024}, shape reconstruction \cite{NMahmoodICCV2019}, and semantic segmentation \cite{SRichterECCV2016}.

However, a critical but underexplored challenge remains: identifying humans from aerial perspectives where subjects occupy only \textit{tens of pixels}. In these cases, detailed appearance cues such as facial features, textures, or fine-grained clothing differences become irrelevant. Instead, recognition must rely on factors like human motion, posture, and behavioral interactions. Despite the growing need for robust long-range human localization—relevant to tasks such as UAV-based surveillance, disaster response, and crowd monitoring—\textbf{no existing dataset jointly addresses this challenge by providing both diverse human actions and extensive aerial viewpoint coverage}.

Prior synthetic human datasets have primarily focused on enhancing diversity through individual traits, such as race, gender, and clothing \cite{PPatelCVPR2021,MBlackCVPR2023}, which are effective for close-range tasks like body/pose estimation or 3D reconstruction. However, their impact diminishes in distant-view scenarios where subjects are too small for these attributes to be resolved \cite{NMahmoodICCV2019}. Moreover, datasets targeting pose diversity, such as SURREAL \cite{GVarolCVPR2017} and AMASS \cite{NMahmoodICCV2019}, rely heavily on motion capture data obtained through physically constrained environments and predefined instructions (e.g., “5 seconds waving above the head with both arms” in HDM05 \cite{NMahmoodICCV2019}). These approaches fail to capture the natural spontaneity of human behaviors, especially for complex multi-human interactions.

Existing aerial-view datasets, such as Okutama-Action \cite{MBarekatainCVPRW2017}, UAVid \cite{YLyuPRS2020}, VisDrone \cite{PZhuTPAMI2022}, and Archangel \cite{YShenAccess2023}, provide broader viewpoint coverage but suffer from limited motion diversity. Most prioritize capturing subjects from various altitudes and angles but recycle a small set of static or scripted motions. As a result, they do not reflect the breadth of real-world activities or emergent behaviors that occur in dynamic, unscripted scenarios. Synthetic datasets like SynDrone \cite{GRizzoliICCVW2023} and CARGO \cite{TZhangCVPR2021} similarly lack behavioral variability, typically limiting human actions to basic locomotion or pose transitions.

In this paper, we introduce \textbf{SynPlay}, a large-scale synthetic human dataset designed to close this gap. SynPlay departs from traditional synthetic data generation pipelines by introducing a rule-guided motion generation framework, where human motions evolve dynamically based on high-level game rules rather than predefined scripts. This framework combines real-world motion capture with motion evolution graphs, enabling a continuous expansion of behavior space. As a result, SynPlay provides \textit{effectively uncountable motion variations}, capturing spontaneous human actions, including collisions, tugging, group interactions, and diverse game dynamics that naturally emerge in multi-human scenarios. For example, while datasets like Archangel-synth \cite{YShenArXiv2024} or SynDrone \cite{GRizzoliICCVW2023} contain only 2–3 distinct motions, SynPlay features a behaviorally diverse action space that cannot be easily enumerated (see Table 1 in the Supplementary material).

Furthermore, SynPlay incorporates a \textit{multi-perspective camera setup} comprising three UAVs at randomized altitudes, three ground-based CCTVs at fixed viewpoints, and a freely roaming Unmanned Ground Vehicle (UGV). This design achieves true near-to-far perspective coverage within a single dataset, enabling robust localization of humans across extreme viewpoint variations. While some ground-level and close-up views are included, the majority of data consists of aerial observations where human subjects appear as small instances—often occupying only tens of pixels. This makes SynPlay uniquely suited for developing models that generalize across long-range aerial-view tasks, where prior datasets fall short.

To create SynPlay, we designed six traditional Korean games, also featured in the Netflix series ``Squid Game'' \cite{SquidGameNetflix2021}, as interactive scenarios. These games naturally elicit a broad range of dynamic human actions, including standing, sitting, grabbing, pulling, and wrestling, making them ideal for generating diverse behavior patterns in a controlled yet emergent manner. Each scenario is instantiated through a combination of motion blending, animation layering, and real-world motion capture, ensuring both realism and variability beyond the constraints of traditional mocap datasets.

\noindent\textbf{Our main contributions are as follows:}
\begin{itemize}
    \item We introduce \textbf{SynPlay}, the first large-scale synthetic human dataset specifically designed for identifying humans from aerial perspectives where subjects occupy only tens of pixels.
    \item We propose a \textbf{rule-guided motion generation framework}, combining motion capture data with motion evolution graphs to create \textbf{effectively uncountable motion variations}, including spontaneous multi-human interactions.
    \item We provide a \textbf{multi-perspective data collection setup} spanning UAVs, CCTVs, and UGVs, achieving true near-to-far viewpoint coverage within a single dataset.
    \item We demonstrate through extensive experiments that SynPlay significantly improves aerial-view human detection and segmentation, especially in few-shot, cross-domain, and data-scarce scenarios.
\end{itemize}

\subsection{Points to Note}

\noindent{\bf SynPlay is synthetic data—but not the usual kind.} Synthetic datasets offer essential benefits for vision tasks requiring large-scale data: effortless generation, cost-free labeling, and controllability. However, they often suffer from limited realism and motion diversity, especially in aerial-view human localization, where subjects occupy only a few dozen pixels in the image. Our core novelty addresses this by expanding the \textit{motion and behavior space}, not just varying static traits. Prior aerial-view synthetic datasets typically focus on capturing multiple viewpoints while reusing a small set of motions, leading to repetitive behaviors seen from different angles. SynPlay resolves this by introducing \textit{rule-guided motion generation}, where human actions evolve dynamically based on high-level game rules rather than predefined scripts. This framework combines real-world motion capture data with a \textit{motion evolution graph}, enabling \textbf{effectively uncountable motion variations}. For example, while datasets like Archangel-synth or SynDrone contain only 2–3 distinct motions, SynPlay captures a wide spectrum of behaviors that naturally emerge from multi-human interactions. This includes spontaneous, unscripted actions that defy simple categorization. As detailed in the supplementary material (Tab.~1), this level of motion diversity is currently unmatched in aerial-view datasets.

\noindent{\bf SynPlay is primarily designed for multi-perspective human localization, with a predominant focus on aerial perception.} SynPlay facilitates robust human localization across a broad range of perspectives and distances, with particular emphasis on aerial viewpoints. While it includes some ground-level and close-up views, the majority of the data consists of aerial observations where subjects often appear as only \textit{tens of pixels in size} (see Fig.~2 in the supplementary material). This presents unique challenges distinct from tasks like pose recovery and mesh reconstruction, which rely on close-range, high-resolution imagery.

SynPlay is the \textbf{only existing dataset to simultaneously achieve}:

\begin{itemize}
    \item High \textbf{instance-per-image density} in aerial-view scenes (88.4 instances per image on average),
    \item True \textbf{near-to-far viewpoint coverage}, including nadir-view observations,
    \item \textbf{Rule-guided human motion diversity}, addressing the lack of behavioral variation in prior aerial-view datasets.
\end{itemize}

These characteristics make SynPlay a unique and valuable resource for advancing aerial-view human analysis, few-shot learning, and data-scarce cross-domain human localization tasks.

\section{Related Works}
\label{sec:related_works}

\noindent{\bf Synthetic human data.} The creation of various synthetic human datasets has been facilitated by the advancement of modern synthetic data rendering engines such as Blender, Unity, and Unreal, alongside human modeling tools like MakeHuman~\citep{MakeHuman} and Character Creator~\citep{Charactercreator}. These rendering engines enable a realistic representation of humans in 3D virtual environments, while the modeling tools give creators precise control over the design of virtual characters. The creators of these datasets leveraged these tools to meticulously control key design factors, ensuring suitability for specific tasks, \textit{e.g.}, SOMAset~\citep{IBarbosaCVIU2018}, PersonX~\citep{XSunCVPR2019}, UnrealPerson~\citep{TZhangCVPR2021}, CARGO~\citep{QZhangCVPR2024} for Re-Identification, SURREAL~\citep{GVarolCVPR2017} for pose estimation, GTA5~\citep{SRichterECCV2016} for semantic segmentation, and Archangel-Synthetic~\citep{YShenAccess2023} for detection.

Recently, various efforts have aimed to enhance the realism of virtual human models, bringing them closer to real-world quality. SMPL-X~\citep{GPavlakosCVPR2019}, AMASS~\citep{NMahmoodICCV2019}, and HardMo~\citep{JLiaoCVPR2024} leveraged motion capture devices to record natural human motions. BEDLAM~\citep{MBlackCVPR2023} expanded diversity in factors like skin tones and clothing, though still based on SMPL-X. However, motion capture systems impose environmental constraints, especially when capturing large-scale motions or multi-human interactions. AGORA~\citep{PPatelCVPR2021} and ScoreHMR~\citep{AStathopoulosCVPR2024} attempted to bypass motion scanners by fitting body models to real-world images and videos, but model quality degrades significantly for distant-view images. A key goal of our dataset was to include multiple viewpoints, including distant perspectives, without sacrificing motion fidelity. To this end, we used motion capture devices but designed our pipeline to ensure the dataset is not limited by the recording environments.\smallskip


\noindent{\bf Natural human motion acquisition.} Whether we are creating a real or synthetic dataset, images of motions captured by directing humans to perform specific actions based on a description often appear awkward rather than natural. Because of that, most datasets aim to include humans engaged in daily activities (\textit{e.g.}, MS COCO~\citep{TLinECCV2014}, MPII Human Pose~\citep{MAndrilukaCVPR2014}), performing tasks such as sports (\textit{e.g.}, UCF-Sports~\citep{MRodriguezCVPR2008}, SoccerNet~\citep{ACioppaCVPRW2022}, SportsMOT~\citep{YCuiICCV2023}, PKU-DyMVHumans~\cite{XZhengCVPR2024}, SportsHHI~\cite{TWuCVPR2024}) or art (\textit{e.g.}, Human-Art~\citep{XJuCVPR2023}) to capture their motions and poses in the most natural states possible. However, it is self-contradictory to artificially create a virtual event to capture natural motions associated with the event. In this paper, we aim to detour this self-contradict by initially designing the virtual events (\textit{i.e.}, aforementioned games) using existing but non-natural virtual motions, which are then replaced by real-world motions captured using a motion capture device.\smallskip

\noindent{\bf Supplemental datasets for training.} Enhancing model performance by supplementing the training with additional data has been a common strategy~\citep{SRenTPAMI2016,HLeeTIP2019}. Initially, this involved combining datasets constructed with the same purpose, like MS COCO~\citep{TLinECCV2014} and PASCAL VOC~\citep{MEveringhamIJCV2015}, for tasks such as object detection. Some approaches utilized large-scale datasets (\textit{e.g.}, ImageNet~\citep{JDengCVPR2009} or Instagram~\citep{DMahajanECCV2018}), which were not necessarily designed for the target task, to build foundational features, followed by a transfer learning such as pretrain-finetune~\citep{RGirshickTPAMI2016} or PTL~\citep{YShenCVPR2023} to adjust the model on the target dataset. As models have grown in size and complexity (\textit{e.g.}, ViT~\citep{ADosovitskiyICLR2021}), the demand for large-scale, high-quality datasets has increased, but the high costs of annotation present a significant barrier. To address this, label-agnostic training methods like self-supervised learning~\citep{TChenICML2020,MCaronNeurIPS2020,JGrillNeurIPS2020,KHeCVPR2022} and synthetic dataset generation with cost-free annotations have emerged as viable solutions. In response, we have specifically designed SynPlay  to supplement various computer vision tasks that require a large-scale, highly-diversified human appearance set.
\section{SynPlay Dataset}
\label{sec:dataset}

\begin{figure*}[t]
\centering
\includegraphics[trim=5mm 5mm 5mm 5mm,clip,width=\linewidth]{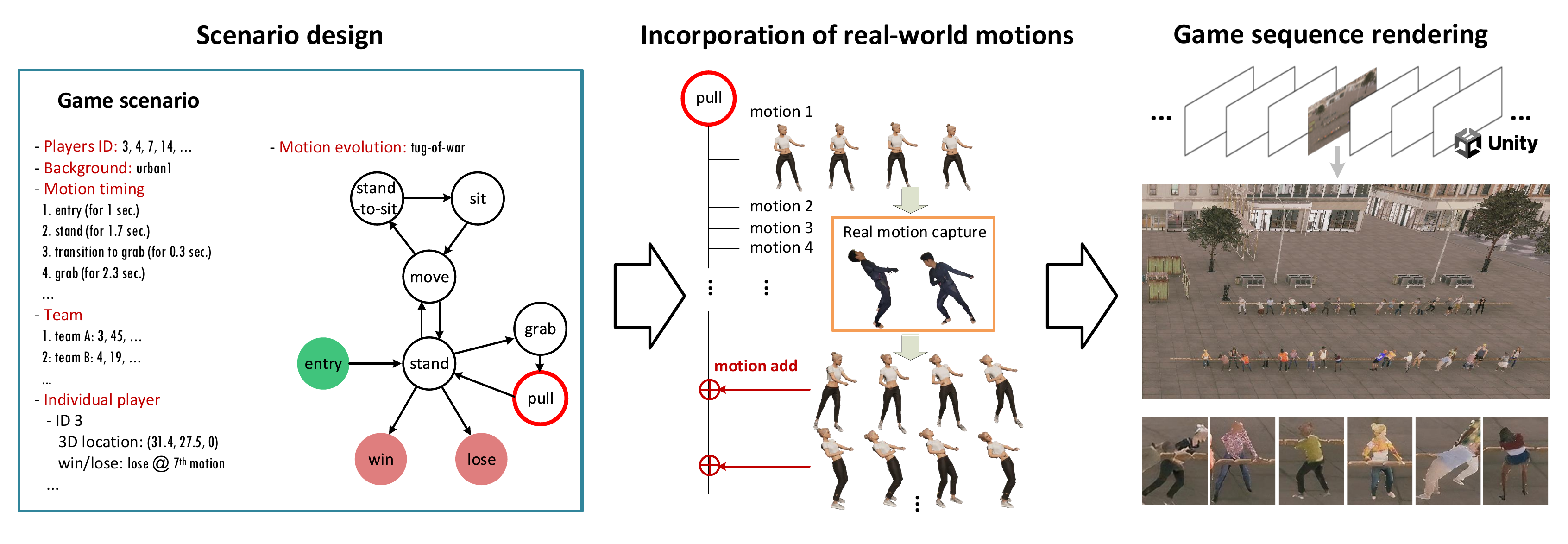}
\vspace{-0.6cm}
\caption{{\bf Game sequence generation pipeline.} This illustrate how we create a sequence for a \texttt{tug-of-war} game which includes an example of how we incorporate real-world motions towards the elementary motion state of \texttt{pull}. In the motion evolution graph, the start and end nodes are indicated by green and red circles, respectively. A diverse set of \texttt{pull} motion instances is shown below the image of the rendered scene.}
\label{fig:two-stage_motion_design}
\end{figure*}

In a nutshell, we aim to create a synthetic dataset in a virtual environment that captures the diversity of human appearances found in the real world. 

Accordingly, we design our dataset with a focus on two key aspects that naturally diversifies the human appearances captured in the scenes: i) expanding the motion set with increased reality, and ii) capturing each scene from a diverse set of camera viewpoints.
\smallskip

\noindent{\bf Diverse yet realistic human motions.} We use \emph{rule-guided motion design} in our SynPlay dataset, borrowing ``rules'' from six traditional Korean games, also featured in the Netflix series ``Squid Game''~\citep{SquidGameNetflix2021}\footnote{Our game scenarios are designed based on the traditional game rules, without taking any specific situations from the show.}. This approach offers coarser motion guidance for the virtual players, facilitating the generation of a wide spectrum of natural motions, even including the ones that defy detailed description.

Our rule-guided motion design is effectively baked into the overall sequence generation pipeline, as shown in Figure~\ref{fig:two-stage_motion_design}, that consists of the \textit{scenario design} followed by the \textit{incorporation of real-world motions}. The scenario design involves the setup of all the knobs that control the appearance, players (winners and losers), game dynamics (\textit{e.g.}, how/when each game ends), and the human motion evolutions for each specific scenario. In each scenario, all items that don’t need to be hard-coded (such as game rules) are selected randomly during scenario design. The motion evolution of each virtual player in a specific game is governed by a graph structure where all possible elementary motions and their potential transitions are represented as nodes and directed edges, respectively. Each node is tied with a pool of motions that fall under the same elementary motion state (\textit{e.g.}, move, sit). As the game progresses, a virtual player evolves its motion following the directed edges and stays at a node according to the `motion timing', also defined in the scenario design. In each state, the virtual human randomly chooses to exhibit one of the motions in the corresponding pool. It is noteworthy to mention that, while our motion evolution graphs are rule-guided, they are compositional and parameterized. This yields a combinatorial space, effectively generating unbounded variants rather than a fixed library.


Before incorporating the real-world motions, we leverage two techniques to pre-diversify the elementary human motions readily available in human motion libraries such as Mixamo~\cite{Mixamo}: i) dynamically blending two existing motions of similar types to generate a new motion type (\textit{e.g.}, blending slow-walking and running to generate hasty-walking), ii) using elementary motions as animation layers to make a new motion (\textit{e.g.}, raising hands while walking). On top of the pre-diversified set of motions for each game, a real human player wearing a motion capture device~\footnote{Each real player used a SmartSuit Pro II and a pair of SmartGloves from Rokoko (\url{http://rokoko.com})} is asked to either similarly mimic or newly create motions that align with the given game rule (thus, \textit{rule-guided}). For example, for the game of tug-of-war, human players were provided with the game \textit{rule}, and then asked to reenact any possible motion with the freedom of choosing the winning or losing side. For some games, more than one players were asked to actually play the game together to capture the motions that can naturally arise at the time of physical interactions. As the result of \textit{incorporating the real-world motions}, the total number of unique motions in SynPlay increased from 104 to 257.\smallskip

\begin{figure*}[t]
\centering

\setlength{\tabcolsep}{0.5pt}
\begin{tabular}{ccccccc}
UAV low-alt & UAV med-alt & UAV high-alt & CCTV front & CCTV side & CCTV back & UGV \\
\includegraphics[width=.141\linewidth]{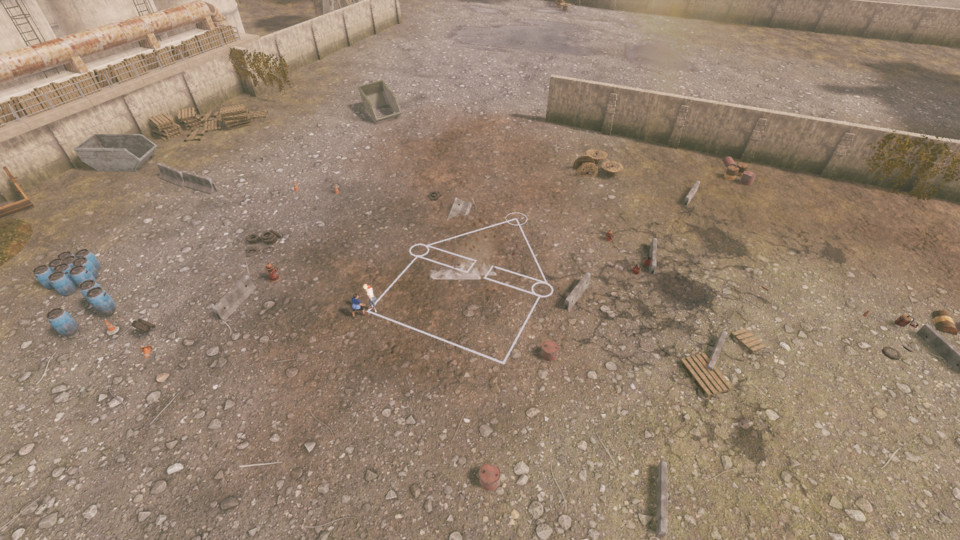} &
\includegraphics[width=.141\linewidth]{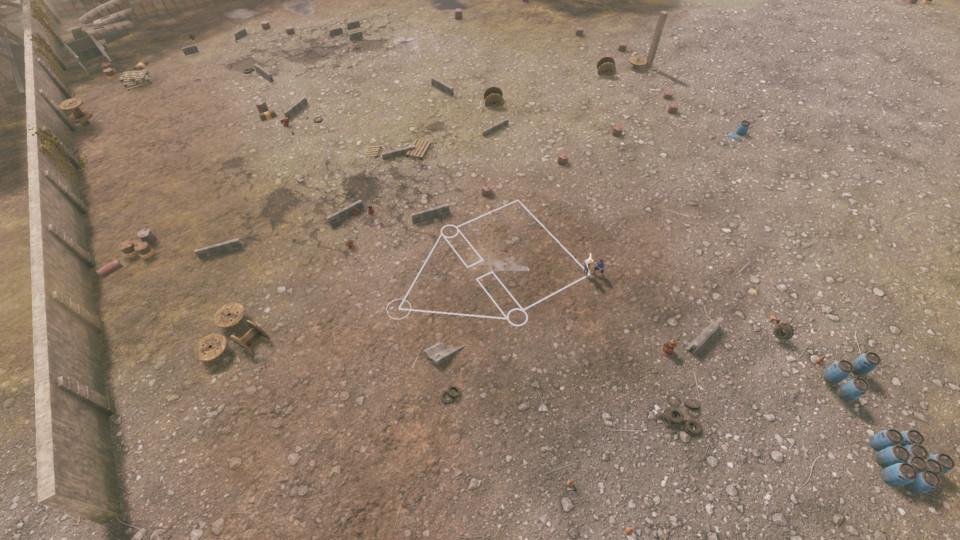} &
\includegraphics[width=.141\linewidth]{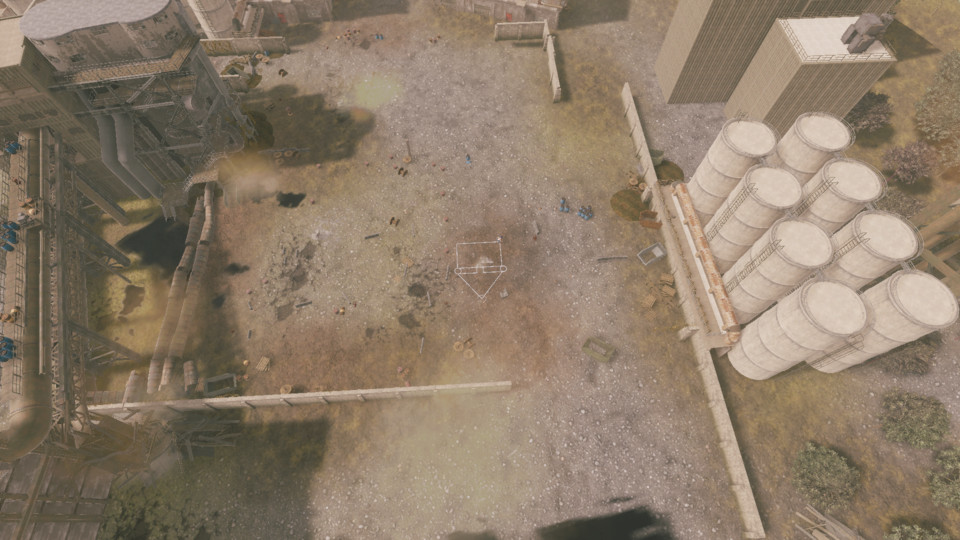} &
\includegraphics[width=.141\linewidth]{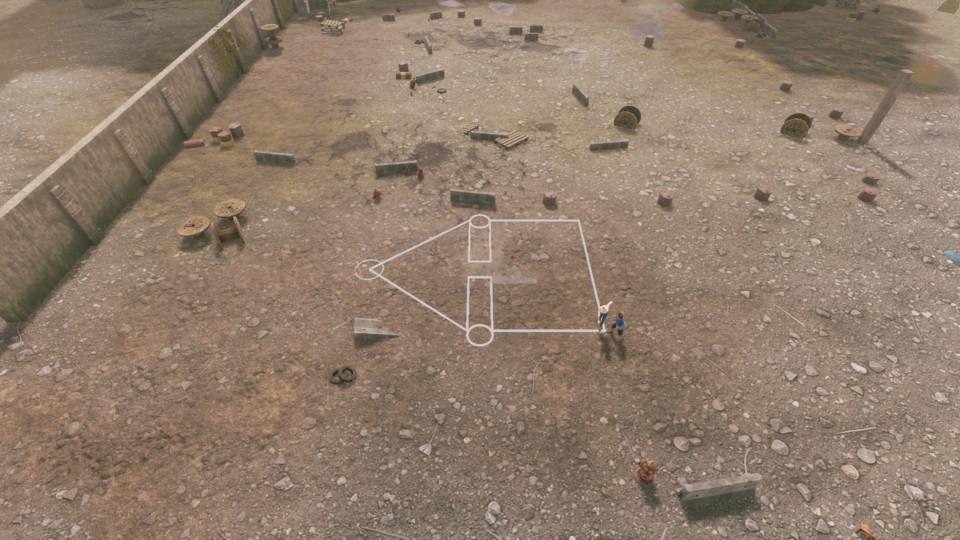} &
\includegraphics[width=.141\linewidth]{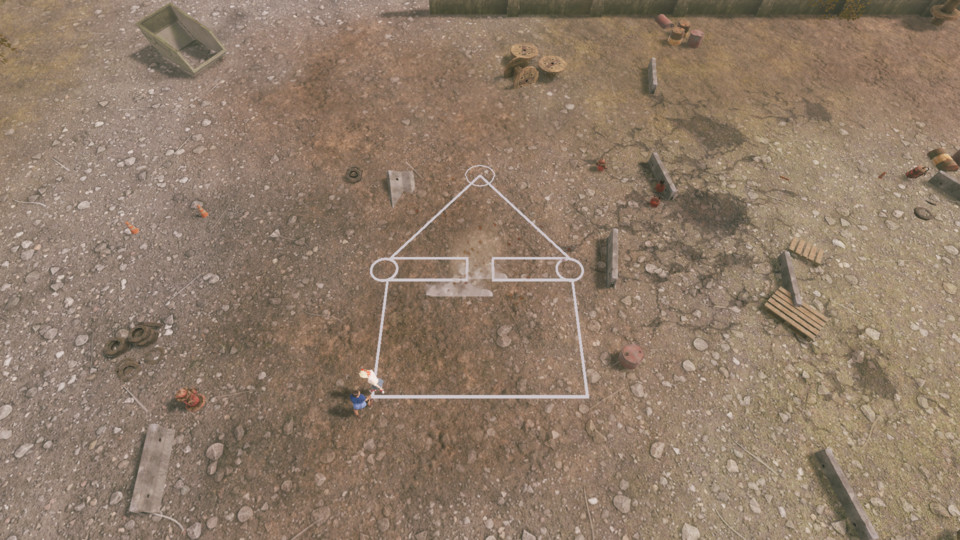} &
\includegraphics[width=.141\linewidth]{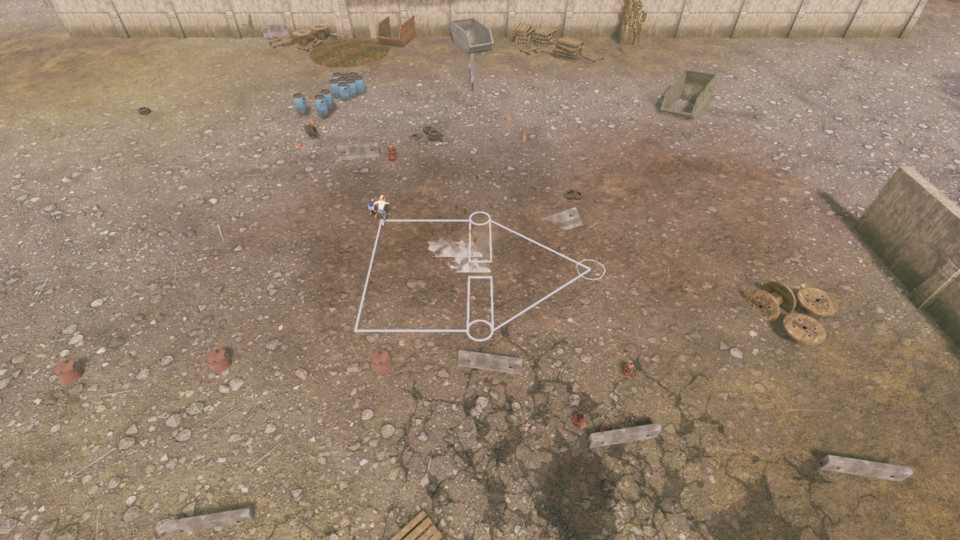} &
\includegraphics[width=.141\linewidth]{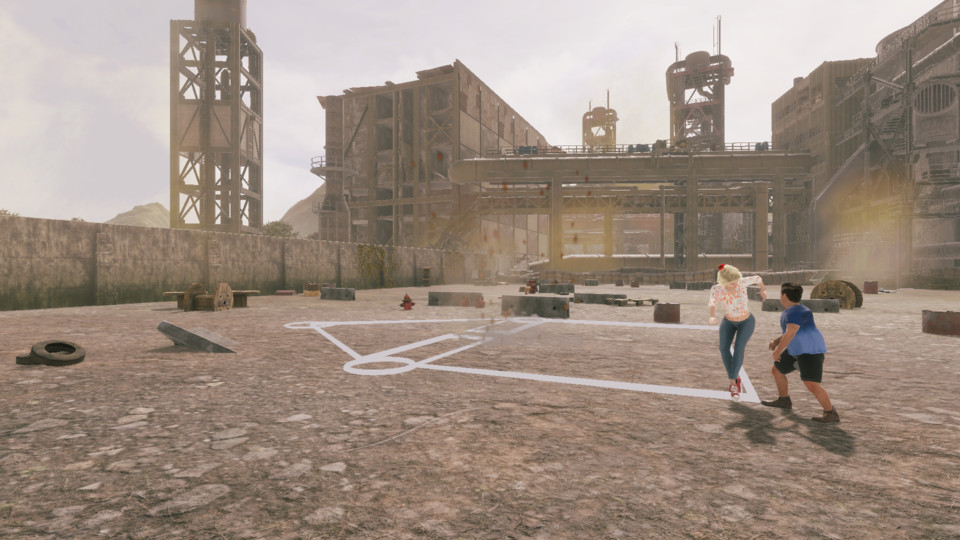} \\
\includegraphics[width=.141\linewidth]{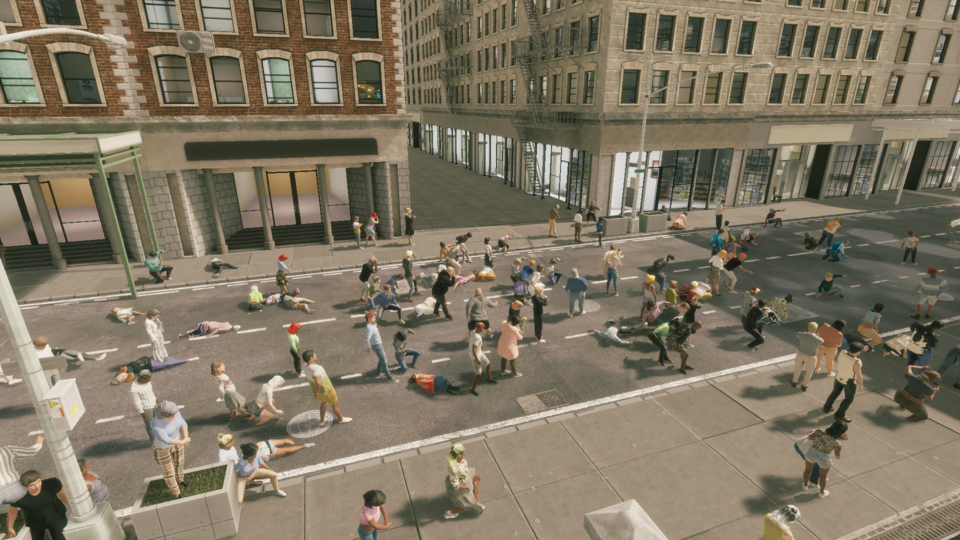} &
\includegraphics[width=.141\linewidth]{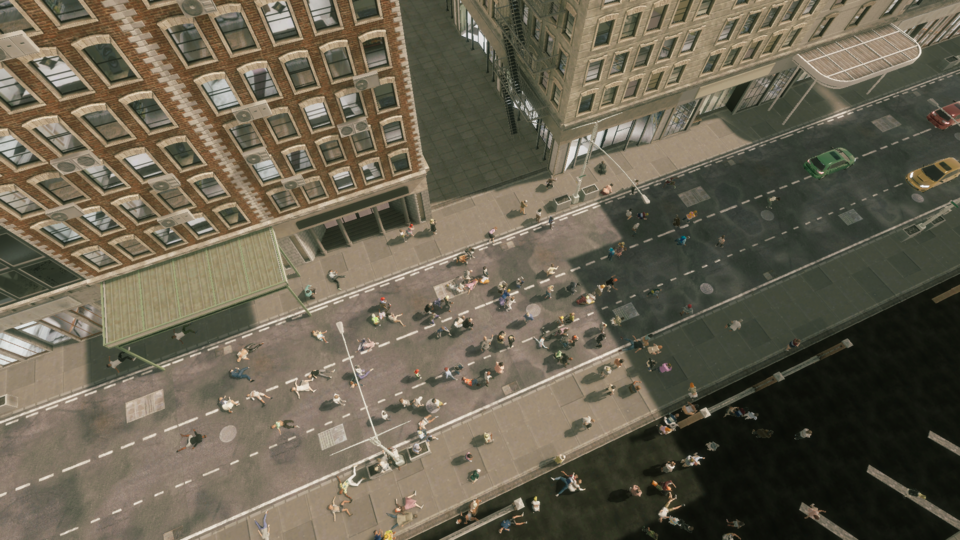} &
\includegraphics[width=.141\linewidth]{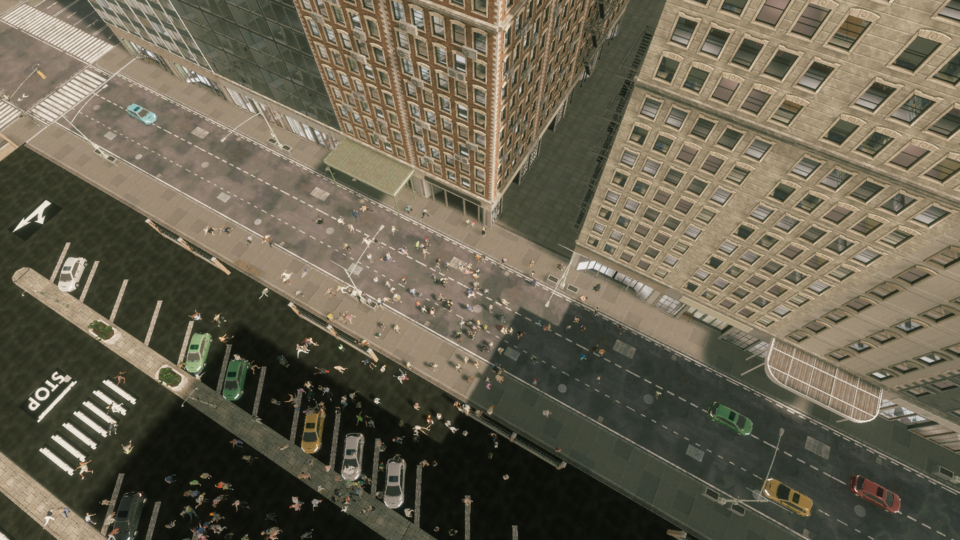} &
\includegraphics[width=.141\linewidth]{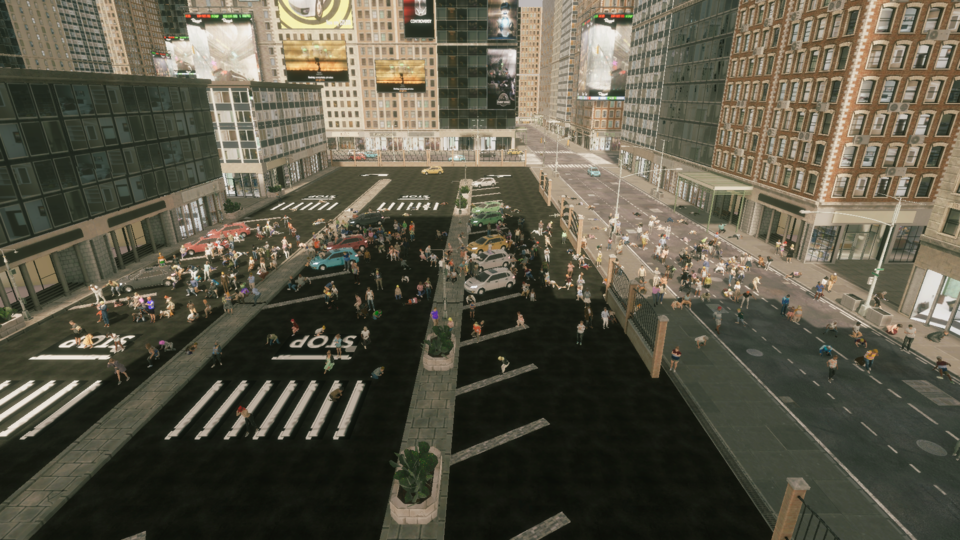} &
\includegraphics[width=.141\linewidth]{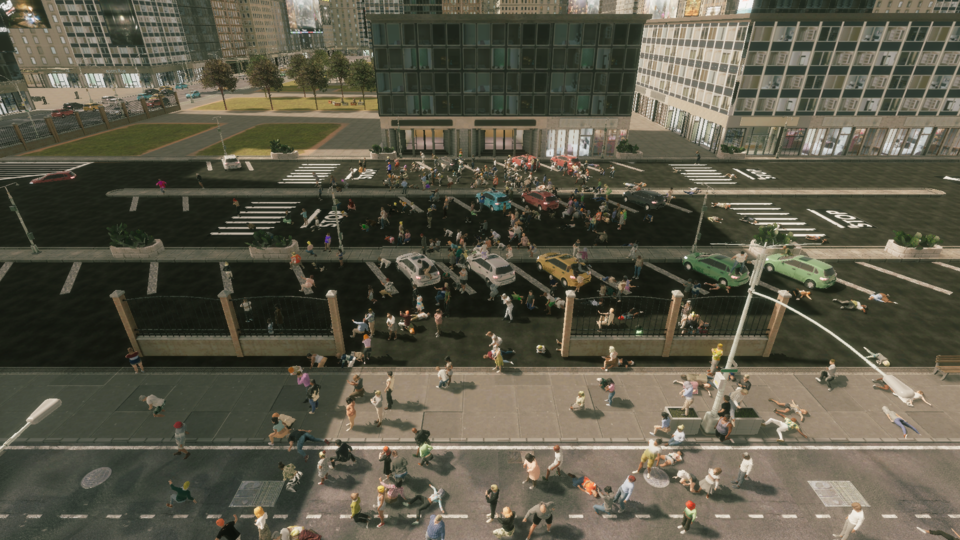} &
\includegraphics[width=.141\linewidth]{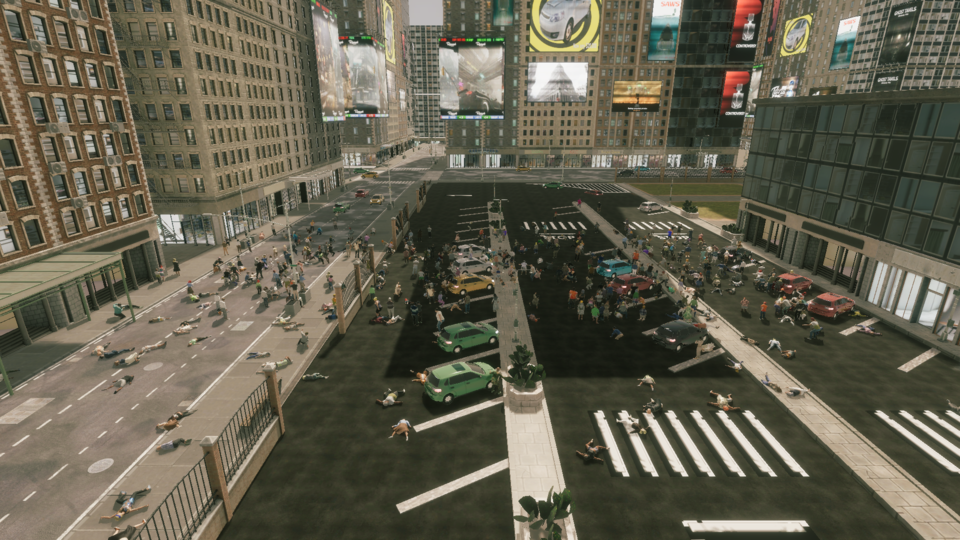} &
\includegraphics[width=.141\linewidth]{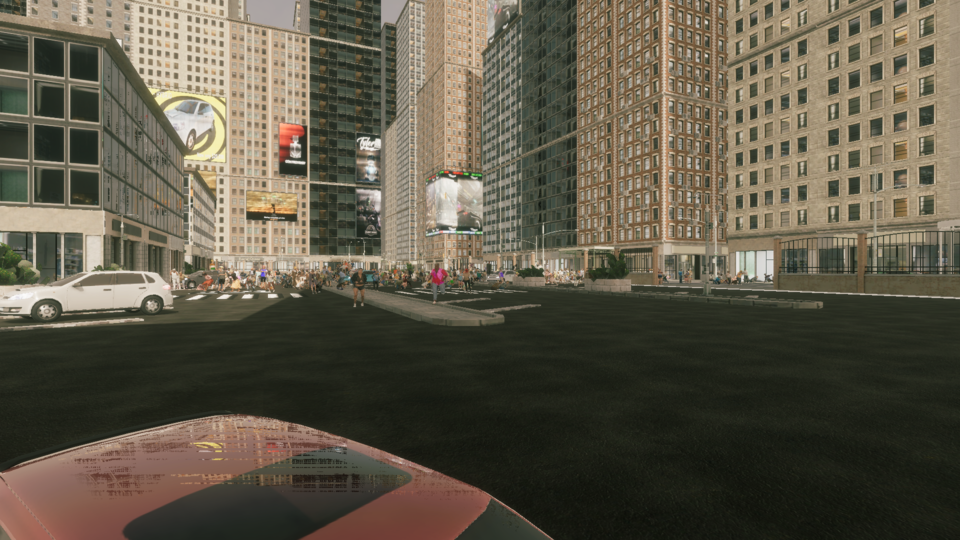} \\
\includegraphics[width=.141\linewidth]{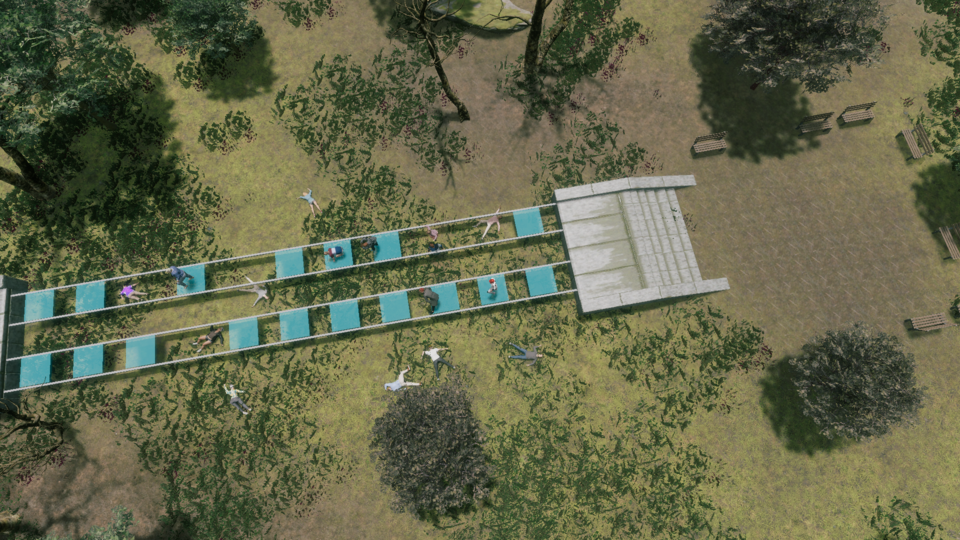} &
\includegraphics[width=.141\linewidth]{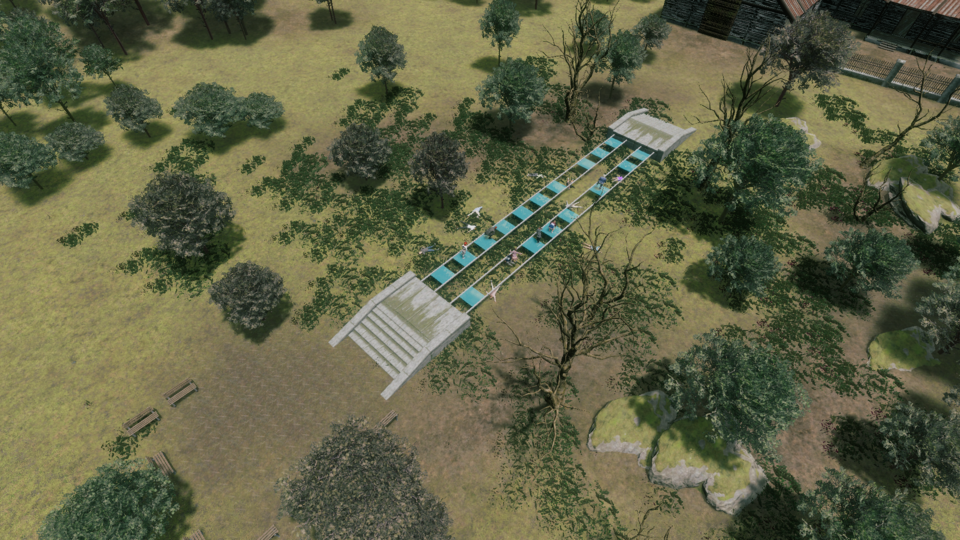} &
\includegraphics[width=.141\linewidth]{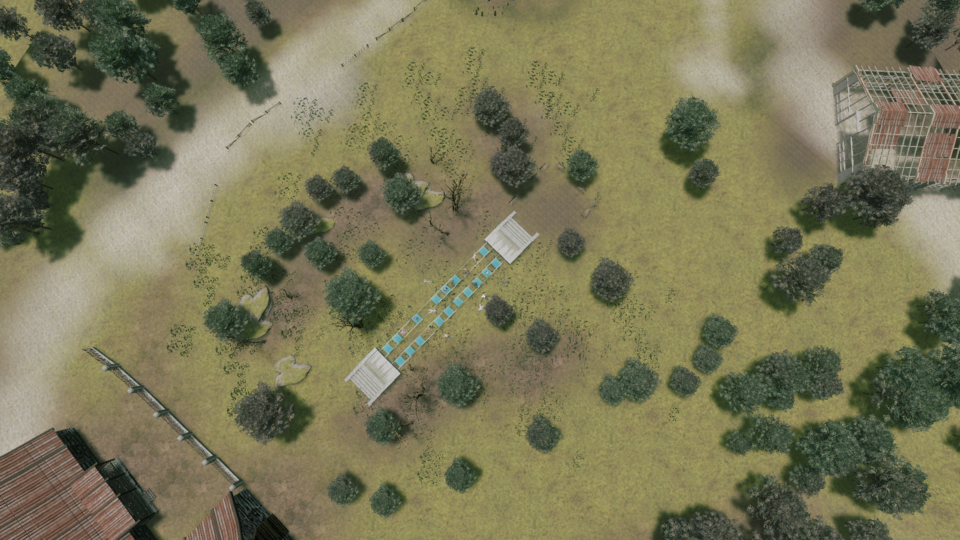} &
\includegraphics[width=.141\linewidth]{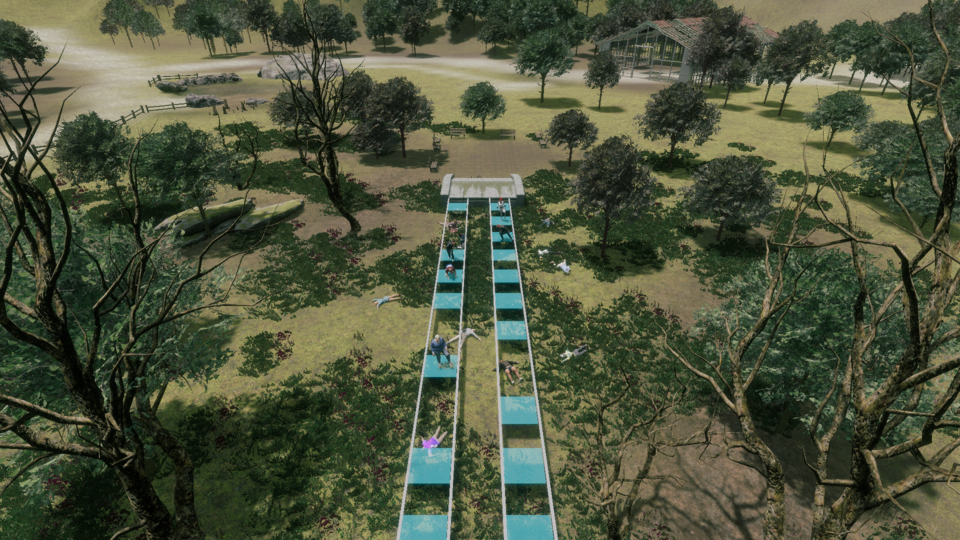} &
\includegraphics[width=.141\linewidth]{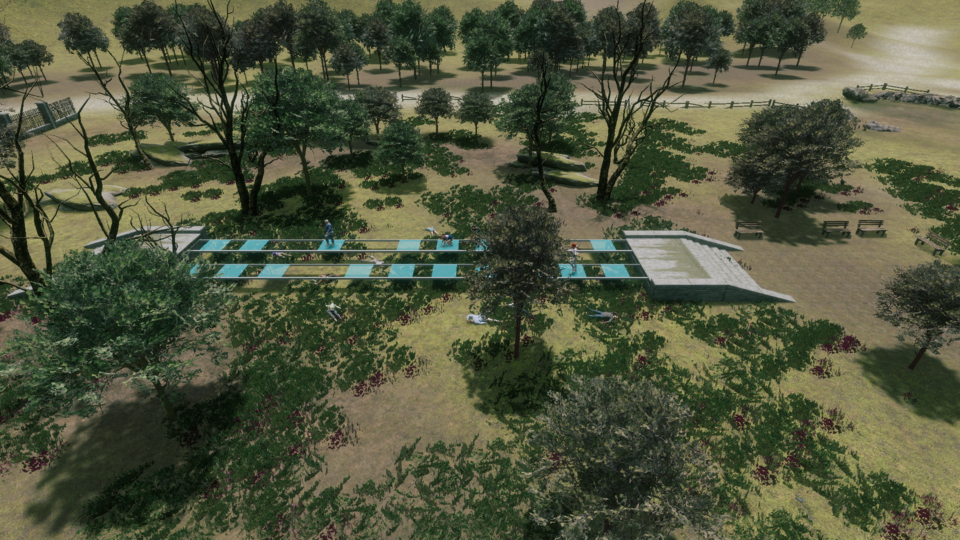} &
\includegraphics[width=.141\linewidth]{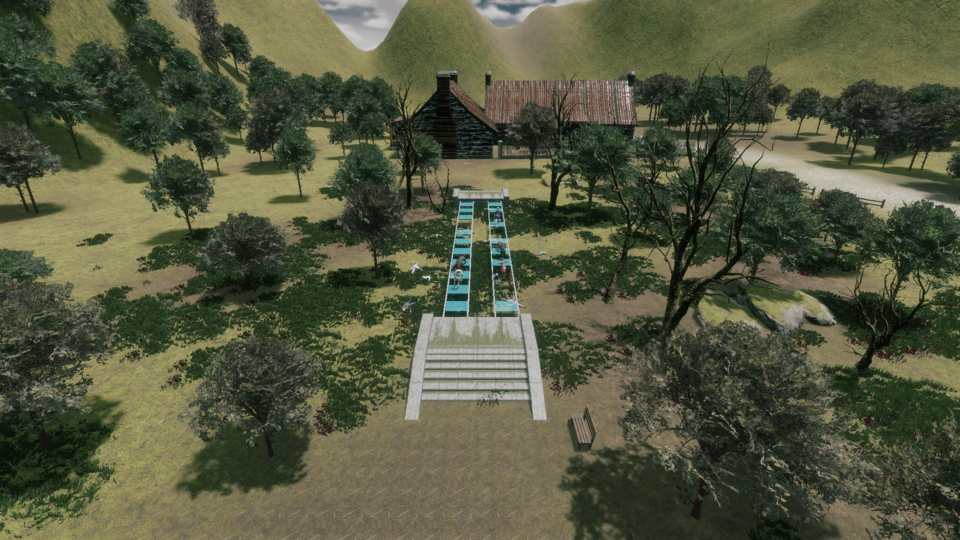} &
\includegraphics[width=.141\linewidth]{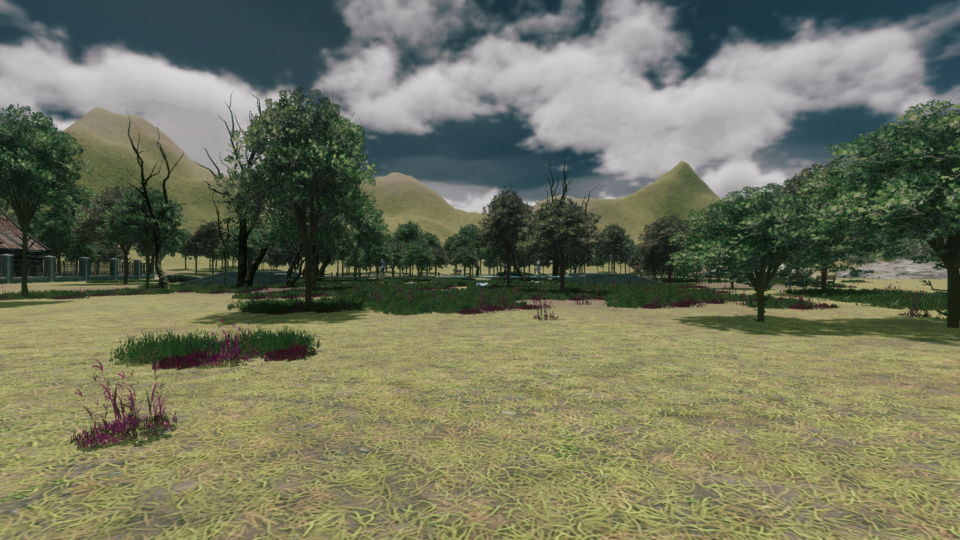} 
\end{tabular}

\vspace{-0.2cm}
\caption{{\bf Multiple viewpoints used in SynPlay.} Multiple camera viewpoints allow substantial variations in appearance for the same human subject with identical pose.}
\label{fig:multi_perspectives}

\end{figure*}


\noindent{\bf Multiple viewpoints.} The camera viewpoints within SynPlay are diversified by implementing three widely used types of image-capturing platforms in the real world: UAV, UGV, and CCTV. They cover a variety of image-capturing properties like static/dynamic and ground/aerial. Viewpoint diversity is acquired by controlling the locations and the focal points of the cameras. Three UAVs, three CCTVs, and one UGV have been deployed (Figure~\ref{fig:multi_perspectives}), resulting in seven unique viewpoints for every game sequence. The UAVs are deployed to fly at various random locations while maintaining altitudes of low ($\sim$30$m$), medium ($\sim$50$m$), and high ($\sim$100$m$). CCTVs are located at a height of 15$m$ at the front, back, and either side of the game playground. UGV images are captured assuming that a vehicle is randomly roaming on the ground. The focal points are set at several locations close to the area where the game usually takes place. For the UAVs, the focal point is changed to a random location every 10 $\mathit{sec}$, where each change takes 5 $\mathit{sec}$ to be fixed at a location for another 5 $\mathit{sec}$. Focal points of the CCTVs and the UGV do not change once determined.\smallskip

\noindent{\bf Dataset specification.} We created ten scenarios for each game, resulting in 60 game scenarios in total. Frames were rendered from seven different camera viewpoints at 1 fps with the resolution of 1920$\times$1080, resulting in a total of 73,892 images with more than 6.5M human instances. The frame generation rate was selected as 1 fps to avoid including highly redundant human poses. Since our pipeline is frame-rate agnostic, we can provide higher-framed dataset upon request. Taking full advantage of the game engine's ability to generate annotations while rendering the scenes, we provided various types of ground truth annotations useful for various computer vision tasks: 2D/3D bounding boxes, instance-level segmentation masks, depth maps, and human keypoint locations.

\subsection{Other design factors}
\label{ssec:other_design_factors}

\noindent{\bf Characters.} We designed 456 virtual characters using the Character Creator, with each character participating in multiple game scenarios. To vary appearances and avoid biases, each character was uniquely designed by gender, skin color, age, height, obesity (body type), hair (styles and colors), and outfits. We kept the gender ratio between male and female at 1:1 and the ratio of skin color among white, black, yellow, and brown at 1:1:1:1. For age, each character was designed to fall into one of three categories: child, middle-aged, and elderly, and the ratio was set at 1:2:1. For each gender and age group, heights were modeled to follow a bell-shaped distribution, resulting in an overall dataset range of 140 to 190 cm. We manually designated every character with a unique outfit, while setting the hair and obesity aspect as diverse as possible. \smallskip

\noindent{\bf Backgrounds.} For each scenario, we set different environmental factors: sites, lighting conditions, and weather. A game takes place at one of ten custom-built sites, including five urban locations (three city areas, a construction site, and a factory) and five natural sites (a green area, snowy field, desert, meadow, and beach). Multiple locations within each site map can serve as playgrounds. To vary lighting, we consider five different times of day: dawn, morning, noon, afternoon, and sunset. We use three types of weather: sunny, foggy, and rainy. All of these are randomly determined for each scenario.
\section{Task Evaluation}
\label{sec:task_eval}

In line with the inherent purpose of synthetic data to serve as supplemental training data, we use the entire SynPlay dataset to train models for various computer vision tasks and evaluate its positive impact towards task performance. Our major counterpart models in evaluation are \emph{trained-from-scratch}, which are trained only on real images (denoted as \textcolor{gray}{real} in evaluation tables). We also validate the advantage of using SynPlay over other synthetic datasets.

\begin{table*}[t]
\centering
\caption{{\bf Comparison with other synthetic datasets on aerial-view human detection and semantic segmentation.} The numbers in parentheses are the performance gaps from the model trained without synthetic data (`\textcolor{gray}{real}'). Positive and negative performance gaps are indicated in \textcolor{teal}{green} and \textcolor{red}{red} fonts, respectively. The best accuracy for each setting is shown in {\bf bold}. \textit{Notations}: `+ real' represents a model pre-trained with synthetic data and fine-tuned on the train split of the corresponding `real' dataset. Performances are evaluated on test split of corresponding `real' data. `s', `m', and `l' represent three YOLO v8 models with different architectures.}
\label{tab:aerial-view_task}
\vspace{-0.6cm}

\begin{tabular}[t]{c@{}c}
\centering

\begin{subtable}[t]{.81\textwidth}
\caption{human detection}
\label{tab:results_det}

\resizebox{\textwidth}{!}{%
\setlength{\tabcolsep}{4.0pt}
\renewcommand{\arraystretch}{1.2}
\begin{tabular}{l|ccc|ccc|ccc}
\multirow{2}{*}{data in training} & \multicolumn{3}{c|}{VisDrone~\citep{PZhuTPAMI2022}} & \multicolumn{3}{c|}{Okutama-action~\citep{MBarekatainCVPRW2017}} & \multicolumn{3}{c}{Semantic Drone~\citep{ICGlink}} \\
& s & m & l & s & m & l & s & m & l \\\Xhline{1.2pt}
\textcolor{gray}{real} & \textcolor{gray}{19.72/47.43}& \textcolor{gray}{21.14/49.52} & \textcolor{gray}{21.60/51.10}& \textcolor{gray}{27.40/75.17} & \textcolor{gray}{28.99/76.60} & \textcolor{gray}{31.53/78.78}& \textcolor{gray}{~~44.00/~~77.20}& \textcolor{gray}{~~44.52/~~78.52}& \textcolor{gray}{~~42.62/~~79.87}\\\hline
Archangel~\citep{YShenAccess2023} & ~~0.23/~~0.63& ~~0.38/~~0.98 & ~~0.59/~~1.48& ~~2.59/~~8.45& ~~3.90/10.13 & ~~2.83/~~9.12 & ~~~~0.64/~~~~1.59& ~~~~2.42/~~~~5.37& ~~~~0.94/~~~~1.62\\
SynDrone~\citep{GRizzoliICCVW2023} & ~~0.31/~~0.81 & ~~0.36/~~0.84 & ~~0.71/~~1.89 & ~~0.00/~~0.00 & ~~0.00/~~0.01 & ~~0.00/~~0.00 & ~~~~0.00/~~~~0.00 & ~~~~0.00/~~~~0.00 & ~~~~0.00/~~~~0.00 \\
{\bf SynPlay} & ~~5.29/11.75& ~~4.31/~~9.12 & ~~2.79/~~5.87& 12.74/40.86& ~~8.19/25.43 & ~~8.15/25.23& ~~~~7.02/~~12.21 &  ~~~~9.60/~~15.51& ~~15.71/~~23.59\\\hline
Archangel + real & 18.77/45.39 & 20.25/48.52 & 20.82/49.51 & 30.72/80.35 & {\bf 32.36}/80.63 & 31.71/79.63 & ~~46.60/~~74.07 & ~~48.60/~~75.86 & ~~44.62/~~73.23\\
& \textcolor{red}{(-0.95/-2.04)} & \textcolor{red}{(-0.89/-1.00)} & \textcolor{red}{(-0.78/-1.59)} & \textcolor{teal}{(+3.32/+5.18)} & \textcolor{teal}{(+3.37/+4.03)} & \textcolor{teal}{(+0.18/+0.85)} & \textcolor{teal}{(~~+2.60}\textcolor{red}{/~~-2.13)} & \textcolor{teal}{(~~+4.08}\textcolor{red}{/~~-2.66)} & \textcolor{teal}{(~~+1.00}\textcolor{red}{/~~-6.64)} \\
SynDrone + real & 18.78/45.79 & 20.94/49.44 & 21.97/51.51 & 29.70/77.71 & 31.39/79.42 & 31.24/78.71 & ~~50.93/~~82.28 & ~~53.71/~~85.47 & ~~59.59/~~85.02 \\
& \textcolor{red}{(-0.94/-1.64)} & \textcolor{red}{(-0.20/-0.08)} & \textcolor{teal}{(+0.37/+1.41)} & \textcolor{teal}{(+2.30/+2.54)} & \textcolor{teal}{(+2.40/+2.82)} & \textcolor{red}{(-0.29/-0.07)} & \textcolor{teal}{(~~+6.93/~~+5.08)} & \textcolor{teal}{(~~+9.19/~~+6.95)} & \textcolor{teal}{(+16.97/~~+5.15)} \\
{\bf SynPlay + real} & {\bf 20.88/49.31} & {\bf 22.34/52.12} & {\bf 22.98/52.93} & {\bf 32.47/81.60} & 31.96/{\bf 81.13} & {\bf 33.17/82.52} & {\bf ~~66.52/~~90.33} & {\bf ~~69.46/~~91.35} & {\bf ~~68.82/~~91.37} \\
& \textcolor{teal}{(+1.16/+1.88)} & \textcolor{teal}{(+1.20/+2.60)} & \textcolor{teal}{(+1.38/+1.83)} & \textcolor{teal}{(+5.07/+6.43)} & \textcolor{teal}{(+2.97/+4.53)} & \textcolor{teal}{(+1.64/+3.74)} & \textcolor{teal}{(+22.52/+13.13)} & \textcolor{teal}{(+24.94/+12.83)} & \textcolor{teal}{(+26.20/+11.50)} \\
\end{tabular}
}
\end{subtable}

&

\begin{subtable}[t]{.183\textwidth}
\caption{semantic seg.}
\label{tab:results_semseg}

\centerline{
\resizebox{\textwidth}{!}{%
\setlength{\tabcolsep}{9.0pt}
\renewcommand{\arraystretch}{1.2}
\begin{tabular}{c|c}
Semantic Drone & Aeroscapes \\
\citep{ICGlink} & \citep{INigamWACV2018} \\\Xhline{1.2pt}
\textcolor{gray}{~~~~0.66}& \textcolor{gray}{22.25} \\\hline
~~~~0.74 & ~~0.04 \\
~~~~0.07 & ~~0.00 \\
~~~~8.03 & ~~6.44 \\\hline
~~~~9.28 & 20.61 \\
\textcolor{teal}{(~~+8.62)} & \textcolor{red}{(-1.64)} \\
~~~~5.56 & 24.59 \\
\textcolor{teal}{(~~+4.90)} & \textcolor{teal}{(+2.34)} \\
{\bf ~~23.32} & {\bf 32.19} \\
\textcolor{teal}{(+22.66)} & \textcolor{teal}{(+9.94)} \\
\end{tabular}
}
}
\end{subtable}
\end{tabular}
\end{table*}

\subsection{General tasks: detection and segmentation}
\label{ssec:general_task}

We evaluate the SynPlay dataset on two tasks, human detection and segmentation. These tasks require the ability to identify diverse human appearances in images captured at a distance. To leverage synthetic data during training, we adopt a \emph{pretrain-finetune} strategy, where a model is pre-trained on synthetic data and fine-tuned on target real-world data. The detectors used in the experiments are YOLO v8 models~\citep{YOLOv8} with three different architecture sizes (small, medium, and large). Mask2former~\citep{BChengCVPR2022} with the Swin-Base~\citep{ZLiuICCV2021} backbone was used for segmentation. For evaluation metrics, we use the COCO-style APs which are two bounding box APs (AP$^\text{bb}$ and AP$^\text{bb}_\text{50}$)\footnote{Detection accuracies in the following tables are reported with two numbers in the form of AP$^\text{bb}$/AP$^\text{bb}_\text{50}$.} for human detection and Intersection-over-Union (IoU) for the segmentation.

The main tasks use aerial-view datasets, featuring a wider range of human appearances, making them ideal for validating the design philosophy of the SynPlay. We also conduct experiments on ground-view datasets to evaluate SynPlay on a more widely studied task in the community. \smallskip

\noindent{\bf Aerial-view tasks.} Table~\ref{tab:aerial-view_task} shows the results for aerial-view human detection and semantic segmentation tasks. Overall, for both tasks, using SynPlay in training provides remarkably better accuracy than all the compared cases, including \textcolor{gray}{real} and all other variations involving other synthetic data. 

Notably, the results show that warming up the model with synthetic data before incorporating real data generally does not improve performance, except in the case of SynPlay. In other words, unless the synthetic dataset is properly designed and constructed, we cannot expect performance improvement simply from adding synthetic data to the training process. Moreover, among cases using synthetic data only in training, SynPlay presents unparalleled accuracy. In fact, the results using other sources of synthetic data are so poor that the other sources can be considered useless for this type of dataset utilization. Based on these two observations, \emph{our design strategies for enhancing the diversity and realism of human appearance are shown to be highly effective in meeting expectations}.\smallskip

\noindent{\bf Ground-view tasks.} Table~\ref{tab:ground_view_task} explores the impact of using SynPlay for the general computer vision tasks of ground-view human detection and semantic segmentation. We also evaluate how models perform when only the subset with matching viewpoint (\textit{i.e.}, UGV images in SynPlay) is used in training. Overall, using the entire SynPlay yields the highest accuracy on both tasks, while using the UGV-subset still outperforms the model trained without SynPlay. These results demonstrate that our insight in ensuring diversity by varying the camera viewpoints is effective even in tasks that do not contain such multiple viewpoints. In addition, the greater improvement in semantic segmentation over object detection shows that ensuring diversity is more effective in tasks that require more detailed human representations.\smallskip

\begin{table}[t]
\centering
\caption{{\bf Impact of SynPlay on MS COCO} (\texttt{person} category). \textit{Notation}: `SynPlay-UGV' and `SynPlay-all' are a UGV subset of SynPlay and the entire SynPlay, respectively.}
\label{tab:ground_view_task}
\vspace{-0.6cm}

\begin{tabular}{c@{}c}
\centering

\begin{subtable}[t]{.835\linewidth}
\caption{human detection}
\label{tab:results_ground_view_det}

\resizebox{\textwidth}{!}{%
\setlength{\tabcolsep}{3.5pt}
\renewcommand{\arraystretch}{1.2}
\begin{tabular}{l|ccc}
data in training & s & m & l \\\Xhline{1.2pt}
\textcolor{gray}{real} & \textcolor{gray}{46.19/65.91} & \textcolor{gray}{50.10/69.86} & \textcolor{gray}{52.52/72.15} \\\hline
{\bf SynPlay-UGV + real} & 46.53/66.18 & 50.70/70.37 & 52.69/72.29 \\
& \textcolor{teal}{(+0.34/+0.27)} & \textcolor{teal}{(+0.60/+0.51)} & \textcolor{teal}{(+0.17/+0.14)} \\
{\bf SynPlay-all + real} & {\bf 46.84/66.70} & {\bf 51.12/70.74} & {\bf 53.00/72.59} \\
& \textcolor{teal}{(+0.65/+0.79)} & \textcolor{teal}{(+1.02/+0.88)} & \textcolor{teal}{(+0.48/+0.44)} \\
\end{tabular}
}
\end{subtable}

&

\begin{subtable}[t]{.165\linewidth}
\caption{sem.seg.}
\label{tab:results_ground_view_instseg}

\resizebox{\textwidth}{!}{%
\setlength{\tabcolsep}{10.0pt}
\renewcommand{\arraystretch}{1.2}
\begin{tabular}{c}
\\\Xhline{1.2pt}
\textcolor{gray}{15.10} \\\hline
20.18 \\
\textcolor{teal}{(+5.08)} \\
{\bf 21.57} \\
\textcolor{teal}{(+6.47)} \\
\end{tabular}
}
\end{subtable}

\end{tabular}
\end{table}

\noindent{\bf Combination with MS-COCO for pre-training dataset.} The effect of using pre-training can be greater when applying two or more datasets with complementary properties. Here, we aim to investigate the potential synergy achieved by integrating MS COCO, a real dataset primarily comprising ground-view images, with SynPlay for the task of aerial-view human detection. Table~\ref{tab:synergy_with_coco}, shows all combinations of SynPlay and MS COCO datasets when used for pre-training. The anticipated synergistic effect appears in all cases except in one case (AP$^\text{bb}_\text{50}$ results on Okutama-action) when fine-tuned on the target dataset. Moreover, when fine-tuned on the real dataset, using only SynPlay (SynPlay + real) achieves accuracy that is comparable to, and sometimes even better than, using MS COCO (COCO + real).

Interestingly, the results without fine-tuning show a different trend. On Okutama-action and Semantic Drone cases, using MS COCO performed better than other two baselines, while SynPlay specifically showing a much lower accuracy. We observe that synthetic data still lags behind real-world data in many respects, highlighting the need for further research to bridge the gap.

\begin{table}[t]
\centering
\caption{{\bf Synergy impact with MS COCO} on aerial-view human detection. YOLO v8 model with a medium size architecture is used.}
\label{tab:synergy_with_coco}
\vspace{-0.3cm}

\resizebox{\linewidth}{!}{%
\setlength{\tabcolsep}{7.0pt}
\renewcommand{\arraystretch}{1.2}
\begin{tabular}{l|ccc}
data in training & VisDrone & Okutama-action & Semantic Drone \\\Xhline{1.2pt}
\textcolor{gray}{real} & \textcolor{gray}{21.14/49.52} & \textcolor{gray}{28.99/76.60} & \textcolor{gray}{44.52/78.52} \\\hline
COCO & ~~7.16/16.46 & 15.17/48.28 & 34.74/56.39 \\
SynPlay & ~~4.31/~~9.12 & ~~8.19/25.43 & ~~9.61/15.52 \\
COCO + SynPlay & 11.49/25.20 & 14.68/49.82 & 18.60/31.03 \\\hline
COCO + real & 22.11/51.73 & 32.26/80.10 & 65.72/89.20 \\
SynPlay + real & 22.34/52.13 & 31.96/{\bf 81.13} & 69.46/91.35 \\
COCO + SynPlay + real & {\bf 22.78/53.01} & {\bf 33.82}/79.44 & {\bf 73.52/92.80} \\
\end{tabular}
}
\end{table}

\subsection{Data-scarce tasks: Few-shot and cross-domain learning}
\label{ssec:few_shot}

We compare SynPlay with other synthetic datasets in addressing data scarcity in aerial-view human detection, which lacks training data more severely than ground-view detection. Following the data-scarce task setups of ~\cite{YShenCVPR2023}, we evaluate few-shot and cross-domain learning. Models are trained with 20 and 50 VisDrone images (`Vis-20/50’) and tested on VisDrone, Okutama-action, and Semantic Drone, with the latter two serving as cross-domain evaluations. To mitigate randomness in selecting real training images, all reported accuracy values are averaged over three runs.

As baseline methods leveraging synthetic data in training, we use a pretrain-finetune strategy (PT-FT) and Progressive Transformation Learning (PTL)~\citep{YShenCVPR2023}. PTL is a progressive data augmentation approach that iteratively expands the training set by adding a subset of synthetic data, which is transformed to look real. In each PTL iteration, a subset of the synthetic data is selected, such that synthetic data that is closer to the real dataset is selected more often. For the data-scarce tasks experimented in \cite{YShenCVPR2023}, PTL was better than PT-FT while both outperformed the cases without synthetic data. We used RetinaNet~\citep{TLinICCV2017} as the detector.\footnote{PTL was designed to be suitable for RetinaNet. For a fair comparison between PTL and PT-FT, we used RetinaNet instead of YOLO v8 for this experiment.}\smallskip

\begin{table*}[t]
\caption{{\bf Few-shot and cross-domain learning accuracy} on aerial-view human detection. To clarify, the accuracy on Okutama-action and Semantic Drone refers to cross-domain learning performance. \textit{Notation}: `Archangel*' is an expanded `Archangel' to be pose-diversified~\citep{YShenArXiv2024}.}
\label{tab:results_data_scarce_Tasks}
\vspace{-0.3cm}
\centering
\resizebox{\linewidth}{!}{%
\setlength{\tabcolsep}{12.0pt}
\renewcommand{\arraystretch}{1.2}
\begin{tabular}{l|c|ccc|ccc}
& & \multicolumn{3}{c|}{Vis-20} & \multicolumn{3}{c}{Vis-50}\\
\multicolumn{1}{c|}{data in training} & method & \multicolumn{1}{c}{VisDrone} & \multicolumn{1}{c}{Okutama-action} & \multicolumn{1}{c|}{Semantic Drone} & \multicolumn{1}{c}{VisDrone} & \multicolumn{1}{c}{Okutama-action} & \multicolumn{1}{c}{Semantic Drone} \\\Xhline{1.2pt}
\textcolor{gray}{real} & & \textcolor{gray}{~~~0.58/~~2.27} & \textcolor{gray}{~~~~~~3.64/~~14.54} & \textcolor{gray}{~~~~0.62/~~~~1.89} & \textcolor{gray}{~~0.76/~~3.30} & \textcolor{gray}{~~~~7.82/~~28.66} & \textcolor{gray}{~~~~1.30/~~~~5.65} \\\hline
~~+ Archagel & \multirow{4}{*}{PTL} & ~~~2.07/~~6.72 & ~~~~~~7.90/~~31.53 & {\bf ~~~~8.81}/{\bf ~~33.71} & ~~2.92/~~9.26 & ~~11.49/~~42.51 & {\bf ~~~~8.98}/{\bf ~~33.21} \\
~~+ Archagel* & & ~~~2.26/~~7.39 & ~~~~~~8.95/~~36.97 & ~~~~6.45/~~26.13 & ~~2.99/~~9.42 & ~~12.89/~~47.24 & ~~~~6.29/~~25.50\\
{\bf ~~+ SynPlay} & & {\bf ~~~3.08/~~9.03} & {\bf ~~~~14.39/~~49.53} & ~~~~6.94/~~24.22 & {\bf ~~3.71}/{\bf 11.20} & {\bf ~~15.67}/{\bf ~~52.06} & ~~~~7.74/~~26.99 \\
& & \textcolor{teal}{(+2.50/+6.76)} & \textcolor{teal}{~(+10.75/+34.99)} & \textcolor{teal}{(~~+6.32/+22.33)} & \textcolor{teal}{(+2.95/+7.90)} & \textcolor{teal}{~~(+7.85/+23.40)} & \textcolor{teal}{~~(+6.44/+21.34)}\\\hline
~~+ Archagel & \multirow{4}{*}{PT-FT} & ~~~0.76/~~2.48 & ~~~~~~4.24/~~17.17 & ~~~~6.53/~~23.67 & ~~1.29/~~3.76 & ~~~~5.32/~~20.96 & ~~~~7.10/~~27.95 \\
~~+ Archagel* & & ~~~1.21/~~4.02 & ~~~~~~9.14/~~34.70 & ~~~~8.20/~~28.80 & ~~1.84/~~5.37 & ~~10.39/~~36.83 & ~~~~8.63/~~30.09 \\
{\bf ~~+ SynPlay} & & {\bf ~~~2.94}/{\bf ~~9.38} & {\bf ~~~~12.19}/{\bf ~~40.88} & {\bf ~~11.32}/{\bf ~~37.30} & {\bf ~~3.72}/{\bf 11.87} & {\bf ~~13.66}/{\bf ~~44.06} & {\bf ~~12.76}/{\bf ~~41.66} \\
& & \textcolor{teal}{(+2.36/+7.11)} & \textcolor{teal}{~(~~+8.55/+26.34)} & \textcolor{teal}{(+10.70/+35.41)} & \textcolor{teal}{(+2.96/+8.57)} & \textcolor{teal}{~~(+5.84/+15.40)} & \textcolor{teal}{(+11.46/+36.01)}\\
\end{tabular}
}
\end{table*}

\noindent{\bf Comparison with other synthetic data.} In Table~\ref{tab:results_data_scarce_Tasks}, we compare the detection accuracy of the models trained with different synthetic datasets on the few-shot and cross-domain learning tasks. With PT-FT, SynPlay achieved significantly better accuracy than other synthetic datasets across all three datasets. With PTL, SynPlay performed the best on VisDrone and Okutama-action for both Vis-20 and Vis-50 settings. Even on Semantic Drone, which shows an unusual performance trend, the best performance was achieved when SynPlay was used via PT-FT.

In addition, compared to SynPlay's performance improvement on general tasks (in Table~\ref{tab:aerial-view_task}), the improvement achieved on data-scarce setting by SynPlay is much greater on VisDrone and Okutama-action on data-scarce tasks. This demonstrates that \emph{SynPlay effectively meets the demand for additional data in data-scarce setting}, which is greater than that in general tasks. We will discuss the unexpected performance trends on Semantic Drone in more detail in Sec.~\ref{sec:discussion}.\smallskip

\begin{figure*}[t]
\centering
\setlength{\tabcolsep}{10pt}
\begin{tabular}{ccc}
\includegraphics[width=.28\linewidth]{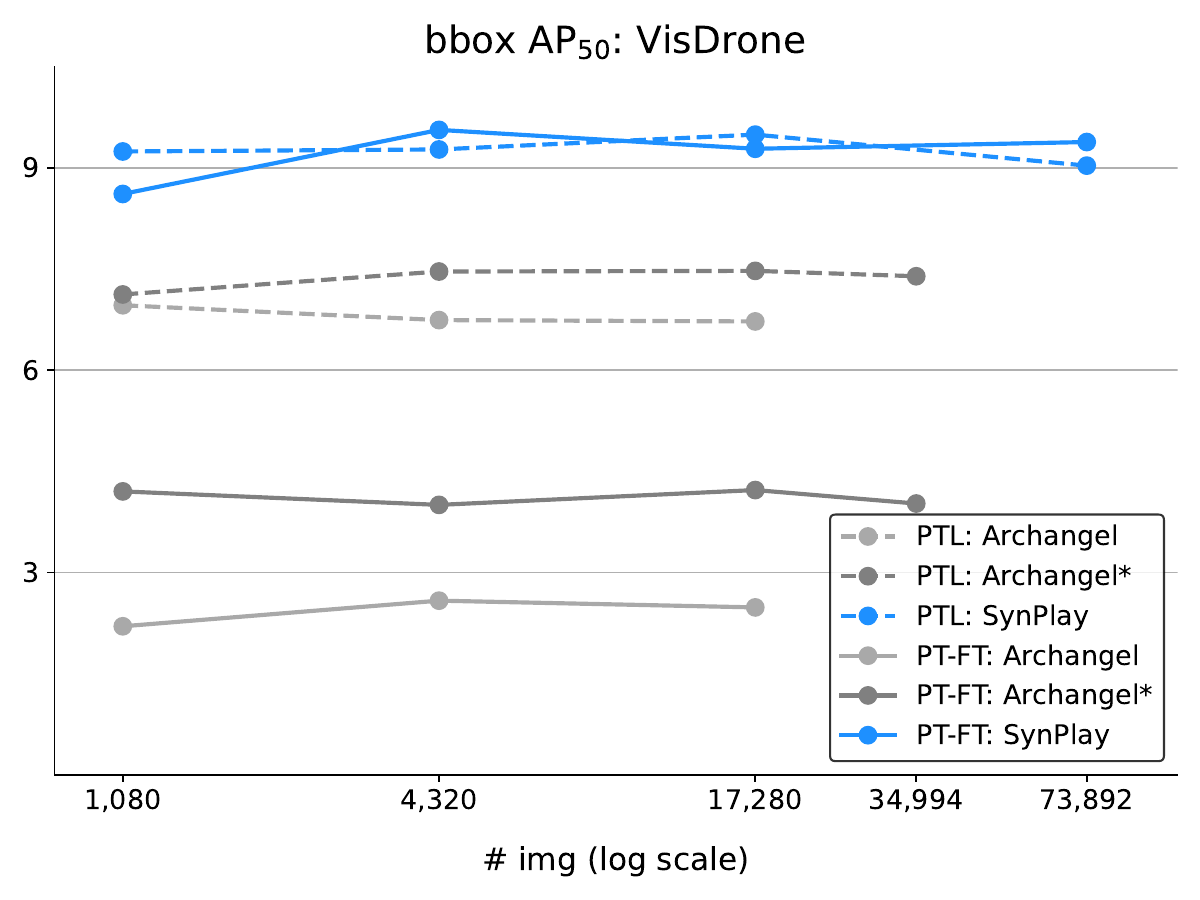} &
\includegraphics[width=.28\linewidth]{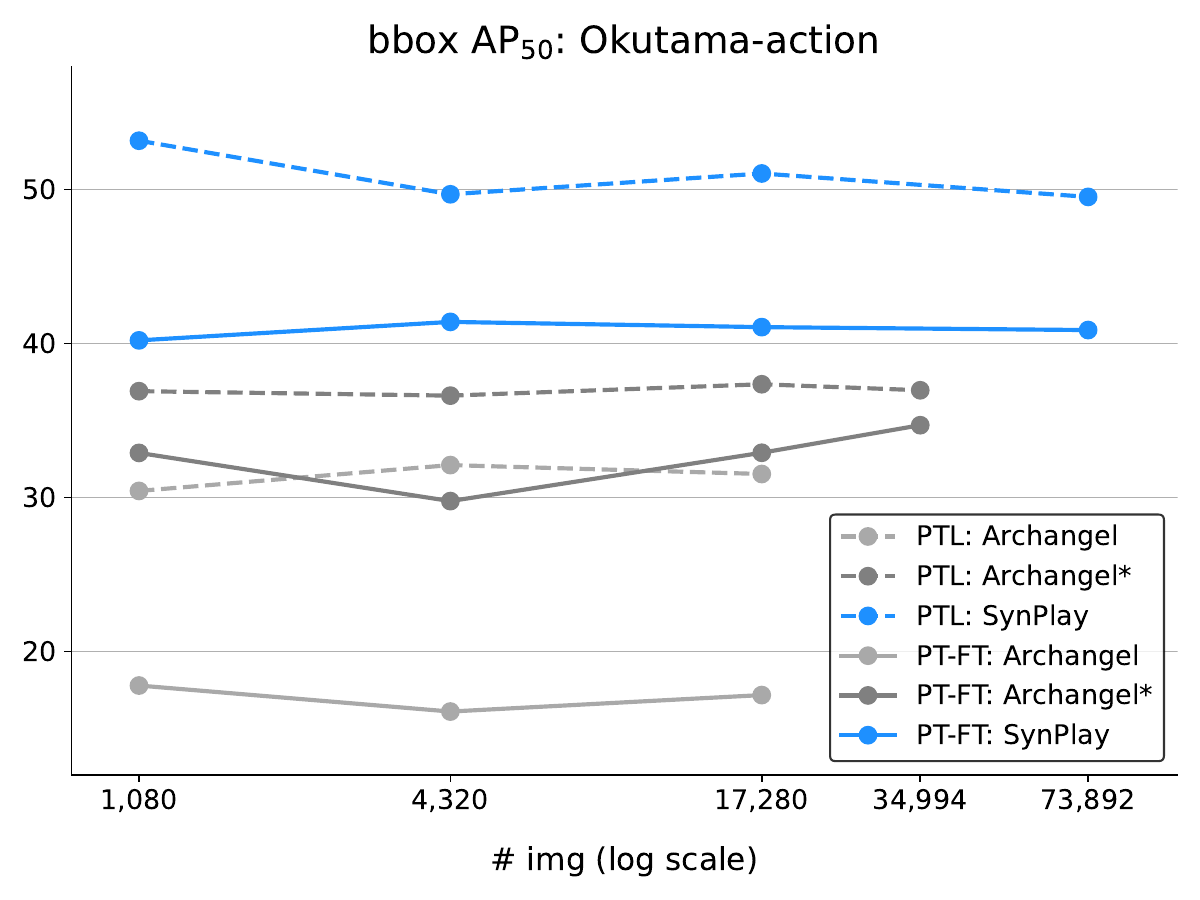} &
\includegraphics[width=.28\linewidth]{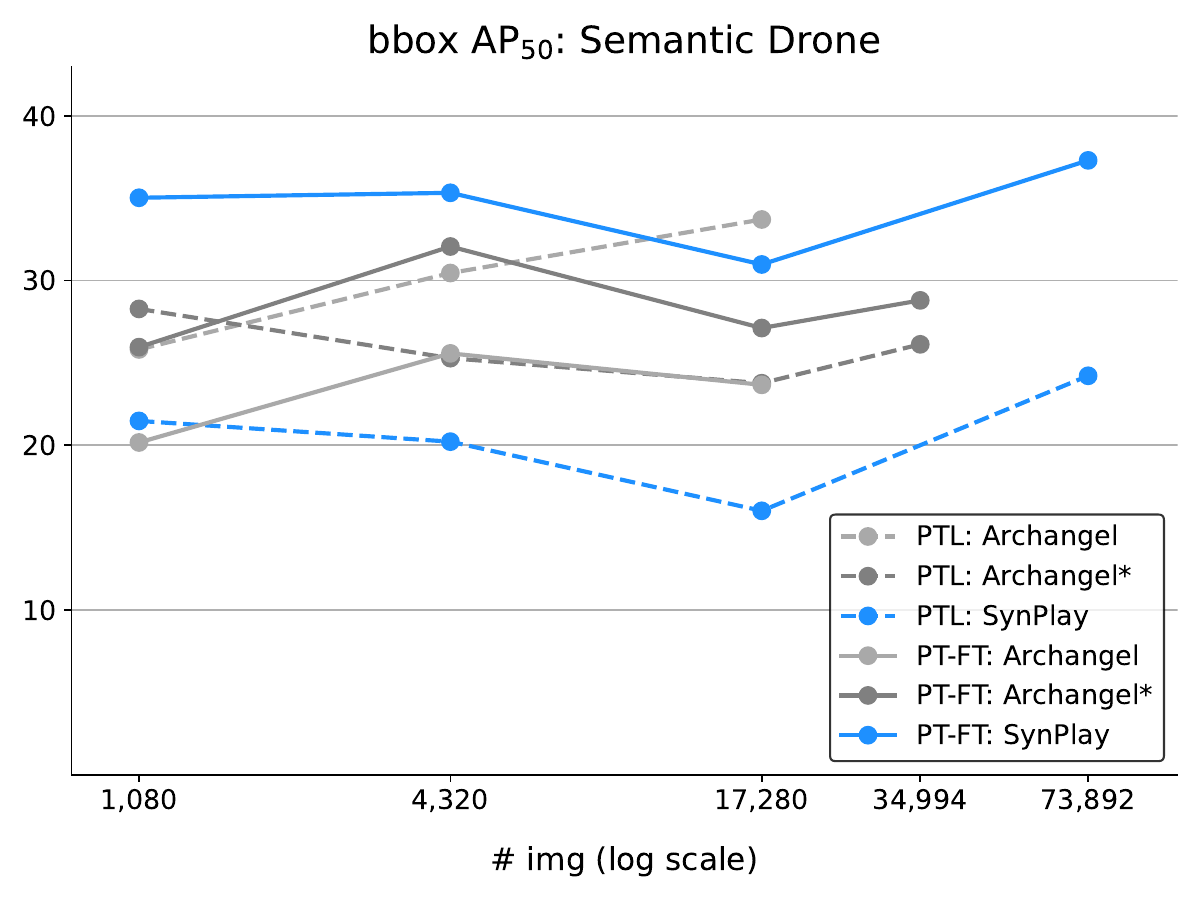} \\
\end{tabular}

\vspace{-0.4cm}
\caption{{\bf Scaling behavior of synthetic datasets} under the Vis-20 setup (AP$_\text{50}^\text{bb}$). Scaling behavior of each dataset is compared by randomly sampled subsets of 1,080, 4,320, and 17,280 images, which correspond to 1/16th the size, 1/4th the size, and the size of Archangel. For reference, the sizes of Archange* and SynPlay are 34,994 and 73,892, respectively.}
\label{fig:scalability_set_size}
\end{figure*}

\noindent{\bf Scaling behaviors.} To fairly validate the performance comparison without being affected by dataset size, we explore the scaling behavior of synthetic datasets. In Fig~\ref{fig:scalability_set_size}, we compare the detection accuracy of three synthetic datasets at multiple points where the datasets are randomly sampled to have the same size. On all three test sets, the best performing models use SynPlay in training, \textit{i.e.}, SynPlay + PTL on VisDrone and Okutama-action, SynPlay + PT-FT on Semantic Drone. \emph{The performance gain achieved using SynPlay is not simply due to the large size of the dataset.}

\subsection{Image quality evaluation}
\label{ssec:qualititative_analysis}

\begin{table}[t]
\caption{{\bf FID comparison.} In FID calculation, VisDrone serves as a reference representing real aerial-view human data.}
\label{tab:FID}
\vspace{-0.3cm}

\centering
\resizebox{\linewidth}{!}{%
\setlength{\tabcolsep}{12.0pt}
\renewcommand{\arraystretch}{1.2}
\begin{tabular}{c|cccc}
COCO & Archangel & Archangel* & SynDrone & {\bf SynPlay} \\\Xhline{1.2pt}
48.16 & 67.20 & 67.20 & 21.66 & {\bf 18.36} \\
\end{tabular}
}
\end{table}

In Table~\ref{tab:FID}, for all synthetic data used in training, we calculate FID (Fr{\' e}chet Inception Distance)~\citep{MHeuselNeurIPS2017} to assess their fidelity and diversity. SynPlay presents the best score, which aligns well with our task results. It suggests that SynPlay's superior task performance is achieved by better fidelity and diversity, which are our goals in designing SynPlay. Moreover, the better FID of SynPlay compared to MS COCO, which mainly includes ground-view images, is also reported, supporting the hypothesis that adopting multiple viewpoints effectively diversifies human appearances.
\section{Discussion and Conclusion}
\label{sec:discussion}

\begin{table}[t]
\caption{{\bf Proportion of nadir-view instances in synthetic datasets} used in data-scarce tasks. Instances with camera viewing angle from ground greater than 71.57$^\circ$ (largest viewing angle from ground in Archangel*) are considered as nadir-view instances.}
\label{tab:nari_view_portion}
\vspace{-0.3cm}

\centering
\resizebox{\linewidth}{!}{%
\setlength{\tabcolsep}{30.0pt}
\renewcommand{\arraystretch}{1.2}
\begin{tabular}{ccc}
Archangel & Archangel* & SynPlay \\\Xhline{1.2pt}
25.00\% & 12.82\% & 4.24\% \\
\end{tabular}
}
\end{table}

\noindent{\bf Peculiar performance trend on Semantic Drone.} Following conflicting phenomena were observed in experimental results when tested on Semantic Drone:
\begin{itemize}
\item[\labelitemii] \textit{When using synthetic data in training via PTL for data-scarce tasks, involving SynPlay under-performed when compared to cases using other synthetic datasets.}
\item[\labelitemii] \textit{The performance gain acquired by incorporating a synthetic data during training is remarkably large for Semantic Drone when compared to other two cases, while SynPlay showcasing the most substantial gain.}
\end{itemize}
Most human instances in Semantic Drone are taken from nadir views, while VisDrone and Okutama-action have fewer nadir-views. The portion of nadir-view instances in SynPlay is the smallest among the synthetic datasets used in data-scarce tasks (Table~\ref{tab:nari_view_portion}). As PTL continues to prioritize samples from synthetic data that closely resemble the seed data (\textit{i.e.,} VisDrone) for training, the reduced selection of nadir-view instances from the SynPlay may result in a lower gain (first phenomenon). On the other hand, the second phenomenon indicates that ensuring greater diversity via using supplemental synthetic data has greater impact on Semantic Drone, which lacks diversity due to its limited viewpoints. Moreover, SynPlay that is less similar to Semantic Drone while being more diverse than other synthetic datasets, shows the largest impact, supporting our claim that improving the diversity is effective in constructing a better synthetic data.\smallskip



\noindent{\bf Conclusion.} 
We presented SynPlay, a large-scale synthetic human dataset that addresses the critical challenge of identifying humans in aerial views where subjects appear as only tens of pixels. SynPlay departs from traditional datasets by introducing a \textit{rule-guided motion generation}, enabling \textit{effectively uncountable motion variations} and spontaneous interactions. Through a \textit{multi-perspective setup}, SynPlay uniquely supports near-to-far human localization across extreme viewpoint variations. Extensive experiments confirm that SynPlay significantly improves human detection, segmentation, and keypoint estimation, particularly in \textit{few-shot and cross-domain settings}. By advancing both dataset design and practical model performance, SynPlay establishes a new foundation for long-range human analysis and aerial-view perception.



\appendix
\section*{Appendix}

\section{Comparison to Other Human Datasets}
\label{sec:dataset_comparison}

\begin{table*}[t!]
\caption{{\bf Comparison of human datasets.} `\#inst/img' is acquired only on images that contain human. `\#motion' indicates the number of unique motions depicted in the dataset, except the ones with the subscript `pose' which indicate the number of static poses. Since a single motion can consist of multiple number of unique poses, \#motion is generally smaller than the number of poses. For certain datasets, the test set without available labels is excluded from this comparison. `uncountable' indicates that the number of human motions included in the set is countless/uncountable.}
\label{tab:data_comparison}
\centering
\resizebox{\linewidth}{!}{%
\setlength{\tabcolsep}{10.0pt}
\renewcommand{\arraystretch}{1.2}
\begin{tabular}{l|c|ccc|ccc}
\multicolumn{1}{c|}{dataset} & domain & \#inst & \#img & \#inst/img & natural motion & \#motion & viewpoint \\\Xhline{1.2pt}
\multicolumn{8}{l}{\textit{ground-view}} \\
VOC \texttt{12}~\citep{MEveringhamIJCV2015} & real & 10K & 11.5K & 2.48 & daily & uncountable & near \\
KITTI~\citep{AGeigerCVPR2012} & real & 4.5K & 7.5K & 2.52 & daily & 2 & near \\
COCO \texttt{Dev17}~\citep{TLinECCV2014} & real & 649K & 164K & 9.72 & daily & uncountable & near \\  
MPII Human Pose~\citep{MAndrilukaCVPR2014} & real & 40K & 24.9K & 1.61 & daily & 20 & near \\
Cityscapes~\citep{MCordtsCVPR2016} & real & 21.4K & 5K & 7.85 & daily & 2 & near \\
ADE20K~\citep{BZhouCVPR2017} & real & 30K & 27.5K & 4.36 & daily & uncountable & near \\
Human-Art~\citep{XJuCVPR2023} & real & 123K & 50K & 2.46 & art & uncountable & near \\\hline
GTA5~\citep{SRichterECCV2016} & synth & 1.4M & 1.4M & 1 & \xmark & $20K_\text{pose}$ & near \\
SURREAL~\citep{GVarolCVPR2017} & synth & 6.5M & 6.5M & 1 & detail+mocap & 23 & near \\
SOMAset~\citep{IBarbosaCVIU2018} & synth & 100K & 100K & 1 & detail+mocap & $250_\text{pose}$ & near \\
PersonX~\citep{XSunCVPR2019} & synth & 273K & 273K & 1 & \xmark & $4_\text{pose}$ & near \\
UnrealPerson~\citep{TZhangCVPR2021} & synth & 120K & 120K & 1 & \xmark & 2 & near \\
AGORA~\citep{PPatelCVPR2021} & synth & $\cdot$ & 19K & 1$\sim$15 & detail+mocap & $4,240_\text{pose}$ & near \\
BEDLAM~\citep{MBlackCVPR2023} & synth & $\cdot$ & 380K & 1$\sim$10 & detail+mocap & $2,311_\text{pose}$ & near
\\\hline
\multicolumn{8}{l}{\textit{aerial-view}} \\
Okutama-action~\citep{MBarekatainCVPRW2017} & real & $\cdot$ & 77K & $\sim$9 & detail & 12 & med \\
Semantic Drone~\citep{ICGlink} & real & 1.5K & 400 & 4.16 & daily & unspecified & med \\
UAVid~\citep{YLyuPRS2020} & real & 4.7K & 420 & 20.06 & daily & unspecified & med$\sim$far \\
VisDrone~\citep{PZhuTPAMI2022} & real & 109K & 40.0K & 15.42 & daily & unspecified & med \\
Archangel-real~\citep{YShenAccess2023} & real & 165.6K & 41.4K & 4 & detail & $3_\text{pose}$ & near$\sim$far \\
Archangel-mannequin~\citep{YShenAccess2023} & real & $\cdot$ & 178.8K & 6$\sim$7 & detail & $3_\text{pose}$ & near$\sim$far
\\\hline
Archangel-synth~\citep{YShenAccess2023} & synth & 4.4M & 4.4M & 1 & \xmark & $3_\text{pose}$ & near$\sim$far \\
SynDrone~\citep{GRizzoliICCVW2023} & synth & 803K & 72K & 11.15 & \xmark & 2 & med$\sim$far \\
CARGO~\citep{QZhangCVPR2024} & synth & 108K & 108K & 1 & \xmark & 2 & near$\sim$far \\
{\bf SynPlay} & synth & 6.5M & 73K & 88.40 & rule+mocap & uncountable & near$\sim$far \\
\multicolumn{8}{l}{* natural motion} \\
\multicolumn{8}{l}{~~$\cdot$ daily: human motions engaged in daily activity} \\
\multicolumn{8}{l}{~~$\cdot$ art: human motions shown in works of art} \\
\multicolumn{8}{l}{~~$\cdot$ detail: human motions captured by `detail-guided design'} \\
\multicolumn{8}{l}{~~$\cdot$ rule: human motions captured by `rule-guided design'} \\
\multicolumn{8}{l}{~~$\cdot$ +mocap: human motions captured using a motion scanner} \\
\end{tabular}
}
\end{table*}

Table~\ref{tab:data_comparison} provides a comparative analysis of various human datasets, categorized as real or synthetic and captured from either ground or aerial perspectives. Key observations from this comparison are outlined below:
\begin{enumerate}
\item Aerial-view sets, \textit{thanks to their wide viewing angles}, generally have \textit{more human instances per image} than ground-view sets, except for few cases that employed a fixed number of actors in a real set or designing one instance per image in a synthetic set.
\item Aerial-view sets generally contain a wider range of viewpoints. (mostly near$\sim$far)
\item For existing synthetic datasets, aerial-view sets typically feature \textit{fewer motion variations} compared to ground-view sets. This is because \textit{aerial-view datasets often prioritize leveraging a wide range of viewpoints over expanding the variety of human motions}.
\item Rule-guided design, which is only leveraged in the SynPlay, can utilize \textit{significantly larger range of human motions} compared to detail-guided design.
\end{enumerate}

The comparison shown in the table also demonstrates that SynPlay successfully addresses the shortfall of aerial-view synthetic datasets (3$^\text{rd}$ observation), while maximizing the benefits of aerial-view datasets (1$^\text{st}$ and 2$^\text{nd}$ observations). 

Moreover, the 4$^\text{th}$ observation supports that our proposed rule-guided design is successful in securing the diversity of human motions in the set. It is noteworthy that while SURREAL~\citep{GVarolCVPR2017} (constructed with `detail+mocap') contains a comparable number (6.5M) of human instances as SynPlay, the number of motions manifested in the dataset is extremely limited when compared to SynPlay (23 vs. uncountable).
\section{SynPlay Statistics}
\label{sec:statistics}

\begin{figure*}[t]
\centering
\includegraphics[trim=5mm 5mm 5mm 5mm,clip,width=\linewidth]{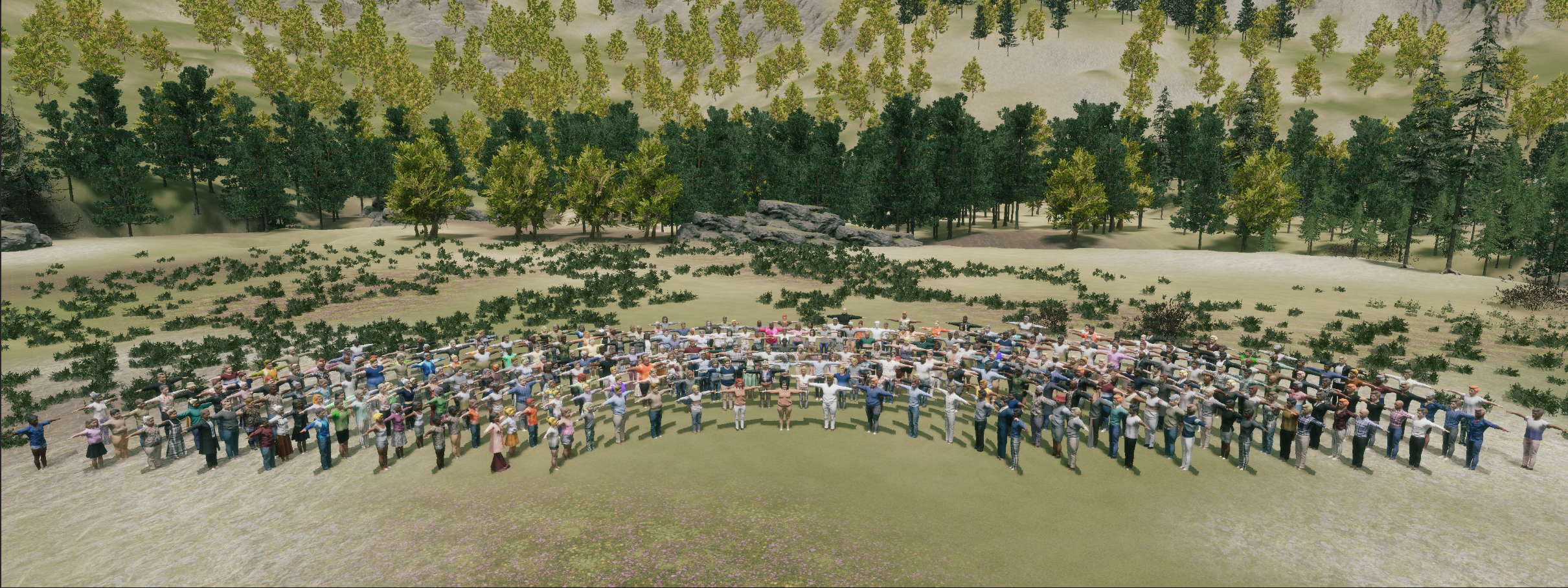}
\caption{{\bf 456 virtual players in SynPlay} created using Character Creator.}
\label{fig:virtual_character}
\end{figure*}

\begin{figure*}
\centering
\begin{minipage}{.45\textwidth}
  \centering
  \includegraphics[width=\linewidth]{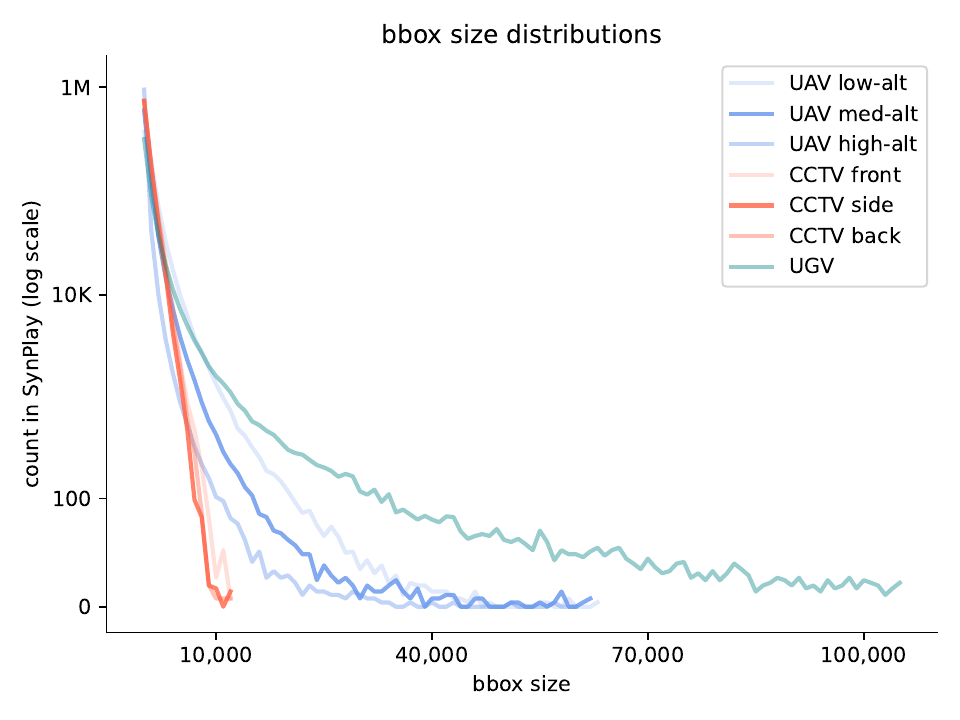}
  \captionof{figure}{{\bf Bounding box size distribution} for each image-capturing device.}
  \label{fig:bbox_size}
\end{minipage}%
~~~~
\begin{minipage}{.45\textwidth}
  \centering
  \includegraphics[width=\linewidth]{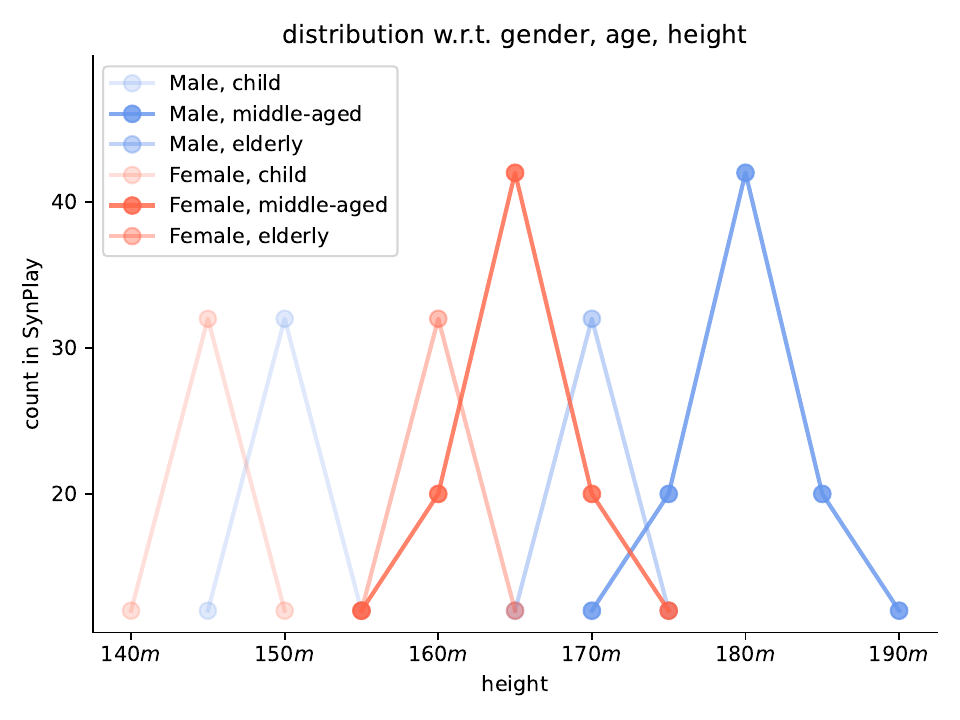}
  \captionof{figure}{{\bf Character height distribution} that varies according to gender and age.}
  \label{fig:height_distribution}
\end{minipage}
\end{figure*}

Here, we provide several statistics from the SynPlay. Fig~\ref{fig:bbox_size} shows the distribution of bounding box sizes over human instances captured by each device. The majority of bounding box sizes are small, which illustrates a characteristic of aerial-view datasets. Interestingly, UAVs can capture human instances with larger bounding boxes than CCTVs. This could be due to the fact that, although UAVs are typically positioned at higher altitudes than CCTVs, there are more cases where UAVs get closer to real-time events and human instances, different from the fixed CCTVs.

Fig~\ref{fig:height_distribution} shows the distribution of human height with respect to gender and age. As mentioned in the main manuscript, each distribution is formed as being bell-shaped. We create 456 virtual characters by controlling human height, gender, and age according to these distributions and uniquely diversifying other factors (skin color, obesity, hair, outfit, \textit{etc}) as much as possible, as shown in Fig~\ref{fig:virtual_character}.

\section{Implementation Details}
\label{sec:implementation_detail}

\subsection{Motion evolution graph}
\label{ssec:motion_evoluton}

\begin{figure*}[t]
\centering
\includegraphics[trim=5mm 5mm 5mm 5mm,clip,width=\linewidth]{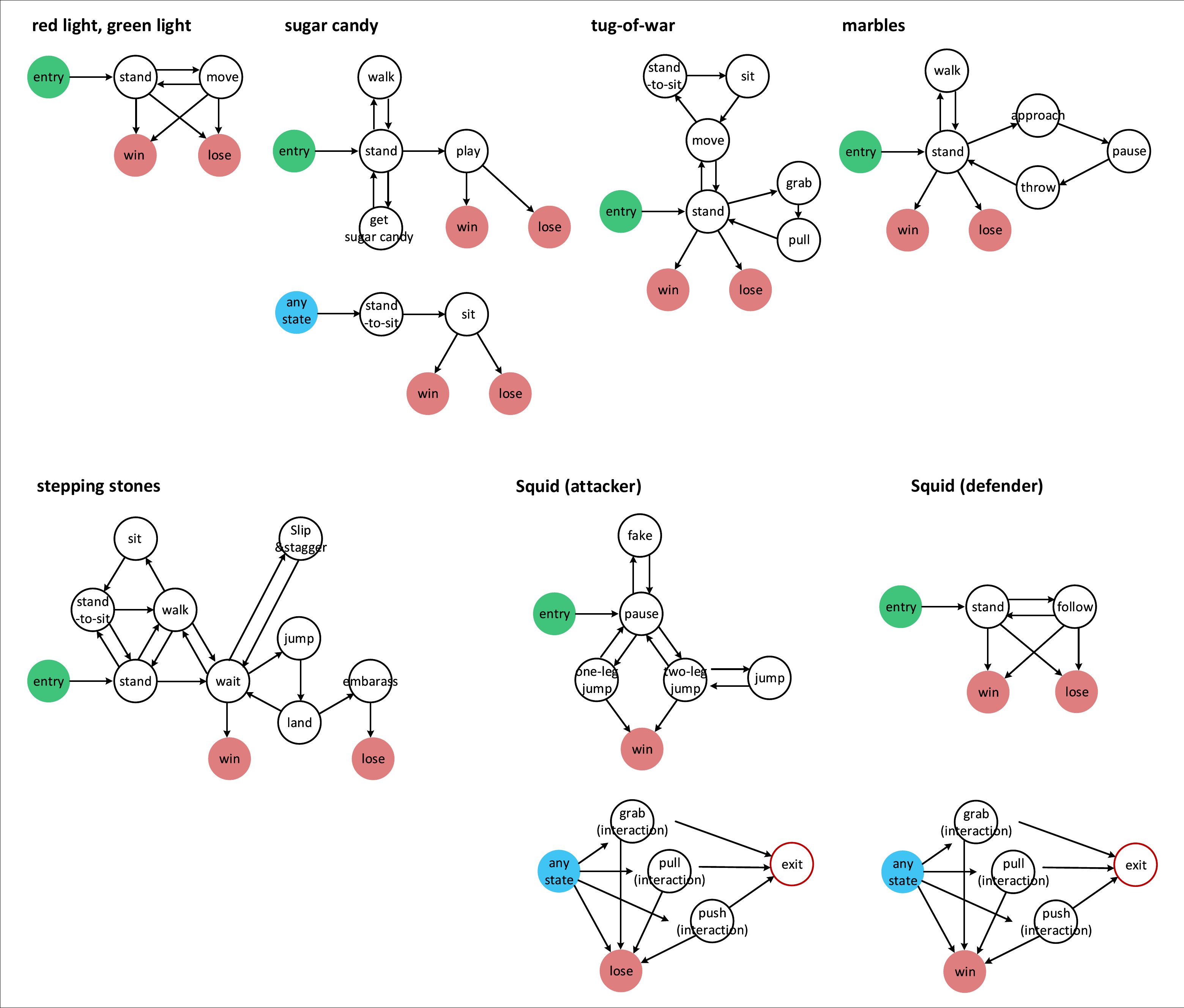}
\caption{{\bf Motion evolution graphs.} The start node (`entry') and the end nodes (`win' or `lose') are indicated by green and red circles, respectively. For the games where secondary graphs are available (i.e., sugar candy or squid), at any given time (except at the start or end node), the current state in the main graph can move to the `any state' node (blue-filled circle) in the secondary graph. When the `end' node (red-bordered circle) is reached within the secondary graph, the current state moves its way back to the latest node that was touched in the main graph before entering the secondary graph.}
\label{fig:motion_evolution_graph}
\end{figure*}

Fig~\ref{fig:motion_evolution_graph} shows motion evolution graphs used in designing the game scenarios for the SynPlay dataset. Even within the same game, the scenario may change, but the motion evolution graph will remain consistent. It is noteworthy to mention that, despite the wide range of situations and a variety of motions involved in the games, the motion evolution graph for each game consists of only a few motion nodes and their transitions. Given that each node (represented as an \textit{elementary motion state} in the main manuscript) encompasses a range of motions, this illustrates the essence of a rule-based design approach where only basic game rules are provided to freely allow the diverse array of human motions to be manifested.

\subsection{Experimental setting}
\label{ssec:training_spec}

In our experiments, our goal is to explore the capabilities of SynPlay as supplementary training data on a variety of tasks. We mostly adhere to the original settings and implementations of the methods used in our experiments, with minimal modifications. The specific modifications used in our experiments are described below.\smallskip

\noindent{\bf Architecture modification.} Our tasks, specifically human detection and semantic segmentation, can be viewed as one-class problems. Therefore, all method architectures, particularly the dimensions of the last layer, have been adjusted accordingly.\smallskip

\noindent{\bf Image size applied in YOLOv8 training/inference.} We use the image size of 1280$\times$1280 for all datasets except for COCO, which uses an image size of 640$\times$640. This decision simply takes into account the original image size of the datasets. Even after rescaling, the size range of human instances in the compared datasets remains similar. When using other models, \textit{i.e.,} Mask2Former in semantic segmentation tasks and retinaNet in data-scarce tasks, the image size/scaling recommended in the original settings was used.\smallskip

\noindent{\bf Training Mask2Former without the large-scale jittering (LSJ) augmentation~\citep{GGhiasiCVPR2021}.} We did not use the default LSJ augmentation in training the Mask2Former segmentation models solely for performance reasons. In all cases, segmentation accuracy were found to be significantly lower when LSJ augmentation was used. LSJ augmentation, which greatly expands the range of image scaling, may not be suitable for aerial-view detection, which mainly includes small-sized human instances. This performance degradation with LSJ augmentation is also observed in~\cite{KHeCVPR2022}, which is a reputable literature in the field of self-supervised learning.\smallskip

\noindent{\bf Settings for PT-FT.} When using PT-FT in the general tasks, training specifications, including training epochs and learning rate, did not differ between pre-training and fine-tuning. In data-scarce tasks, we follow all the settings of \cite{YShenCVPR2023} as outlined in PTL, while leaving out the progressive component.\smallskip

\noindent{\bf Settings for data-scarce tasks.} For all experiments performed for data-scarce tasks including the scaling behavior study, we follow all the settings and experimental environments of \cite{YShenArXiv2024}.\smallskip

\subsection{Quantitative measures}
\label{ssec:quan_measure}

We provide the detail on how we calculate the two metrics used for the quantitative analysis in the main manuscript.\smallskip

\noindent{\bf Fréchet Inception Distance (FID)~\citep{MHeuselNeurIPS2017}.} We utilized the PyTorch implementation of FID in~\cite{Seitzer2020FID} with the default setup to assess the fidelity and diversity for all the training datasets involved in our experiments. We did not perform image scaling on the input for any dataset, and the final average pooling features were used to compute FID.\smallskip

\noindent{\bf Proportion of nadir-view instances.} An instance with an elevation angle greater than 71.57$^\circ$ relative to the UAV is considered to be a nadir-view instance, representing the maximum elevation angle for Archangel*~\citep{YShenArXiv2024}. To identify nadir-view instances for Archangel, we utilized the dataset metadata, i.e., UAV position. Similarly, for Archangel*, we determined if an instance was an nadir-view instance based on the source instance, also using the dataset metadata. In the case of SynPlay, we computed the elevation angle for each instance using the absolute 3D coordinates of the instance and the UAV provided by SynPlay.
\section{Qualitative Analysis}
\label{sec:qualitative_analysis}

\subsection{Blending and animation layer}
\label{ssec:two_virtual_technique}

\begin{figure*}[t]
\centering
\setlength{\tabcolsep}{0.5pt}
\begin{tabular}{ccccccccc}
\multicolumn{4}{c}{{\bf crawling}} & & \multicolumn{4}{c}{{\bf running}} \\
\includegraphics[trim=10mm 0mm 10mm 0mm, width=.12\linewidth]{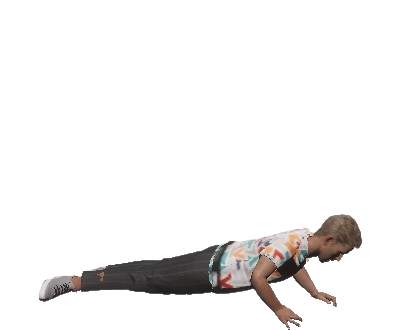} &
\includegraphics[trim=10mm 0mm 10mm 0mm, width=.12\linewidth]{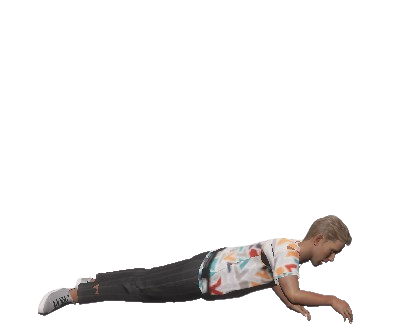} &
\includegraphics[trim=10mm 0mm 10mm 0mm, width=.12\linewidth]{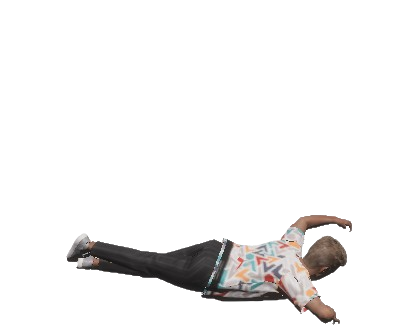} &
\includegraphics[trim=10mm 0mm 10mm 0mm, width=.12\linewidth]{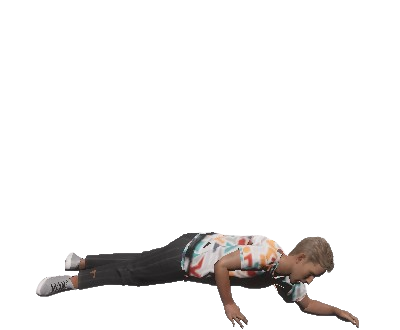} &~~~~~~~
~~~~~~~&
\includegraphics[width=.09\linewidth]{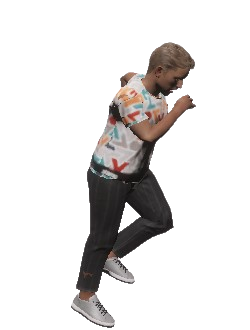} &
\includegraphics[width=.09\linewidth]{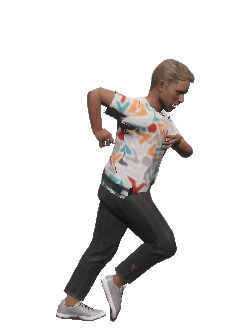} &
\includegraphics[width=.09\linewidth]{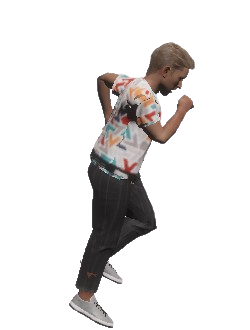} &
\includegraphics[width=.09\linewidth]{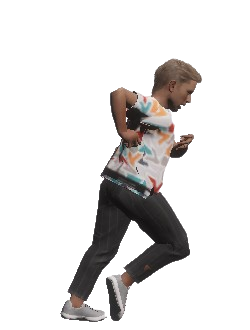} \\\\
\multicolumn{4}{c}{{\bf crippling}} & & \multicolumn{4}{c}{{\bf dragging}} \\
\includegraphics[width=.07\linewidth]{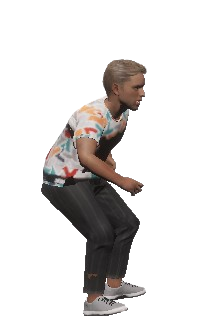} &
\includegraphics[width=.07\linewidth]{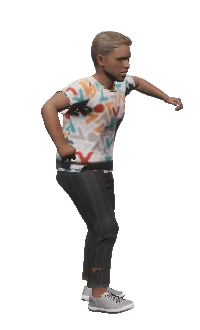} &
\includegraphics[width=.07\linewidth]{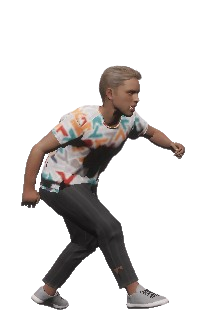} &
\includegraphics[width=.07\linewidth]{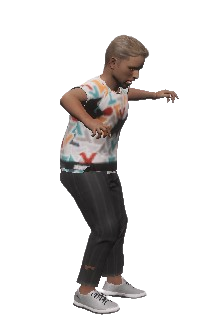} &
&
\includegraphics[width=.09\linewidth]{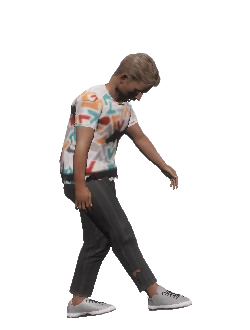} &
\includegraphics[width=.09\linewidth]{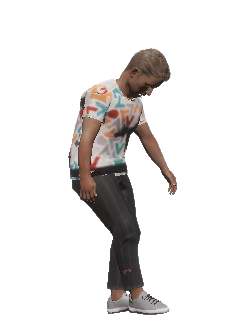} &
\includegraphics[width=.09\linewidth]{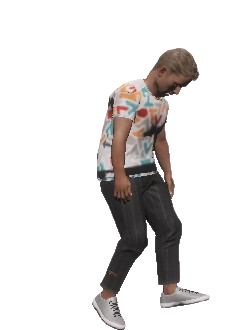} &
\includegraphics[width=.09\linewidth]{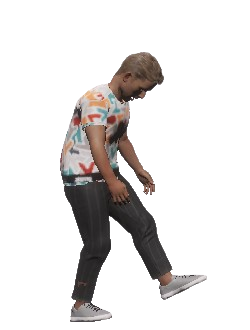} \\\\\cdashline{1-4}\cdashline{6-9}\\
\multicolumn{4}{c}{{\bf walking while bending}} & & \multicolumn{4}{c}{{\bf running of exhausted person}} \\
\includegraphics[width=.09\linewidth]{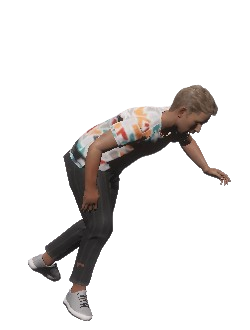} &
\includegraphics[width=.09\linewidth]{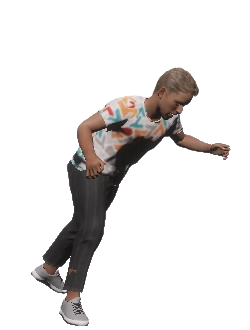} &
\includegraphics[width=.09\linewidth]{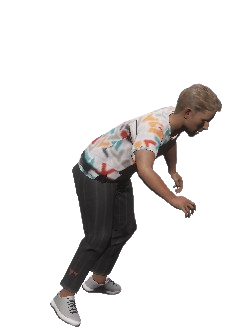} &
\includegraphics[width=.09\linewidth]{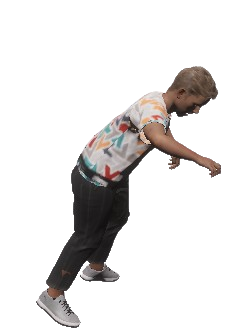} &
&
\includegraphics[width=.09\linewidth]{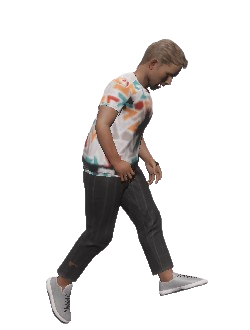} &
\includegraphics[width=.09\linewidth]{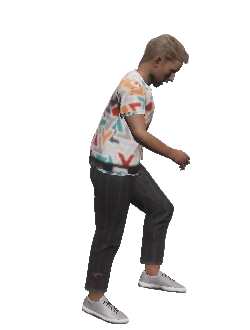} &
\includegraphics[width=.09\linewidth]{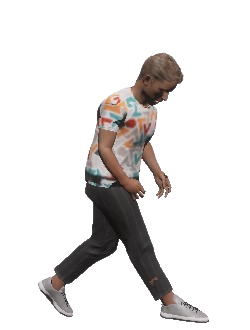} &
\includegraphics[width=.09\linewidth]{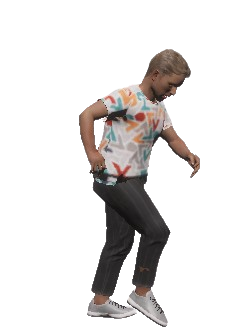} \\
\end{tabular}
\caption{{\bf Two motion blending examples}. For each example (left or right column), the two motions (top and middle row) is blended together to generate a new motion (bottom row). The blending ratio between the two input motions can be controlled. The names for the motions are not computationally involved in the blending process.}
\label{fig:motion_blending}
\end{figure*}
\begin{figure*}[t]
\centering
\setlength{\tabcolsep}{0.5pt}
\begin{tabular}{cccccccccccccc}
\multicolumn{4}{c}{{\bf sitting}} & & \multicolumn{4}{c}{{\bf walking}} & & \multicolumn{4}{c}{{\bf kneeling down}} \\
\includegraphics[trim=10mm 0mm 10mm 0mm, width=.07\linewidth]{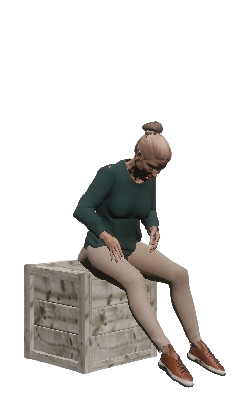} &
\includegraphics[trim=10mm 0mm 10mm 0mm, width=.07\linewidth]{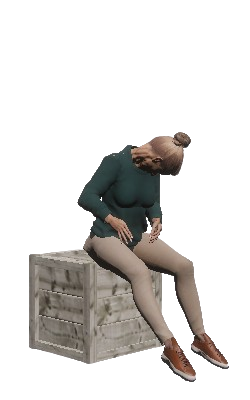} &
\includegraphics[trim=10mm 0mm 10mm 0mm, width=.07\linewidth]{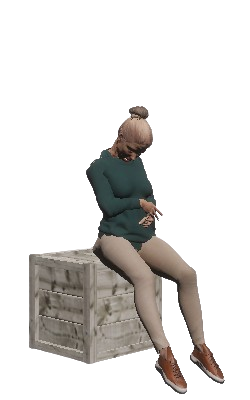} &
\includegraphics[trim=10mm 0mm 10mm 0mm, width=.07\linewidth]{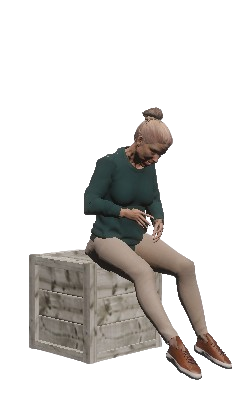} &~~~~
~~~~&
\includegraphics[trim=10mm 0mm 10mm 0mm, width=.07\linewidth]{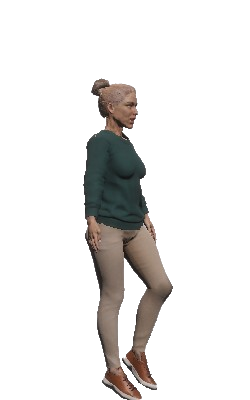} &
\includegraphics[trim=10mm 0mm 10mm 0mm, width=.07\linewidth]{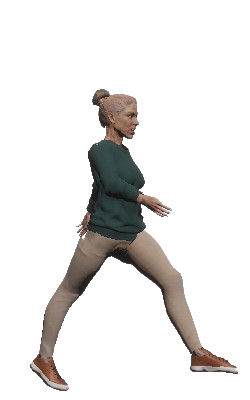} &
\includegraphics[trim=10mm 0mm 10mm 0mm, width=.07\linewidth]{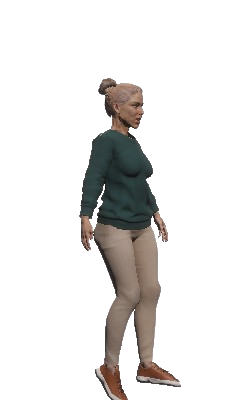} &
\includegraphics[trim=10mm 0mm 10mm 0mm, width=.07\linewidth]{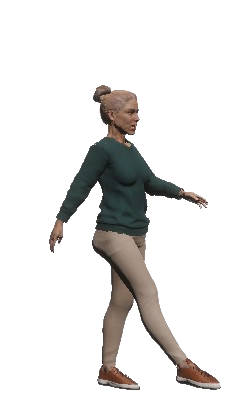} &~~~~
~~~~&
\includegraphics[trim=10mm 0mm 10mm 0mm, width=.07\linewidth]{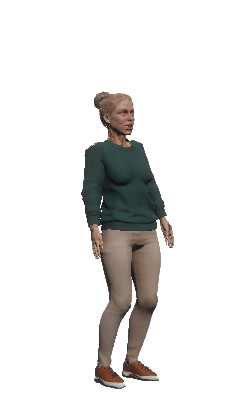} &
\includegraphics[trim=10mm 0mm 10mm 0mm, width=.07\linewidth]{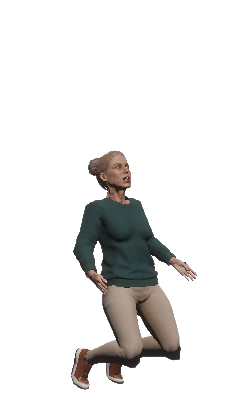} &
\includegraphics[trim=10mm 0mm 10mm 0mm, width=.07\linewidth]{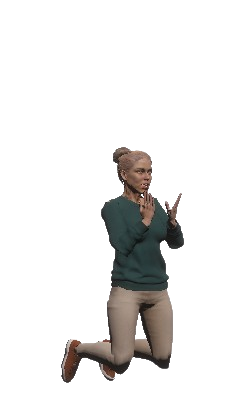} &
\includegraphics[trim=10mm 0mm 10mm 0mm, width=.07\linewidth]{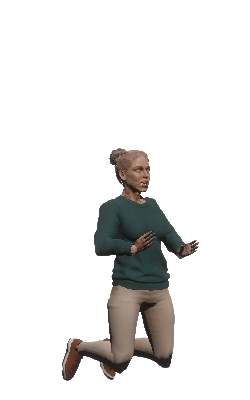}\\\\
\multicolumn{4}{c}{{\bf raising something up}} & & \multicolumn{4}{c}{\multirow{2}{*}{{\bf throwing something}}} & & \multicolumn{4}{c}{\multirow{2}{*}{{\bf cheering}}} \\
\multicolumn{4}{c}{{\bf and looking at it}} & &  \\
\includegraphics[trim=10mm 0mm 10mm 0mm, width=.07\linewidth]{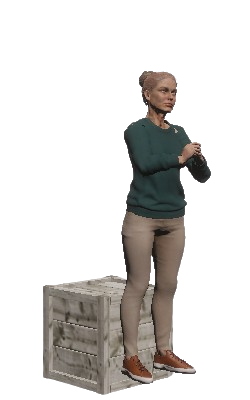} &
\includegraphics[trim=10mm 0mm 10mm 0mm, width=.07\linewidth]{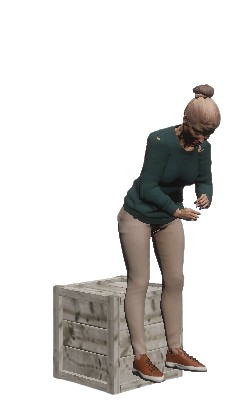} &
\includegraphics[trim=10mm 0mm 10mm 0mm, width=.07\linewidth]{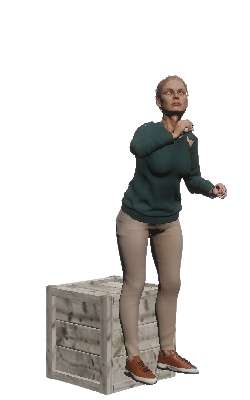} &
\includegraphics[trim=10mm 0mm 10mm 0mm, width=.07\linewidth]{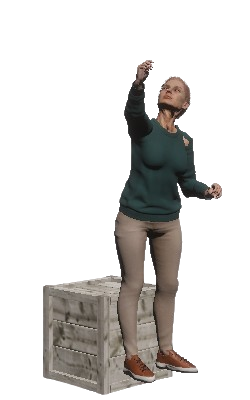} &~~~
~~~&
\includegraphics[trim=10mm 0mm 10mm 0mm, width=.07\linewidth]{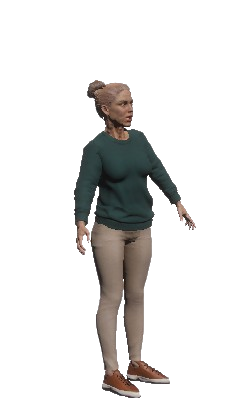} &
\includegraphics[trim=10mm 0mm 10mm 0mm, width=.07\linewidth]{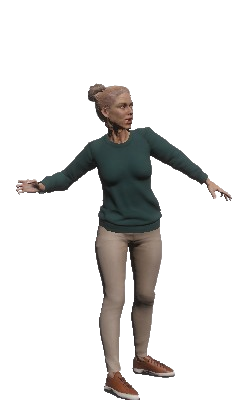} &
\includegraphics[trim=10mm 0mm 10mm 0mm, width=.07\linewidth]{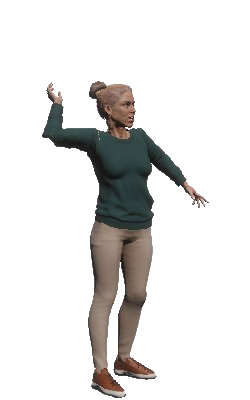} &
\includegraphics[trim=10mm 0mm 10mm 0mm, width=.07\linewidth]{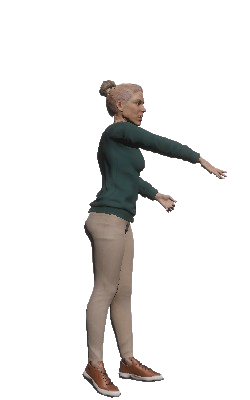} &~~~
~~~&
\includegraphics[trim=10mm 0mm 10mm 0mm, width=.07\linewidth]{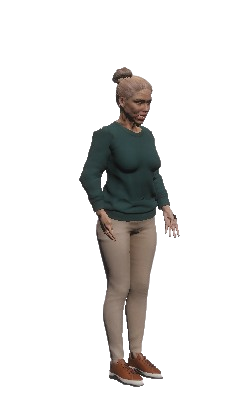} &
\includegraphics[trim=10mm 0mm 10mm 0mm, width=.07\linewidth]{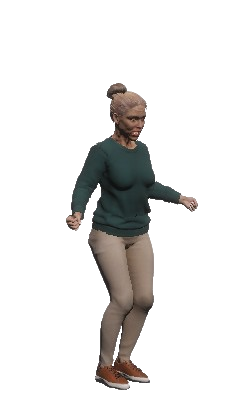} &
\includegraphics[trim=10mm 0mm 10mm 0mm, width=.07\linewidth]{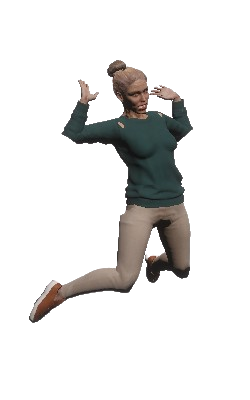} &
\includegraphics[trim=10mm 0mm 10mm 0mm, width=.07\linewidth]{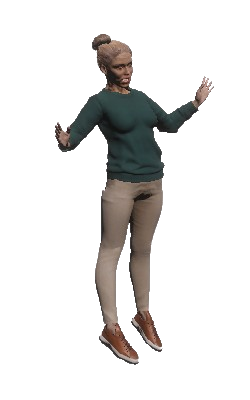} \\\\\cdashline{1-4}\cdashline{6-9}\cdashline{11-14}\\
\multicolumn{4}{c}{{\bf sitting, raising something up}} & & \multicolumn{4}{c}{{\bf throwing something}} & & \multicolumn{4}{c}{{\bf cheering}} \\
\multicolumn{4}{c}{{\bf and looking at it}} & & \multicolumn{4}{c}{{\bf while walking}} & & \multicolumn{4}{c}{{\bf while kneeling down}} \\
\includegraphics[trim=10mm 0mm 10mm 0mm, width=.07\linewidth]{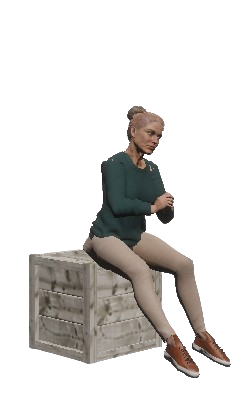} &
\includegraphics[trim=10mm 0mm 10mm 0mm, width=.07\linewidth]{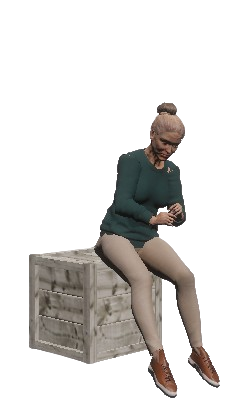} &
\includegraphics[trim=10mm 0mm 10mm 0mm, width=.07\linewidth]{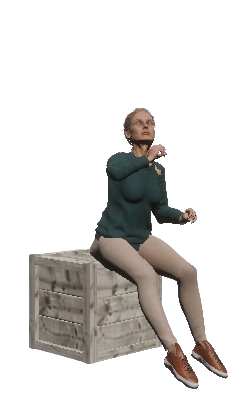} &
\includegraphics[trim=10mm 0mm 10mm 0mm, width=.07\linewidth]{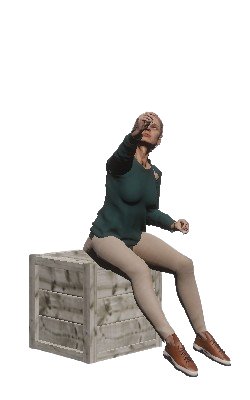} &~~~
~~~&
\includegraphics[trim=10mm 0mm 10mm 0mm, width=.07\linewidth]{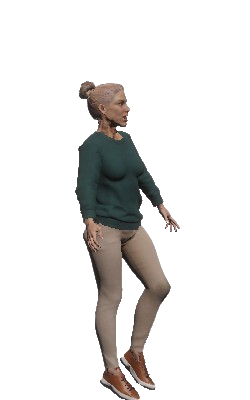} &
\includegraphics[trim=10mm 0mm 10mm 0mm, width=.07\linewidth]{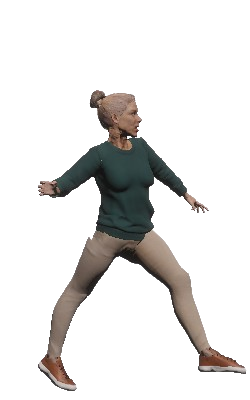} &
\includegraphics[trim=10mm 0mm 10mm 0mm, width=.07\linewidth]{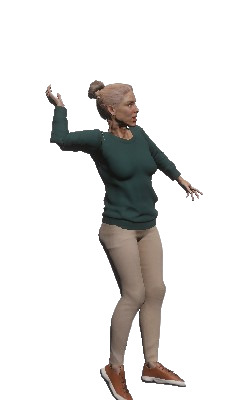} &
\includegraphics[trim=10mm 0mm 10mm 0mm, width=.07\linewidth]{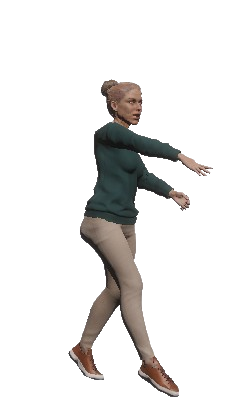} &~~~
~~~&
\includegraphics[trim=10mm 0mm 10mm 0mm, width=.07\linewidth]{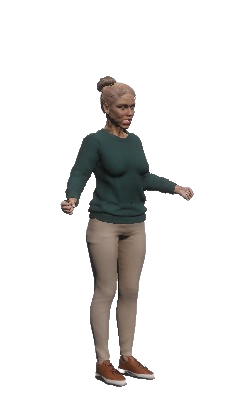} &
\includegraphics[trim=10mm 0mm 10mm 0mm, width=.07\linewidth]{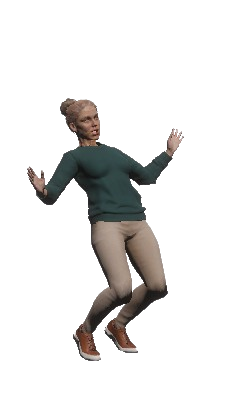} &
\includegraphics[trim=10mm 0mm 10mm 0mm, width=.07\linewidth]{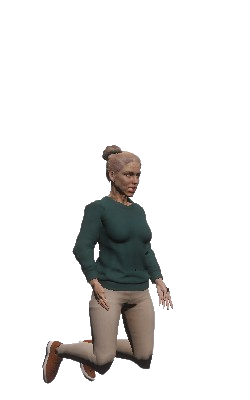} &
\includegraphics[trim=10mm 0mm 10mm 0mm, width=.07\linewidth]{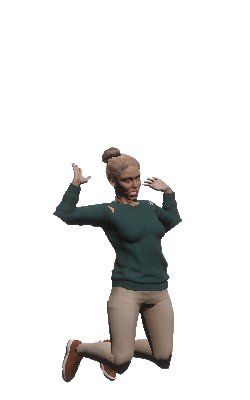} \\
\end{tabular}
\caption{{\bf Three examples of leveraging animation layers}. For each example (left, middle, or right), the resulting motion of leveraging the animation layers over two input motions (top and middle rows) is shown in the bottom row. Note that the semantic labels (e.g., walking, cheering) were not provided at the time of capture; they are included in the figure only for the convenience of the readers.}
\label{fig:animation_layer}
\end{figure*}

Fig~\ref{fig:motion_blending} and~\ref{fig:animation_layer} show several examples of the blending process and how the animation layers are leveraged: the two techniques for expanding human motions within the virtual environments, respectively. Interestingly, the motions created by blending is largely different from their corresponding input motions, while the motions created via leveraging the animation layers still exhibit the appearances and dynamics resembling both the input motions. These two techniques are readily available for use within the Unity environment.

\subsection{Virtual motion and real-world motion}
\label{ssec:motion_comparison}

\begin{figure*}[t]

\centering
\setlength{\tabcolsep}{0.5pt}
\begin{tabular}{ccccc:cccccccccc}
\multicolumn{14}{c}{(a) Using virtual motions as reference}\\\\
\multicolumn{4}{c}{{\bf Virtual motion}} & \multicolumn{2}{c}{} & \multicolumn{9}{c}{{\bf Real-world motion}} \\\\
\includegraphics[trim=10mm 0mm 10mm 0mm, width=.05\linewidth]{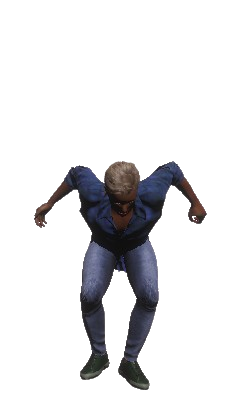} &
\includegraphics[trim=10mm 0mm 10mm 0mm, width=.05\linewidth]{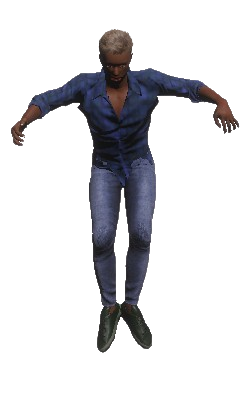} &
\includegraphics[trim=10mm 0mm 10mm 0mm, width=.05\linewidth]{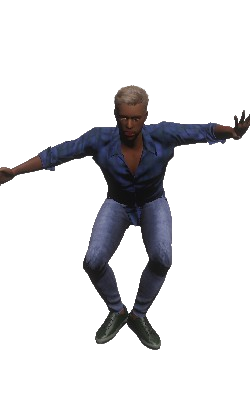} &
\includegraphics[trim=10mm 0mm 10mm 0mm, width=.05\linewidth]{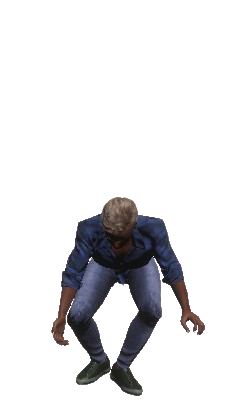} &~~~~~~~
& ~~~~~~~&
\includegraphics[trim=10mm 0mm 10mm 0mm, width=.05\linewidth]{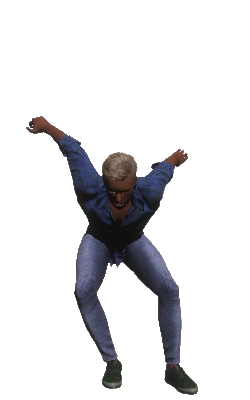} &
\includegraphics[trim=10mm 0mm 10mm 0mm, width=.05\linewidth]{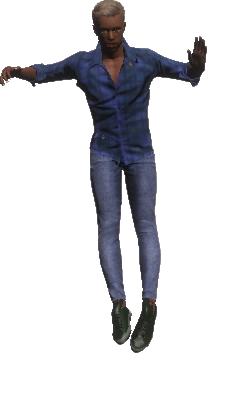} &
\includegraphics[trim=10mm 0mm 10mm 0mm, width=.05\linewidth]{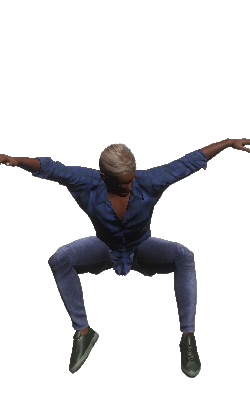} &
\includegraphics[trim=10mm 0mm 10mm 0mm, width=.05\linewidth]{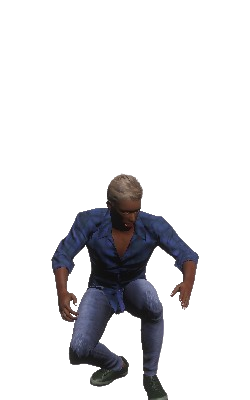} &~~~
~~~&
\includegraphics[trim=10mm 0mm 10mm 0mm, width=.05\linewidth]{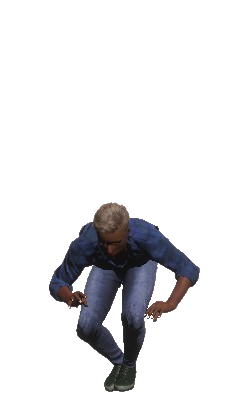} &
\includegraphics[trim=10mm 0mm 10mm 0mm, width=.05\linewidth]{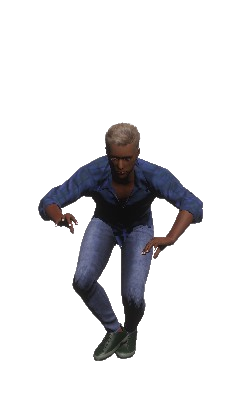} &
\includegraphics[trim=10mm 0mm 10mm 0mm, width=.05\linewidth]{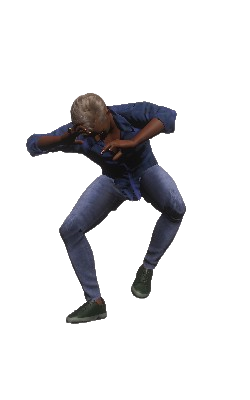} &
\includegraphics[trim=10mm 0mm 10mm 0mm, width=.05\linewidth]{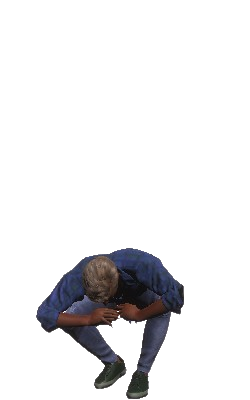} \\
\multicolumn{4}{c}{{\bf jumping}} & & \\
&&&&&&
\includegraphics[trim=10mm 0mm 10mm 30mm, width=.05\linewidth]{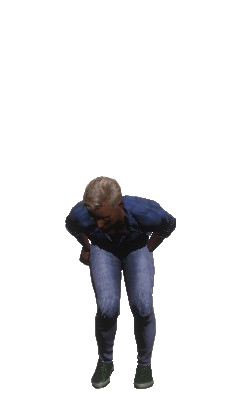} &
\includegraphics[trim=10mm 0mm 10mm 30mm, width=.05\linewidth]{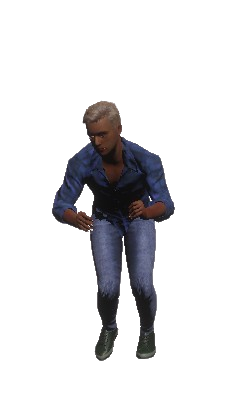} &
\includegraphics[trim=10mm 0mm 10mm 30mm, width=.05\linewidth]{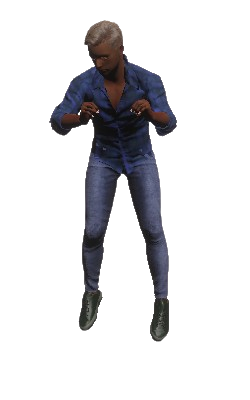} &
\includegraphics[trim=10mm 0mm 10mm 30mm, width=.05\linewidth]{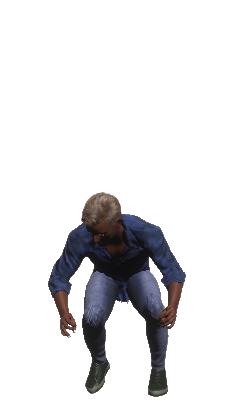} &~~~
~~~&&&& \\\\
\includegraphics[trim=10mm 0mm 10mm 0mm, width=.05\linewidth]{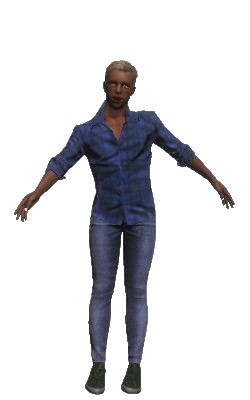} &
\includegraphics[trim=10mm 0mm 10mm 0mm, width=.05\linewidth]{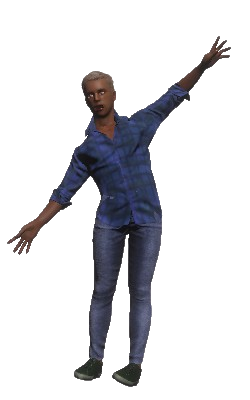} &
\includegraphics[trim=10mm 0mm 10mm 0mm, width=.05\linewidth]{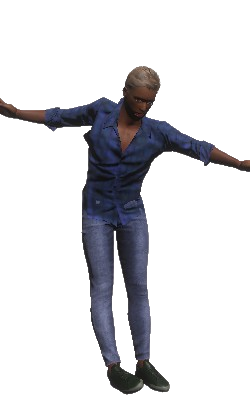} &
\includegraphics[trim=10mm 0mm 10mm 0mm, width=.05\linewidth]{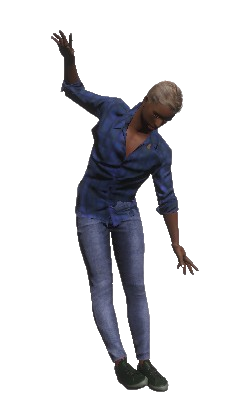} &~~~~~~~
& ~~~~~~~&
\includegraphics[trim=10mm 0mm 10mm 0mm, width=.05\linewidth]{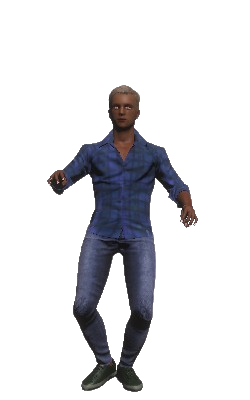} &
\includegraphics[trim=10mm 0mm 10mm 0mm, width=.05\linewidth]{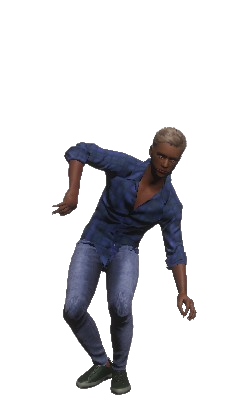} &
\includegraphics[trim=10mm 0mm 10mm 0mm, width=.05\linewidth]{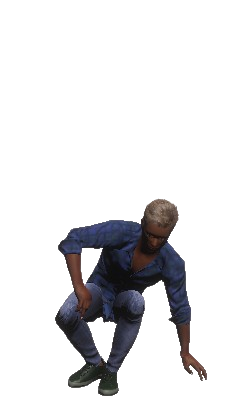} &
\includegraphics[trim=10mm 0mm 10mm 0mm, width=.05\linewidth]{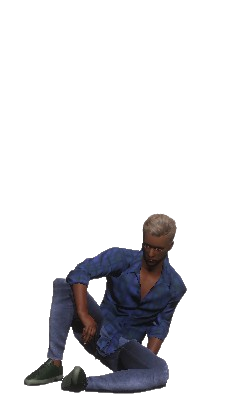} &~~~
~~~&
\includegraphics[trim=10mm 0mm 10mm 0mm, width=.05\linewidth]{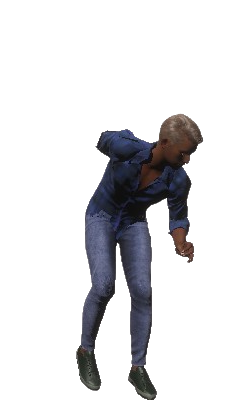} &
\includegraphics[trim=10mm 0mm 10mm 0mm, width=.05\linewidth]{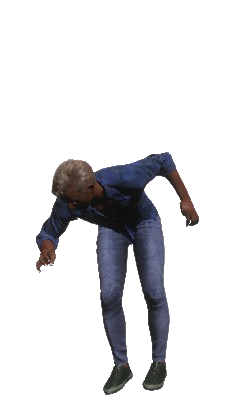} &
\includegraphics[trim=10mm 0mm 10mm 0mm, width=.05\linewidth]{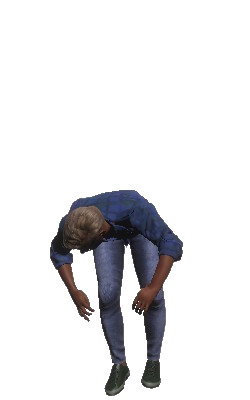} &
\includegraphics[trim=10mm 0mm 10mm 0mm, width=.05\linewidth]{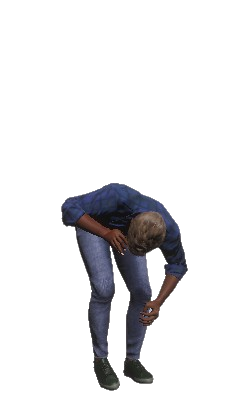} \\
\multicolumn{4}{c}{{\bf slipping}} & & \\
&&&&&&
\includegraphics[trim=10mm 0mm 10mm 40mm, width=.05\linewidth]{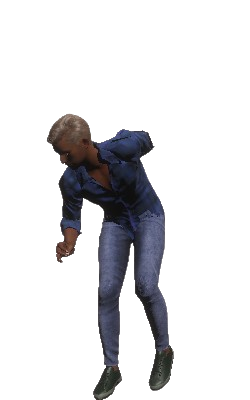} &
\includegraphics[trim=10mm 0mm 10mm 40mm, width=.05\linewidth]{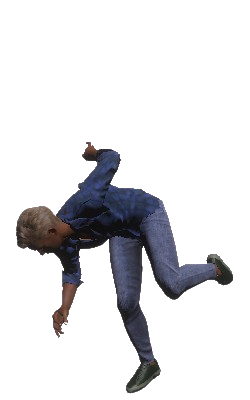} &
\includegraphics[trim=10mm 0mm 10mm 40mm, width=.05\linewidth]{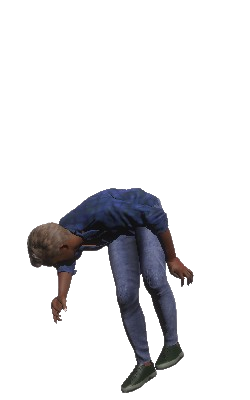} &
\includegraphics[trim=10mm 0mm 10mm 40mm, width=.05\linewidth]{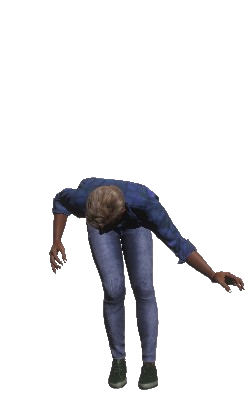} &~~~
~~~&
\includegraphics[trim=10mm 0mm 10mm 40mm, width=.05\linewidth]{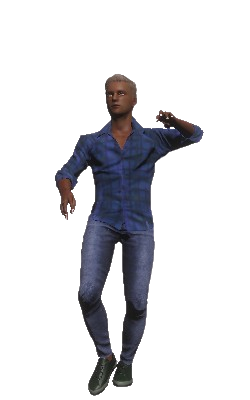} &
\includegraphics[trim=10mm 0mm 10mm 40mm, width=.05\linewidth]{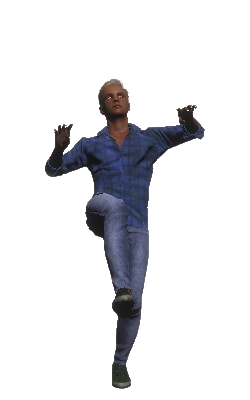} &
\includegraphics[trim=10mm 0mm 10mm 40mm, width=.05\linewidth]{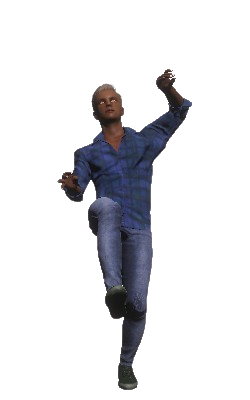} &
\includegraphics[trim=10mm 0mm 10mm 40mm, width=.05\linewidth]{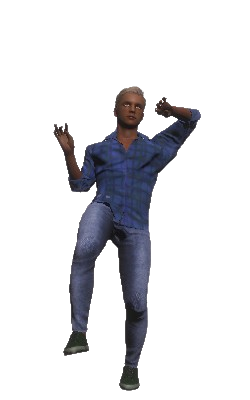} \\\\\\
\end{tabular}

\begin{tabular}{ccccccccc}
\multicolumn{9}{c}{(b) Without reference}\\
\includegraphics[trim=10mm 0mm 10mm 10mm, width=.067\linewidth]{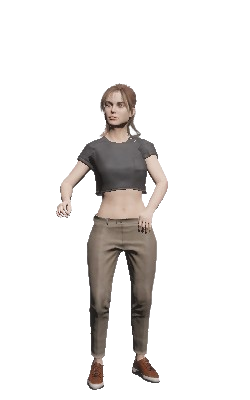} &
\includegraphics[trim=10mm 0mm 10mm 10mm, width=.067\linewidth]{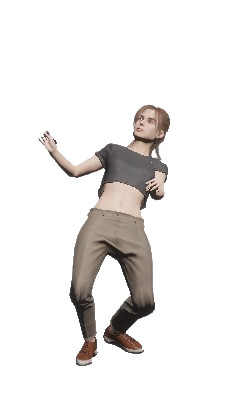} &
\includegraphics[trim=10mm 0mm 10mm 10mm, width=.067\linewidth]{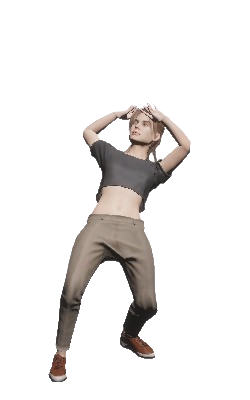} &
\includegraphics[trim=10mm 0mm 10mm 10mm, width=.067\linewidth]{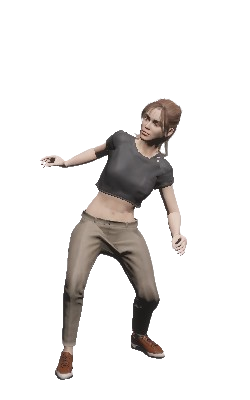} &~~~~~~~ ~~~~~~~&
\includegraphics[trim=10mm 0mm 10mm 10mm, width=.067\linewidth]{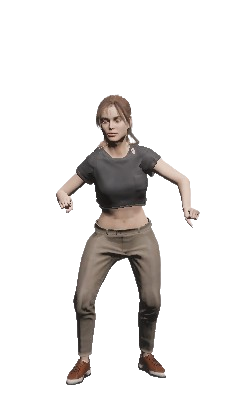} &
\includegraphics[trim=10mm 0mm 10mm 10mm, width=.067\linewidth]{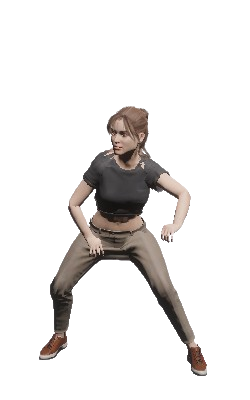} &
\includegraphics[trim=10mm 0mm 10mm 10mm, width=.067\linewidth]{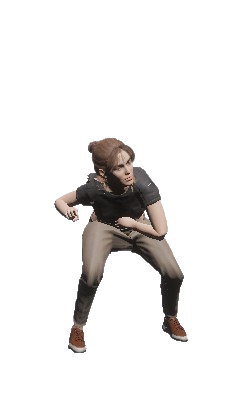} &
\includegraphics[trim=10mm 0mm 10mm 10mm, width=.067\linewidth]{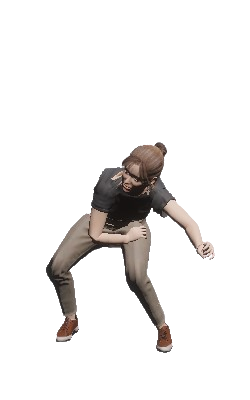} \\
\multicolumn{9}{c}{{\bf dodging}} \\\\
\end{tabular}

\begin{tabular}{ccccccccc}
\includegraphics[trim=10mm 0mm 10mm 0mm, width=.106\linewidth]{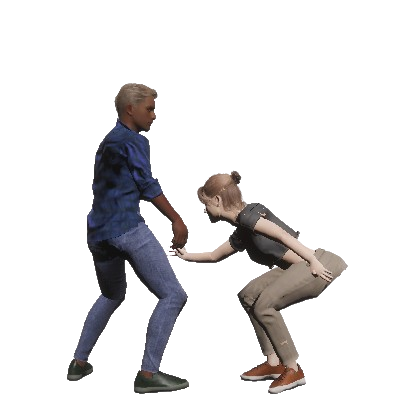} &
\includegraphics[trim=10mm 0mm 10mm 0mm, width=.106\linewidth]{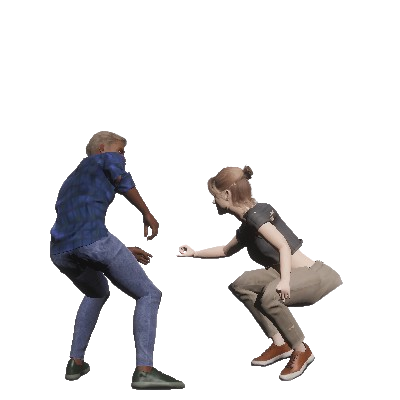} &
\includegraphics[trim=10mm 0mm 10mm 0mm, width=.106\linewidth]{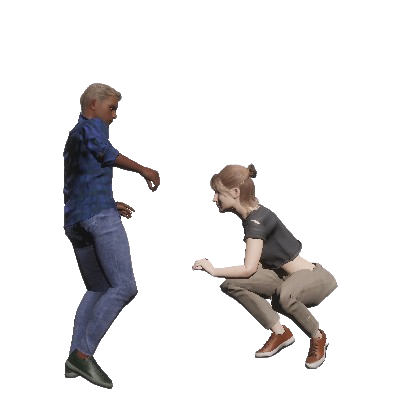} &
\includegraphics[trim=10mm 0mm 10mm 0mm, width=.106\linewidth]{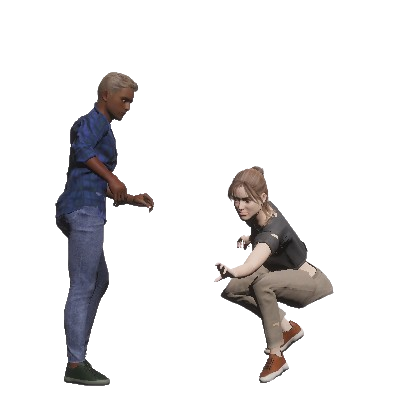} &~~~~~~~ ~~~~~~~&
\includegraphics[trim=10mm 0mm 10mm 0mm, width=.106\linewidth]{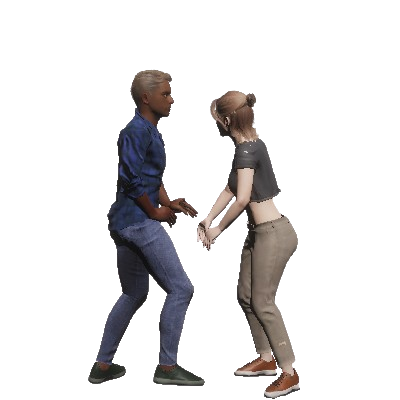} &
\includegraphics[trim=10mm 0mm 10mm 0mm, width=.106\linewidth]{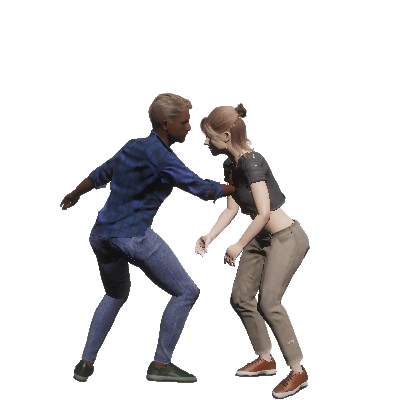} &
\includegraphics[trim=10mm 0mm 10mm 0mm, width=.106\linewidth]{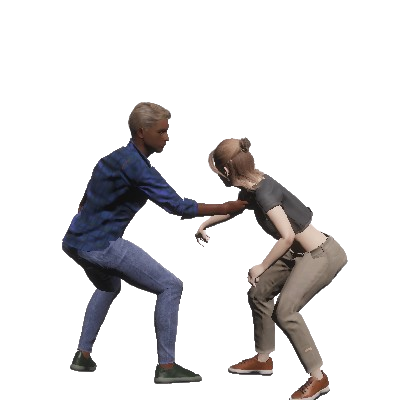} &
\includegraphics[trim=10mm 0mm 10mm 0mm, width=.106\linewidth]{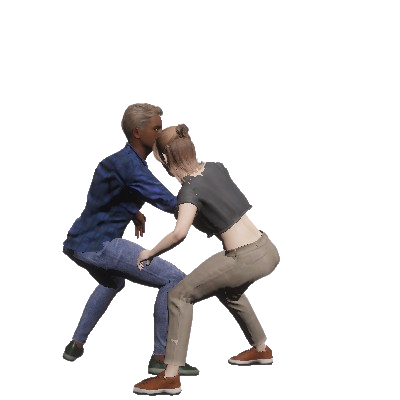} \\
\end{tabular}

\begin{tabular}{ccccccccc}
\includegraphics[trim=10mm 0mm 10mm 20mm, width=.106\linewidth]{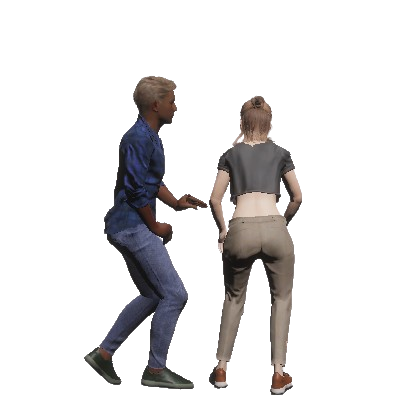} &
\includegraphics[trim=10mm 0mm 10mm 20mm, width=.106\linewidth]{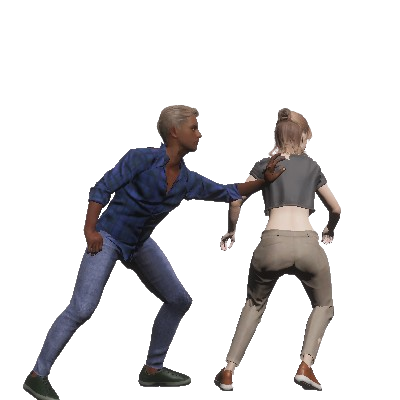} &
\includegraphics[trim=10mm 0mm 0mm 20mm, width=.110\linewidth]{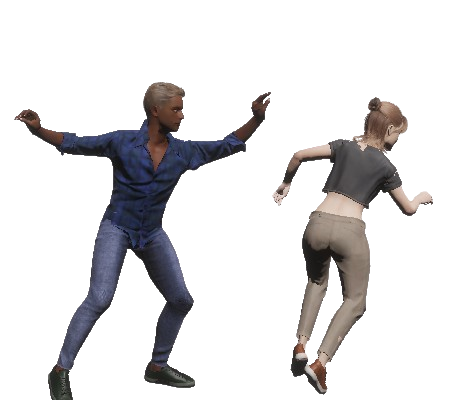} &
\includegraphics[trim=0mm 0mm 10mm 20mm, width=.124\linewidth]{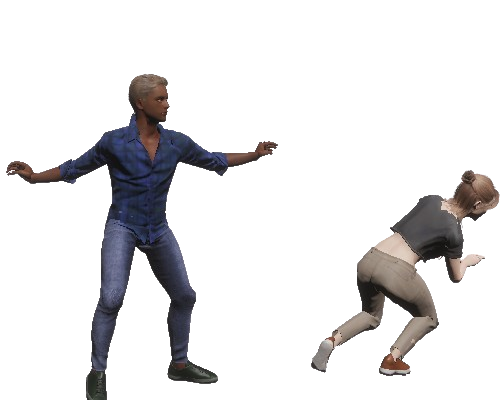} &~~~~~ ~~~~~&
\includegraphics[trim=10mm 0mm 30mm 20mm, width=.128\linewidth]{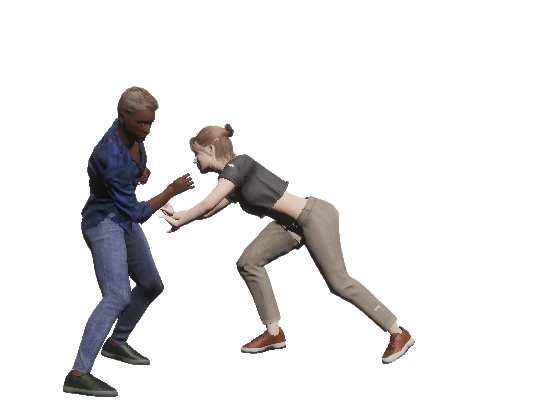} &
\includegraphics[trim=10mm 0mm 30mm 20mm, width=.124\linewidth]{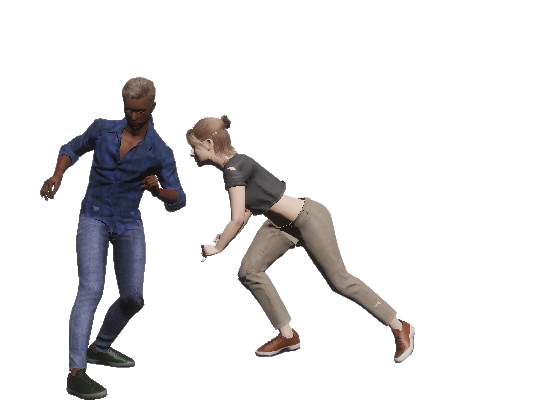} &
\includegraphics[trim=10mm 0mm 30mm 20mm, width=.13\linewidth]{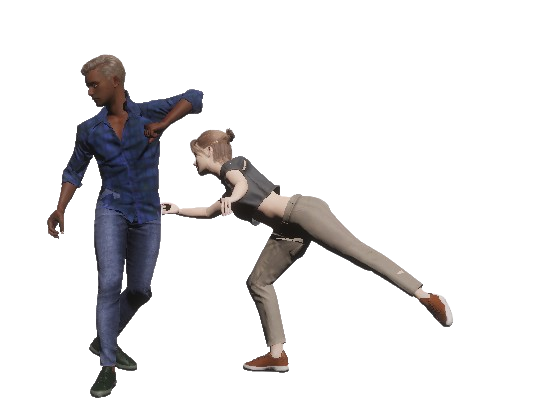} &
\includegraphics[trim=10mm 0mm 30mm 20mm, width=.13\linewidth]{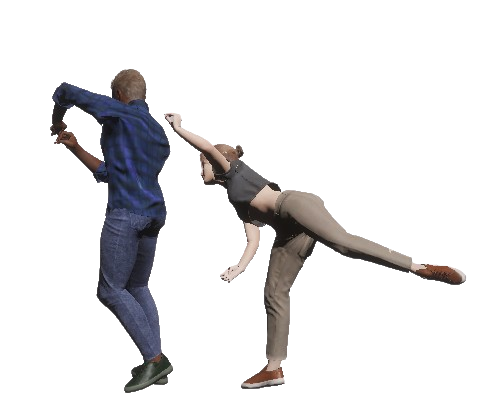} \\
\end{tabular}

\begin{tabular}{ccccccccc}
\includegraphics[trim=10mm 0mm 10mm 20mm, width=.106\linewidth]{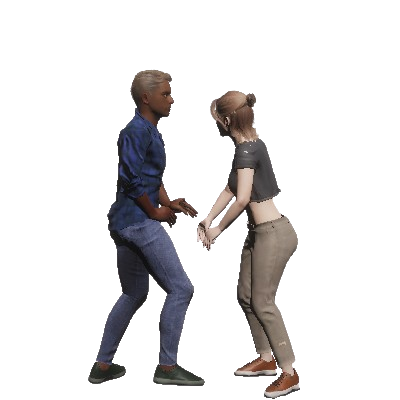} &
\includegraphics[trim=10mm 0mm 10mm 20mm, width=.106\linewidth]{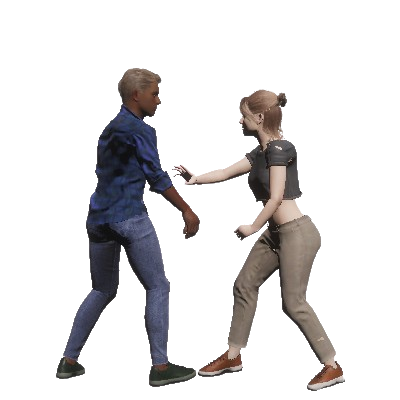} &
\includegraphics[trim=10mm 0mm 10mm 20mm, width=.106\linewidth]{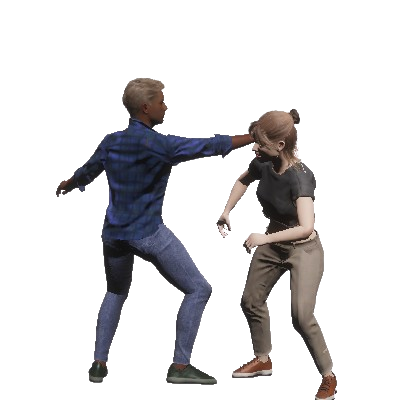} &
\includegraphics[trim=10mm 0mm 10mm 20mm, width=.106\linewidth]{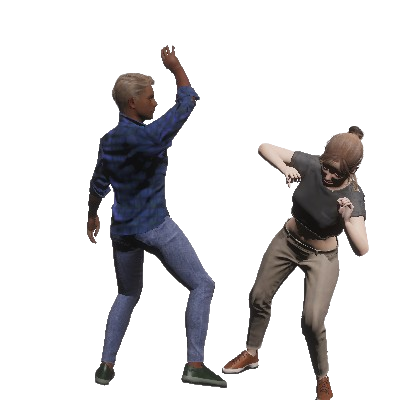} &~~~~~~~ ~~~~~~~&
\includegraphics[trim=10mm 0mm 10mm 20mm, width=.106\linewidth]{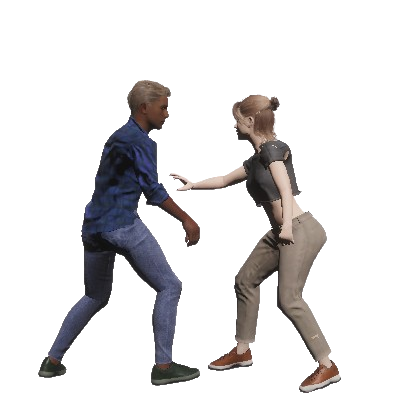} &
\includegraphics[trim=10mm 0mm 10mm 20mm, width=.106\linewidth]{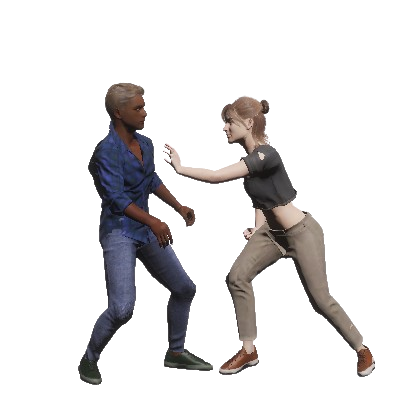} &
\includegraphics[trim=10mm 0mm 10mm 20mm, width=.106\linewidth]{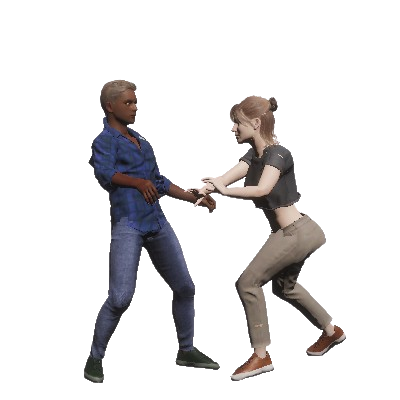} &
\includegraphics[trim=10mm 0mm 10mm 20mm, width=.106\linewidth]{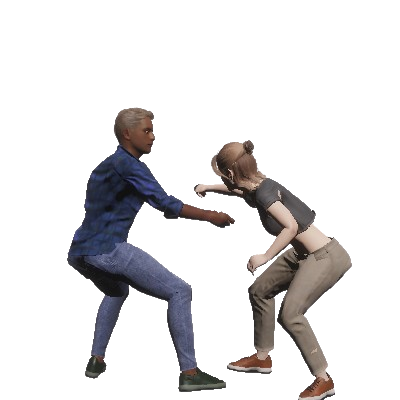} \\
\multicolumn{9}{c}{{\bf wrestling (multi-player motion)}} \\
\end{tabular}

\caption{{\bf Real-world motion examples.} Real-world motions are acquired either (a) by mimicking reference motions or (b) by exhibiting potential in-game motions without any reference that align with the given game rules. Wearable motion scanners are used for all the cases. Note that the semantic labels (e.g., jumping, dodging) were not provided at the time of capture; they are included in the figure only for the convenience of the readers. }
\label{fig:real-world_motions}
\end{figure*}

Fig~\ref{fig:real-world_motions} shows several examples of real-world motions. Real-world motions are created either by having the real human wearing the motion capture device mimic the pre-provided reference motions or by demonstrating potential in-game motions under the given game rules. It is observed that real-world motions can express a wider range of specific actions while maintaining a sense of realism. Moreover, motions that are difficult to pinpoint or describe can also be created, \textit{e.g.,} multi-person wrestling motions.

\subsection{SynPlay sample images}
\label{ssec:sample_images}

\begin{figure*}[h]
\centerline{
\setlength{\tabcolsep}{0.5pt}
\begin{tabular}{ccc}
\multicolumn{3}{c}{(1) red light, green light}\vspace{0.2cm}\\
\includegraphics[width=.32\linewidth]{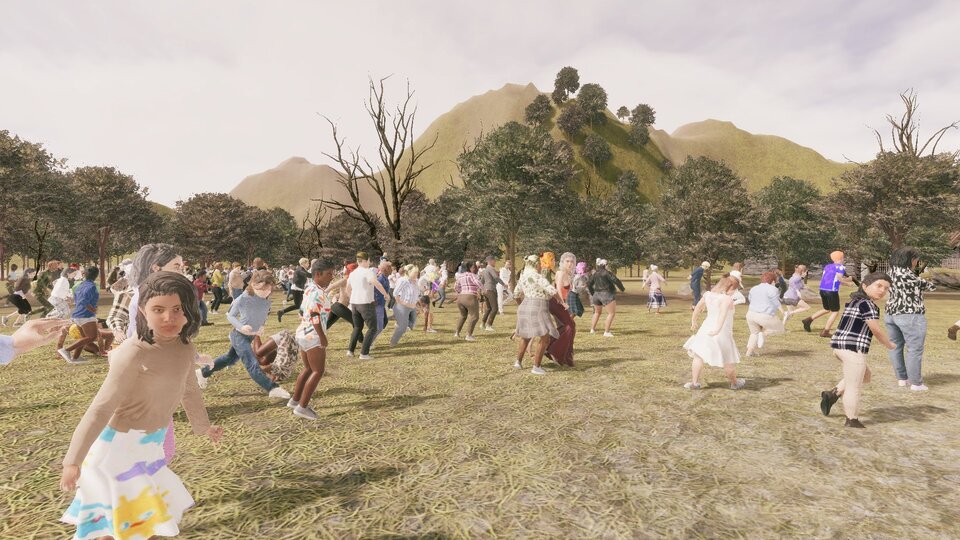} ~&~
\includegraphics[width=.32\linewidth]{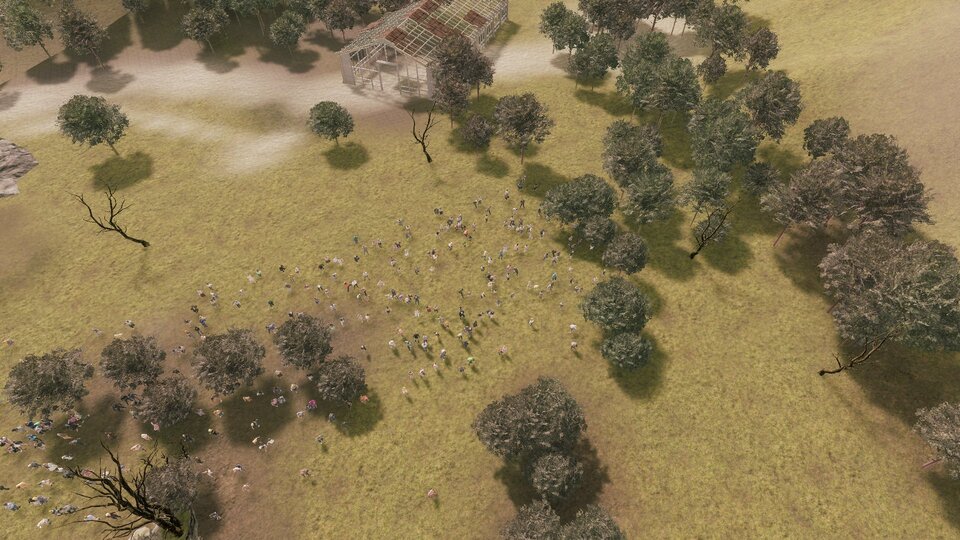} ~&~
\includegraphics[width=.32\linewidth]{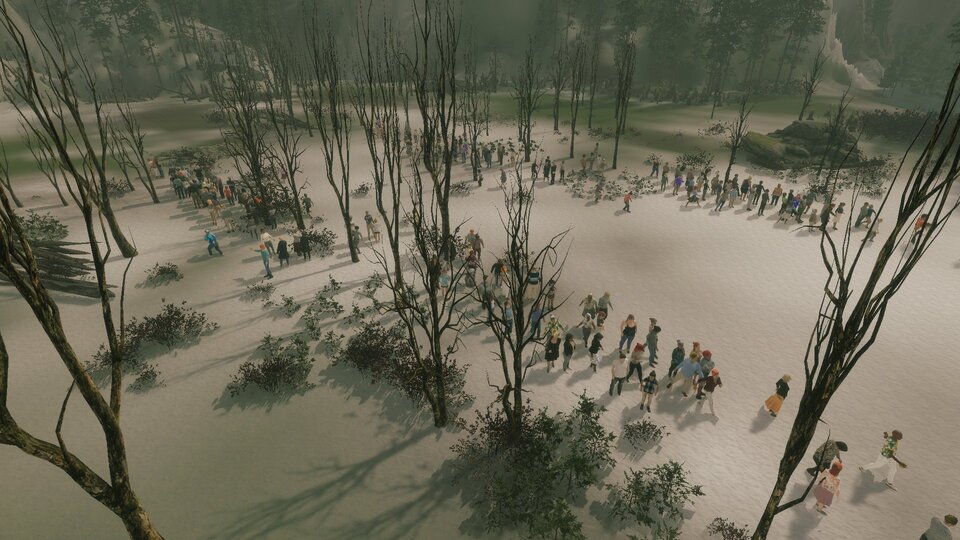} \\
UGV & UAV med-alt & UAV low-alt \vspace{0.1cm}\\
\includegraphics[width=.32\linewidth]{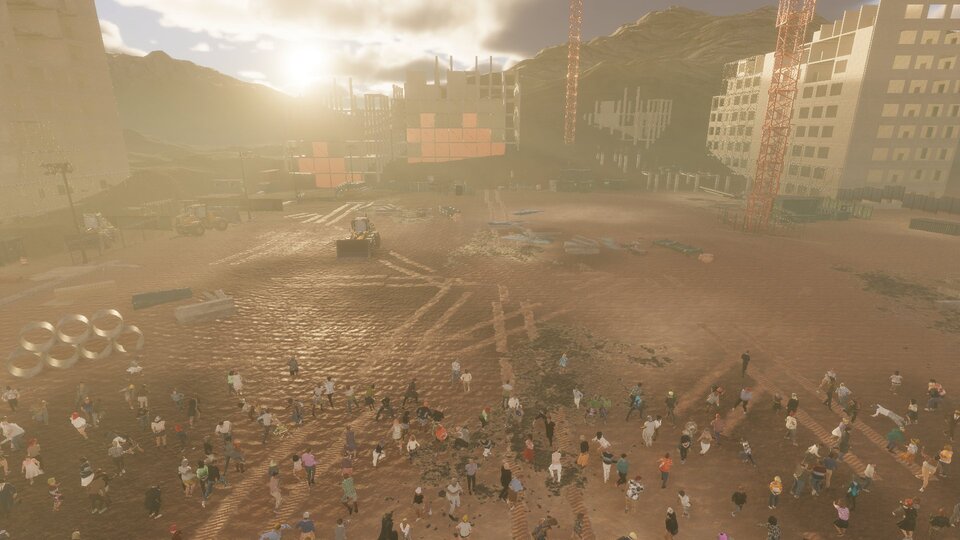} ~&~
\includegraphics[width=.32\linewidth]{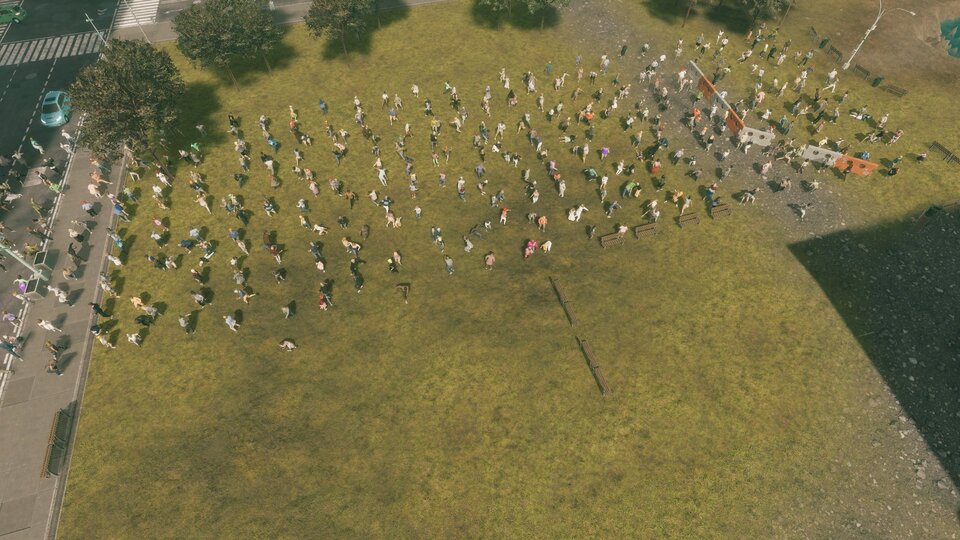} ~&~
\includegraphics[width=.32\linewidth]{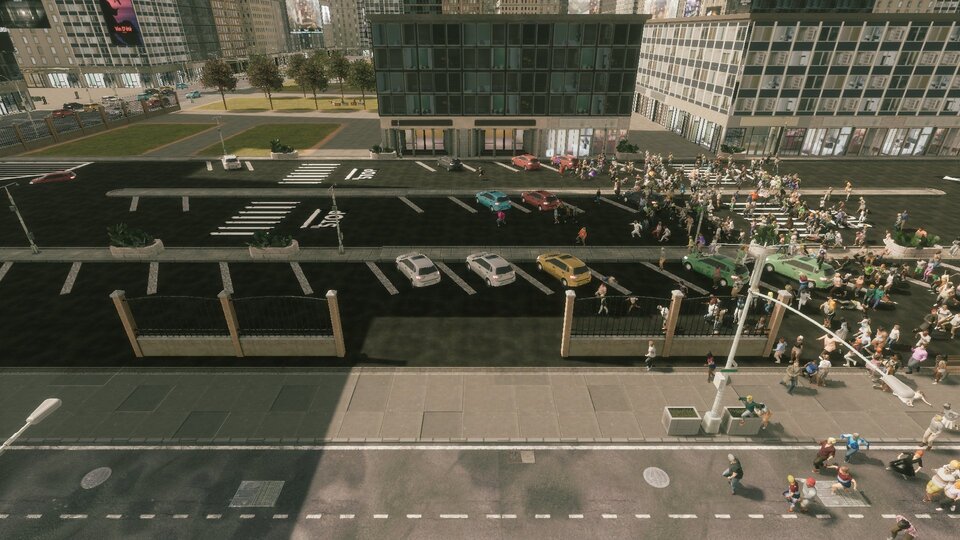} \\
CCTV back & UAV high-alt & CCTV side \\\\
\multicolumn{3}{c}{(2) sugar candy}\vspace{0.2cm}\\
\includegraphics[width=.32\linewidth]{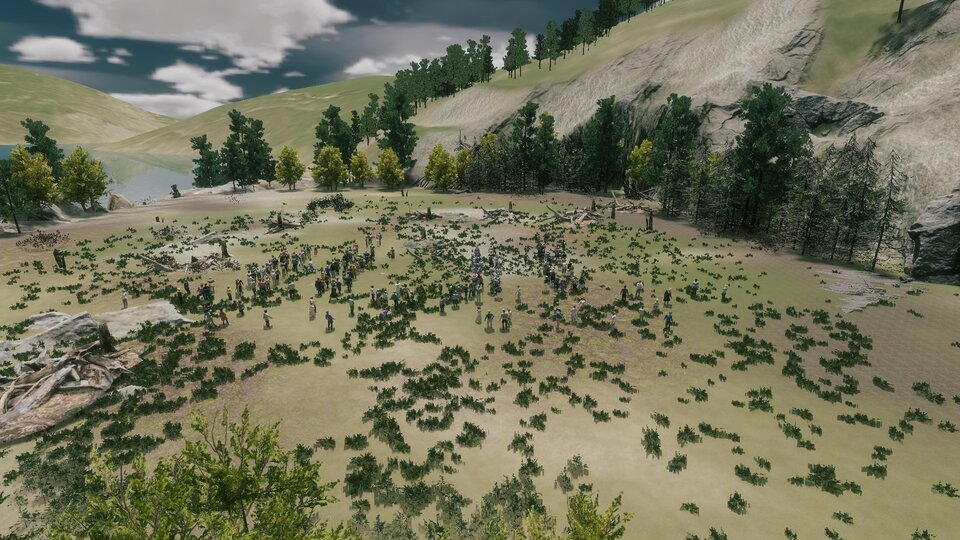} ~&~
\includegraphics[width=.32\linewidth]{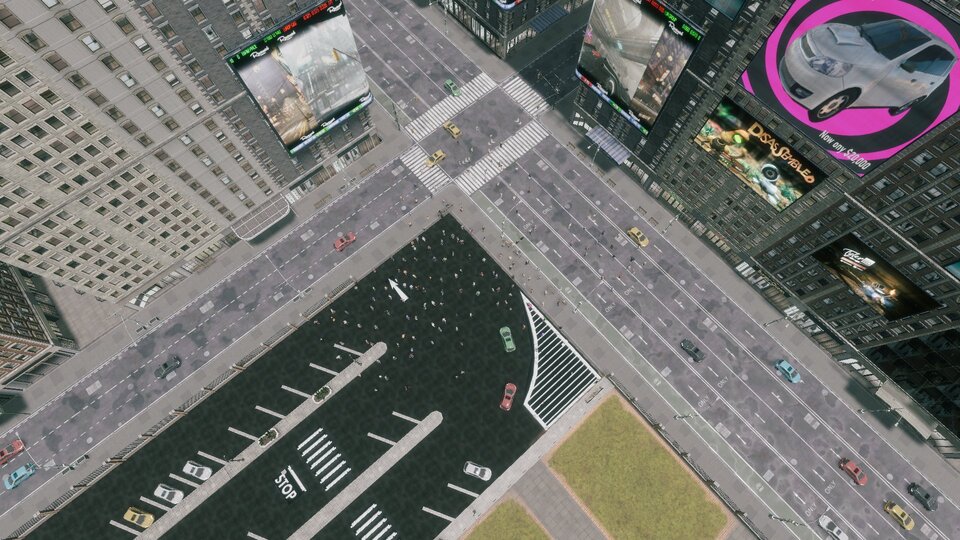} ~&~
\includegraphics[width=.32\linewidth]{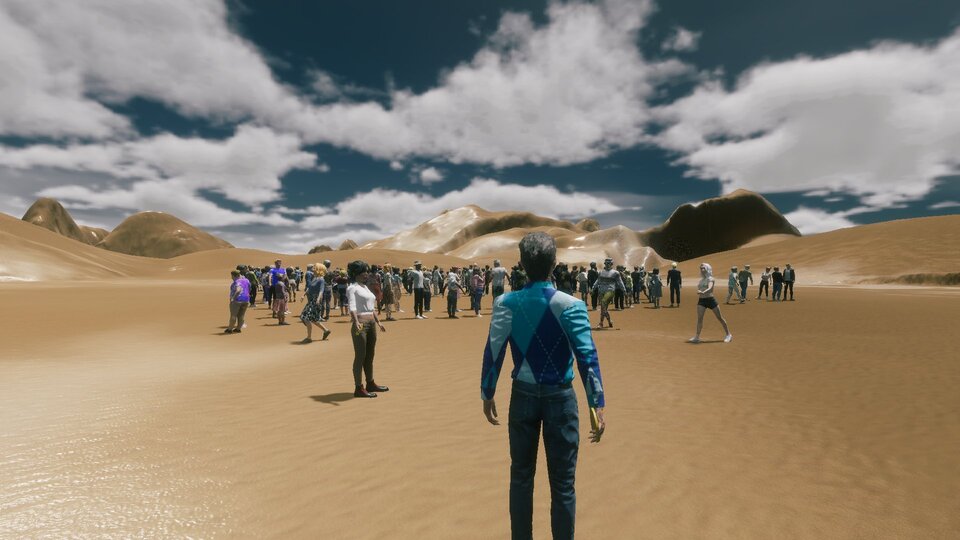} \\
CCTV side & UAV high-alt & UGV \vspace{0.1cm}\\
\includegraphics[width=.32\linewidth]{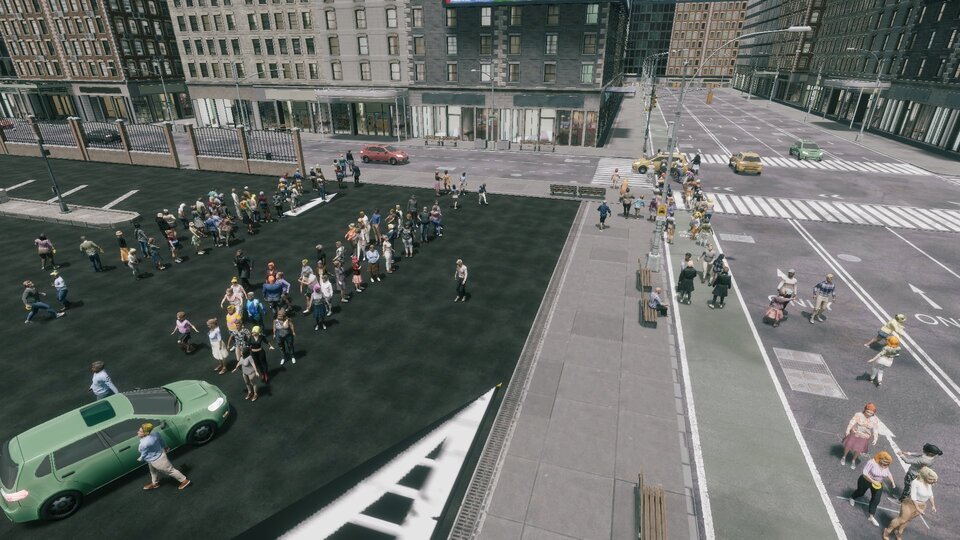} ~&~
\includegraphics[width=.32\linewidth]{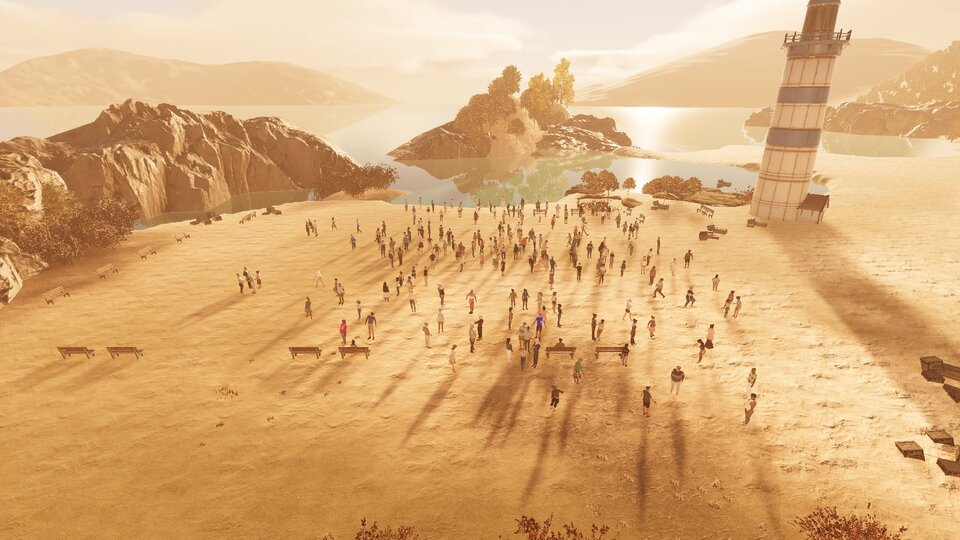} ~&~
\includegraphics[width=.32\linewidth]{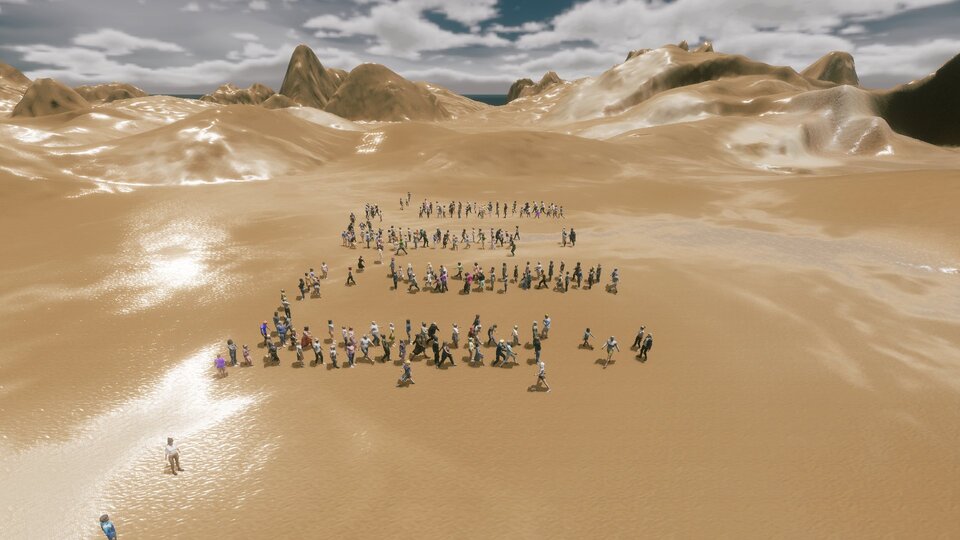} \\
UAV low-alt & CCTV front & CCTV back \\
\end{tabular}
}
\end{figure*}

\begin{figure*}[h]
\centerline{
\setlength{\tabcolsep}{0.5pt}
\begin{tabular}{ccc}
\multicolumn{3}{c}{(3) tug-of-war}\vspace{0.2cm}\\
\includegraphics[width=.32\linewidth]{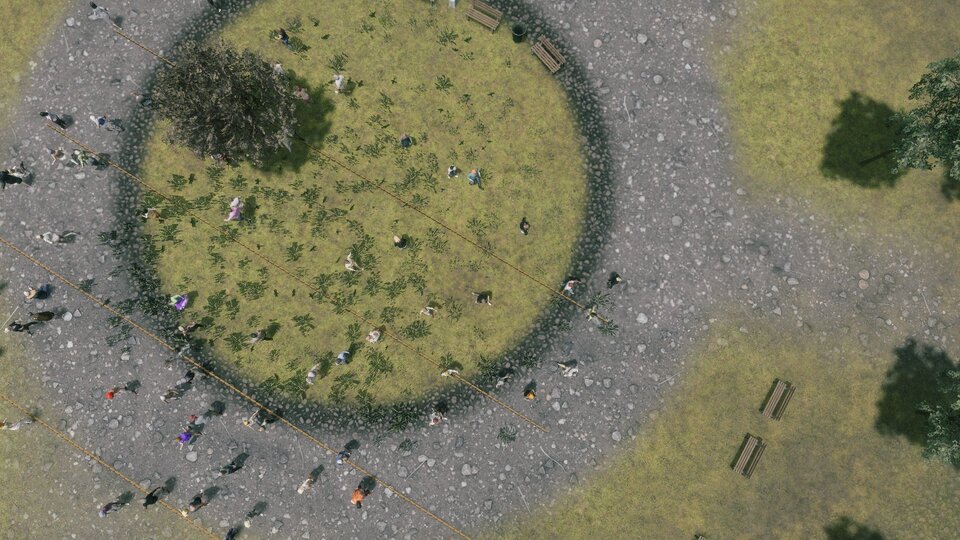} ~&~
\includegraphics[width=.32\linewidth]{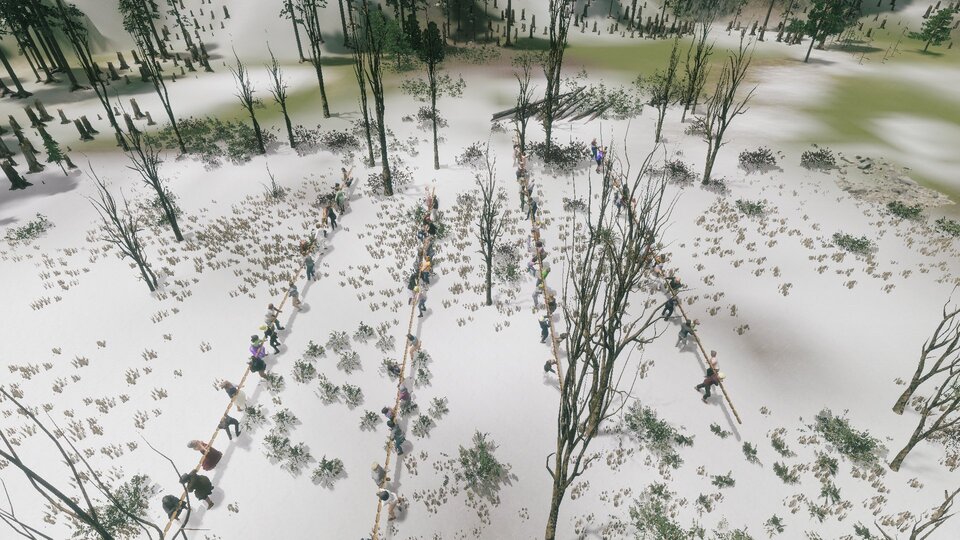} ~&~
\includegraphics[width=.32\linewidth]{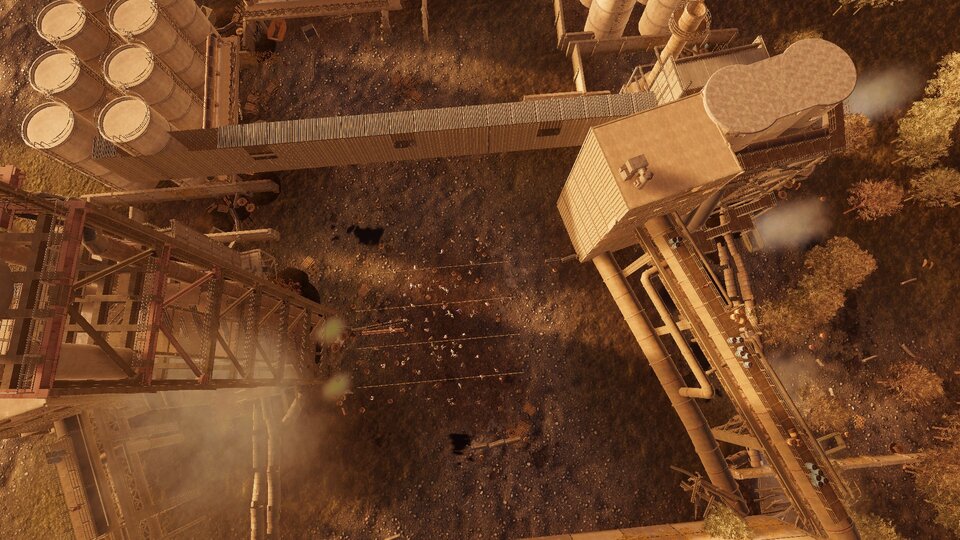} \\
UAV low-alt & CCTV front & UAV high-alt \vspace{0.1cm}\\
\includegraphics[width=.32\linewidth]{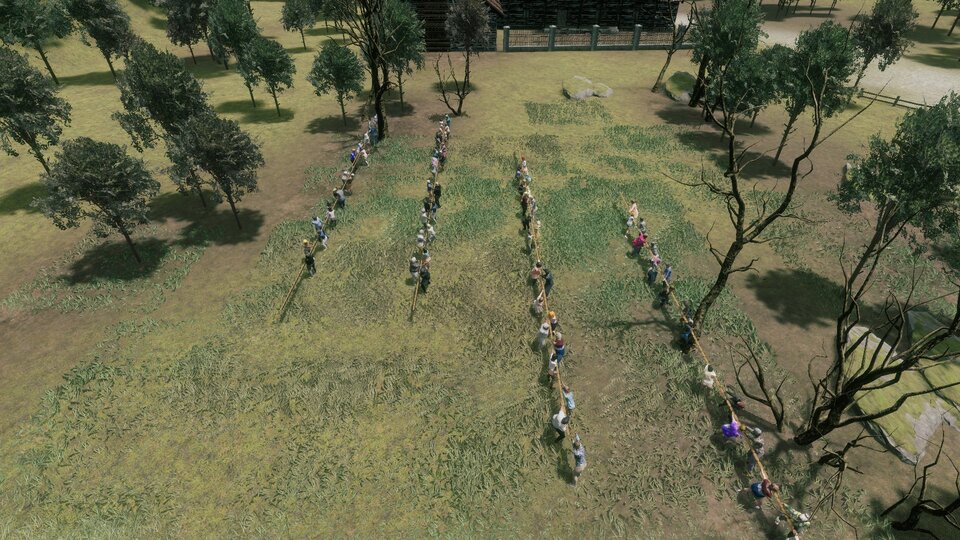} ~&~
\includegraphics[width=.32\linewidth]{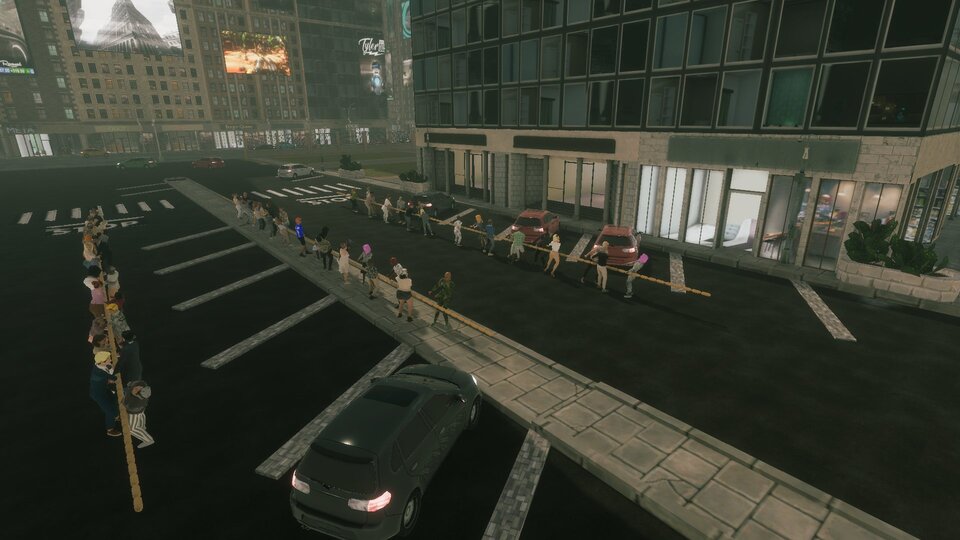} ~&~
\includegraphics[width=.32\linewidth]{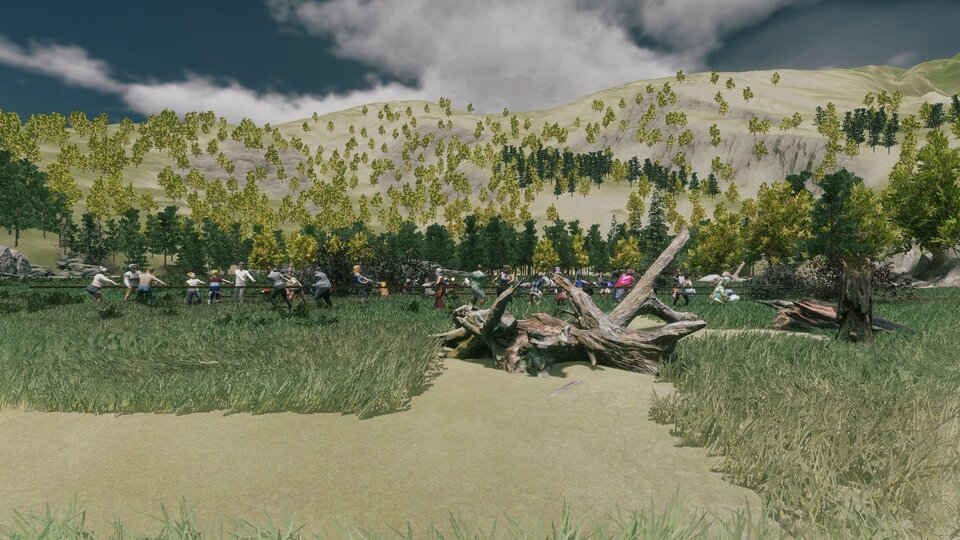} \\
CCTV back & UAV low-alt & UGV \\\\
\multicolumn{3}{c}{(4) marbles}\vspace{0.2cm}\\
\includegraphics[width=.32\linewidth]{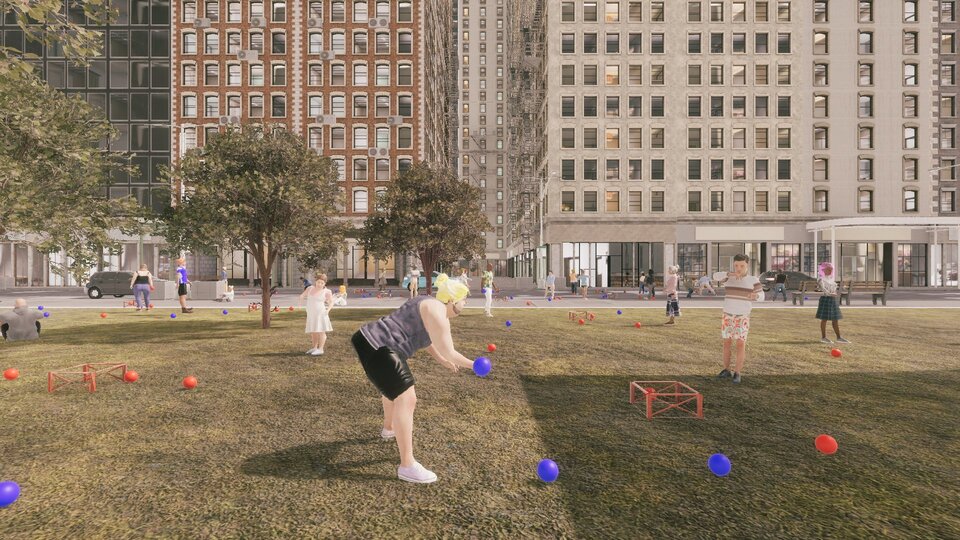} ~&~
\includegraphics[width=.32\linewidth]{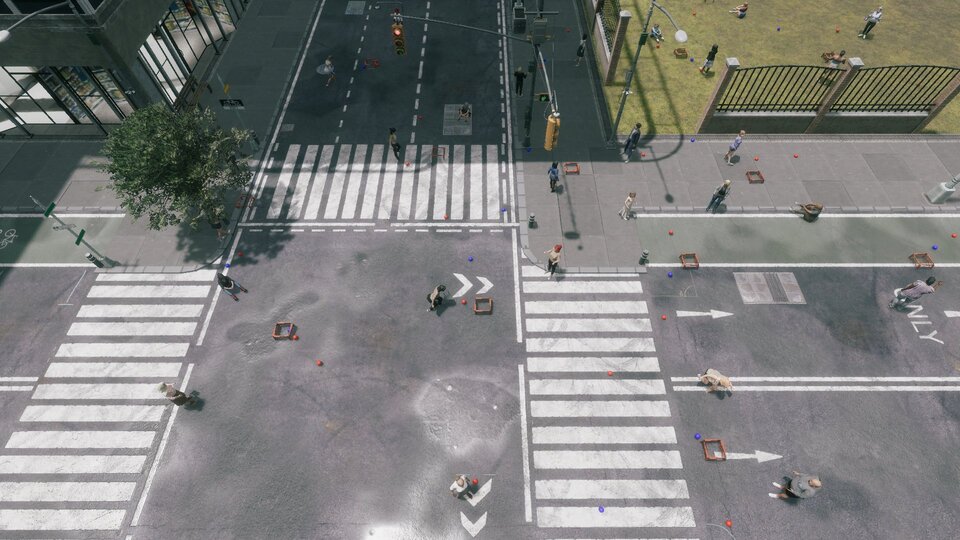} ~&~
\includegraphics[width=.32\linewidth]{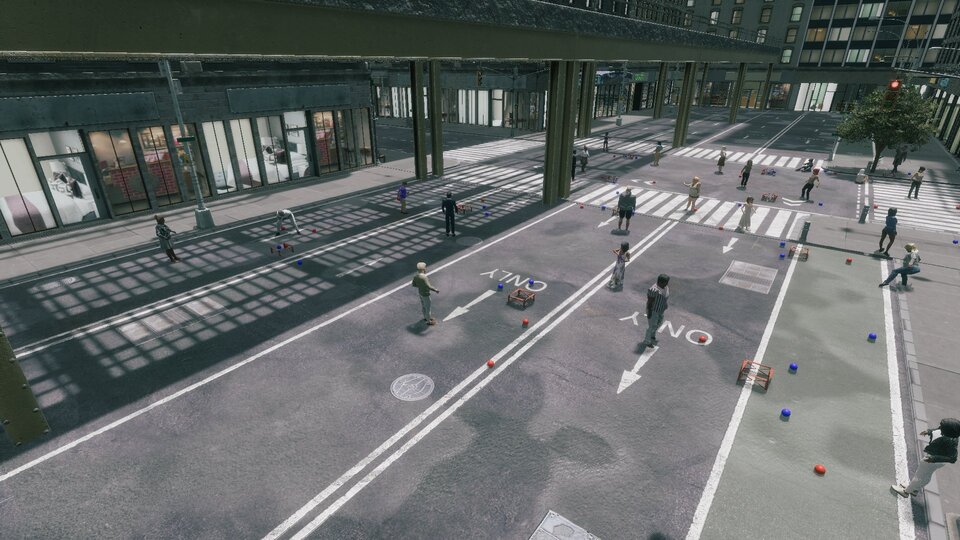} \\
UGV & CCTV front & UAV med-alt \vspace{0.1cm}\\
\includegraphics[width=.32\linewidth]{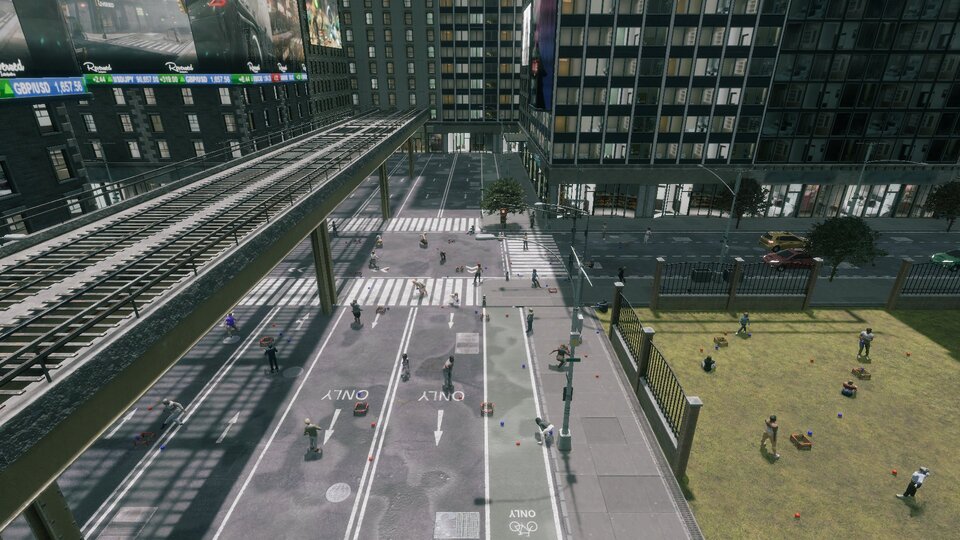} ~&~
\includegraphics[width=.32\linewidth]{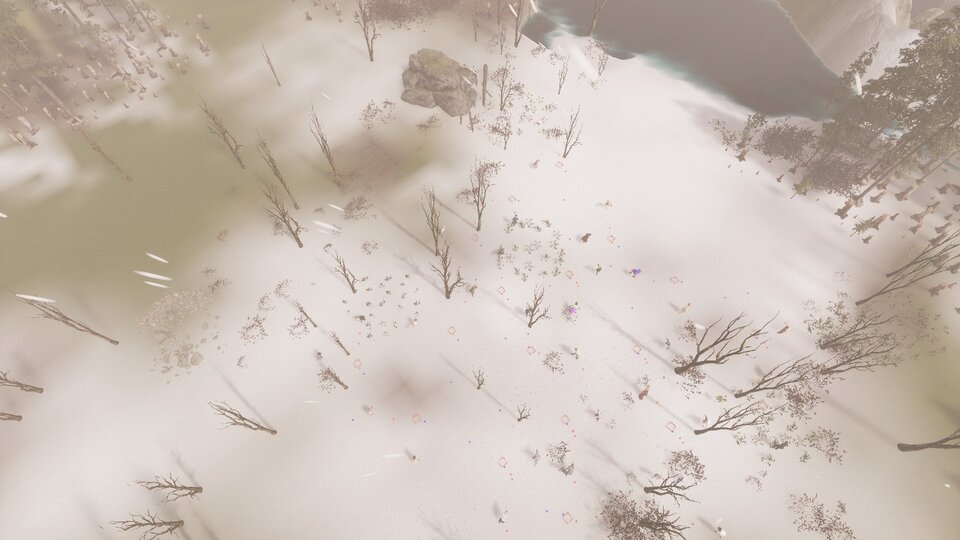} ~&~
\includegraphics[width=.32\linewidth]{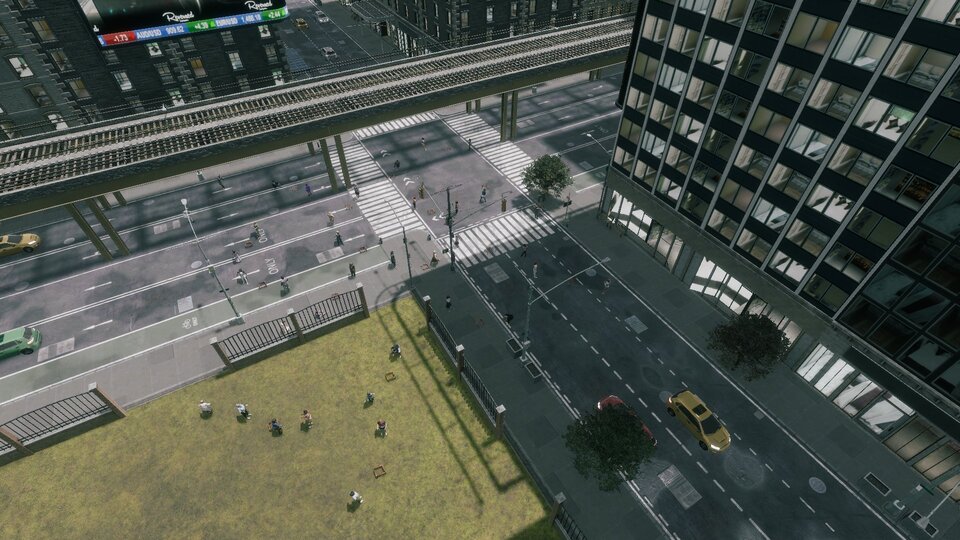} \\
CCTV side & UAV med-alt & UAV low-alt \\
\end{tabular}
}
\end{figure*}

\begin{figure*}[h]
\centerline{
\setlength{\tabcolsep}{0.5pt}
\begin{tabular}{ccc}
\multicolumn{3}{c}{(5) stepping stones}\vspace{0.2cm}\\
\includegraphics[width=.32\linewidth]{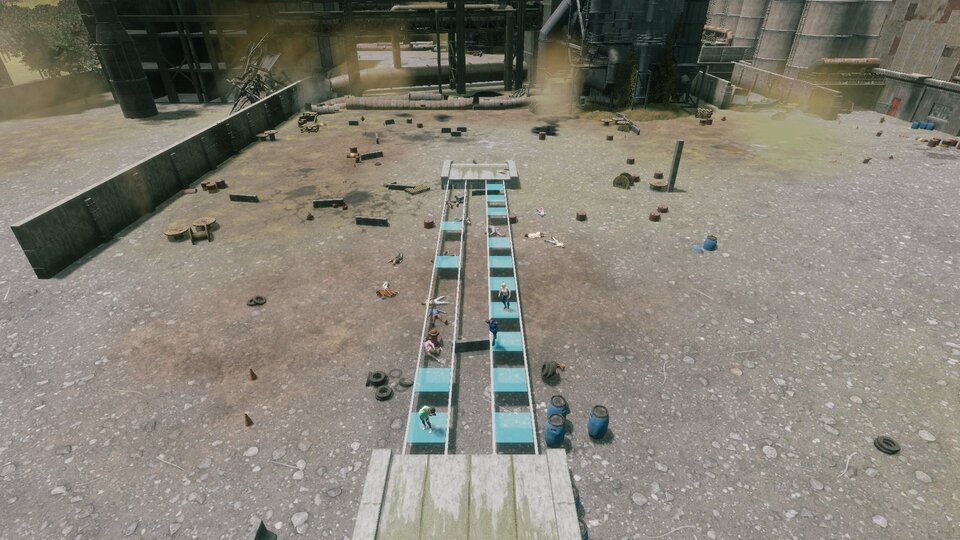} ~&~
\includegraphics[width=.32\linewidth]{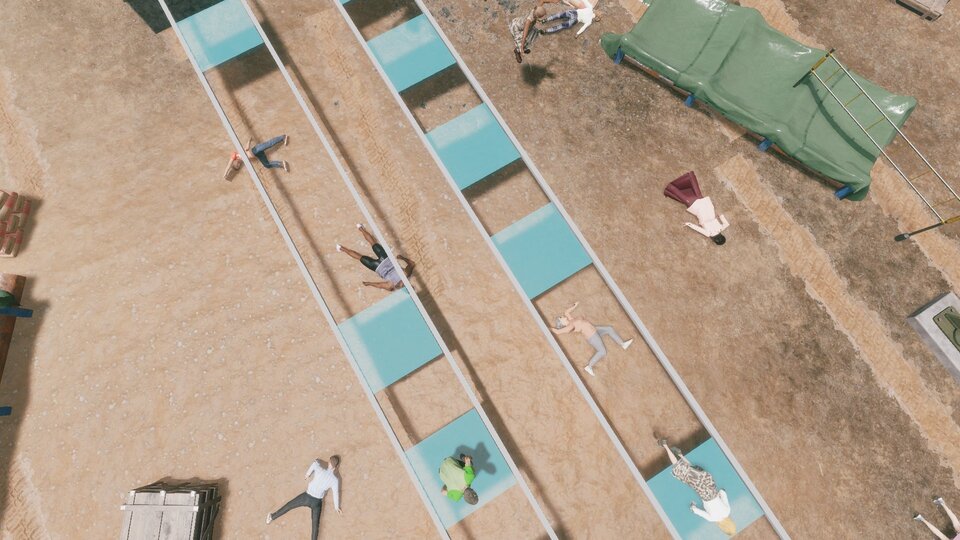} ~&~
\includegraphics[width=.32\linewidth]{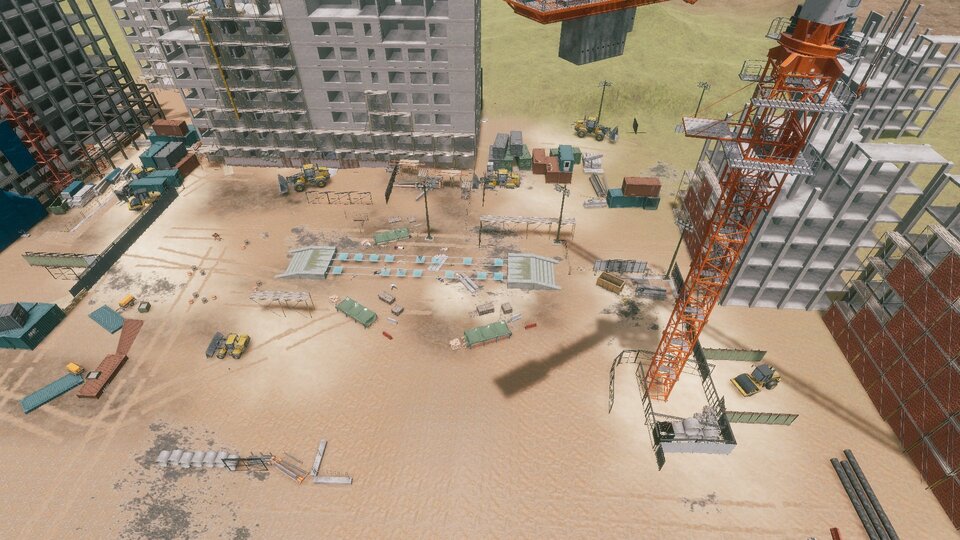} \\
CCTV front & UAV med-alt & UAV high-alt \vspace{0.1cm}\\
\includegraphics[width=.32\linewidth]{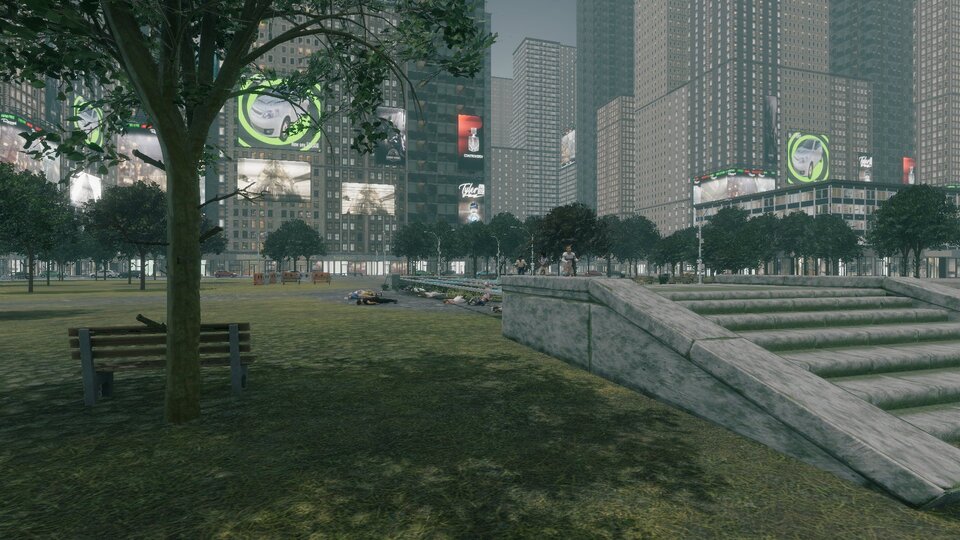} ~&~
\includegraphics[width=.32\linewidth]{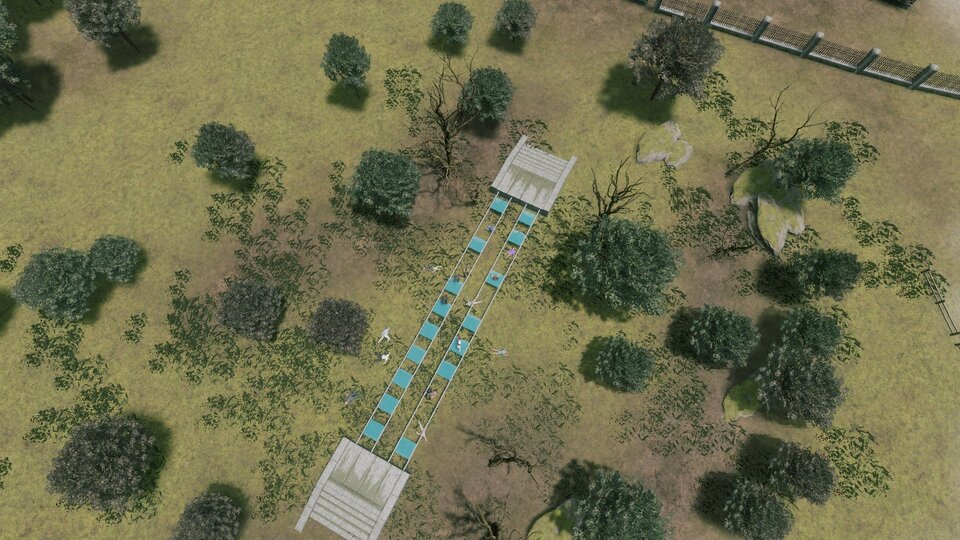} ~&~
\includegraphics[width=.32\linewidth]{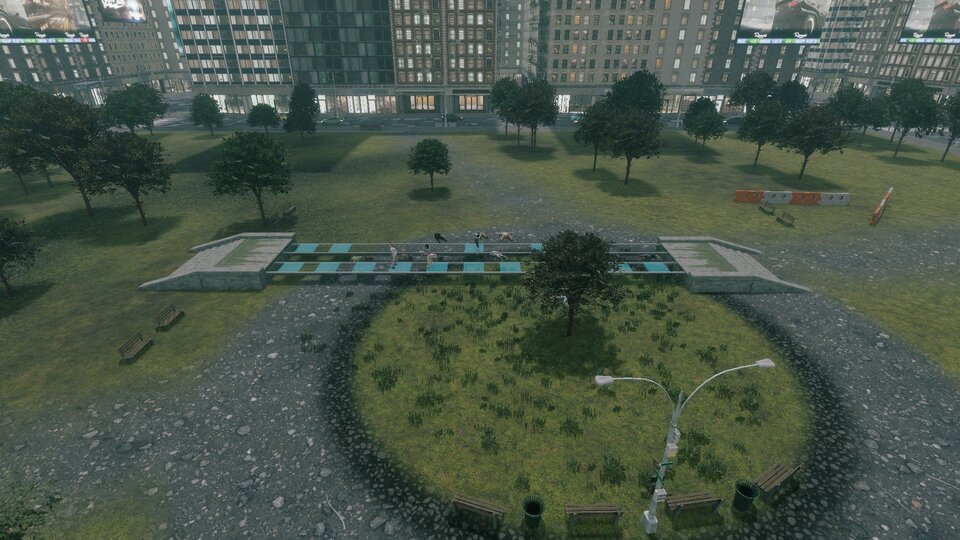} \\
UGV & UAV med-alt & CCTV side \\\\
\multicolumn{3}{c}{(6) squid}\vspace{0.2cm}\\
\includegraphics[width=.32\linewidth]{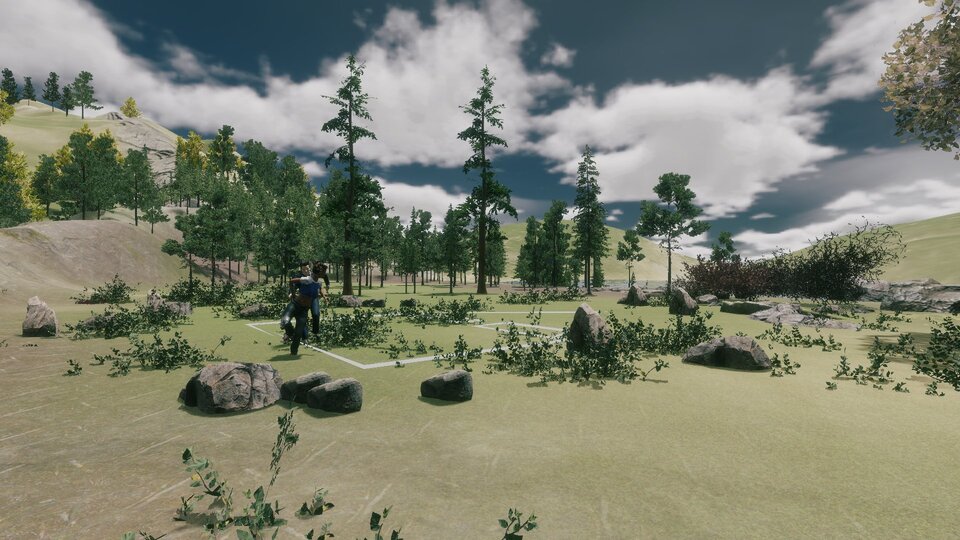} ~&~
\includegraphics[width=.32\linewidth]{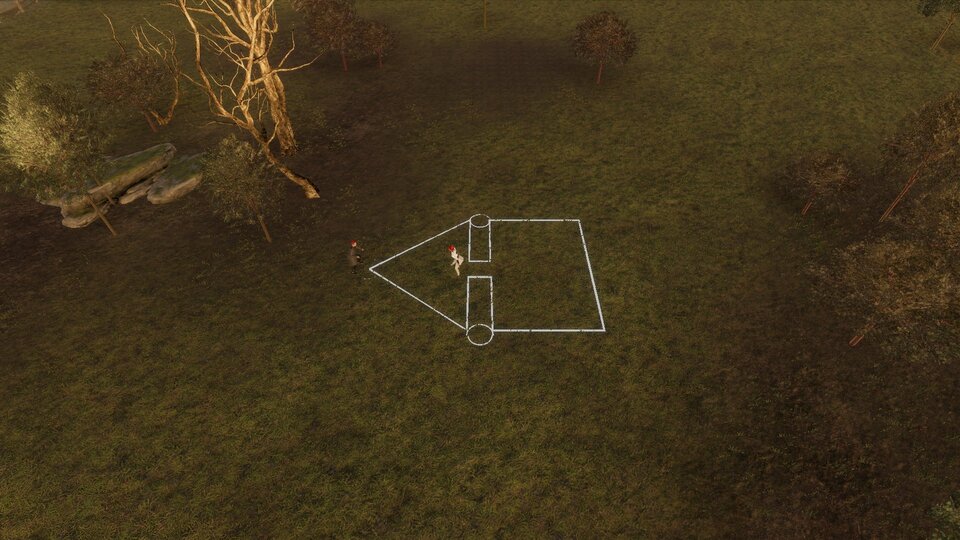} ~&~
\includegraphics[width=.32\linewidth]{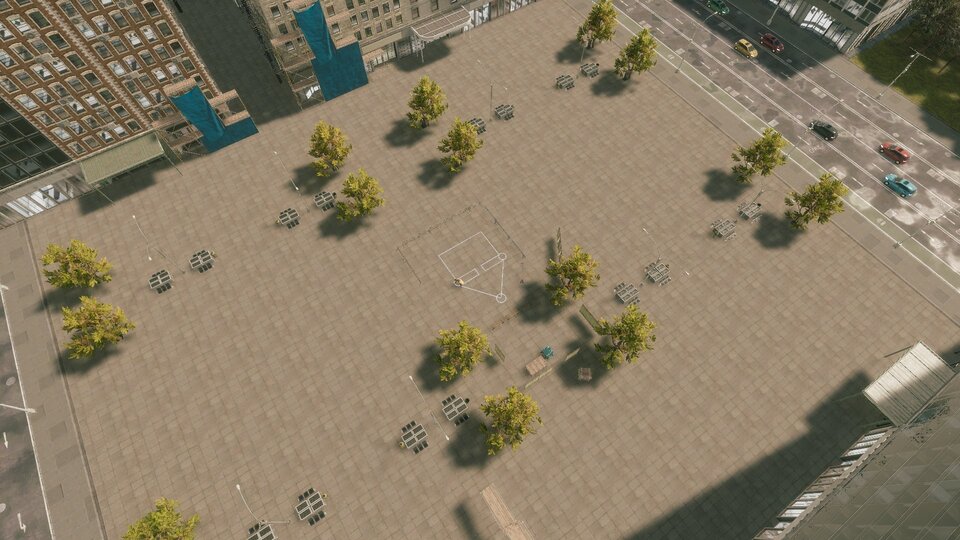} \\
UGV & CCTV front & UAV high-alt \vspace{0.1cm}\\
\includegraphics[width=.32\linewidth]{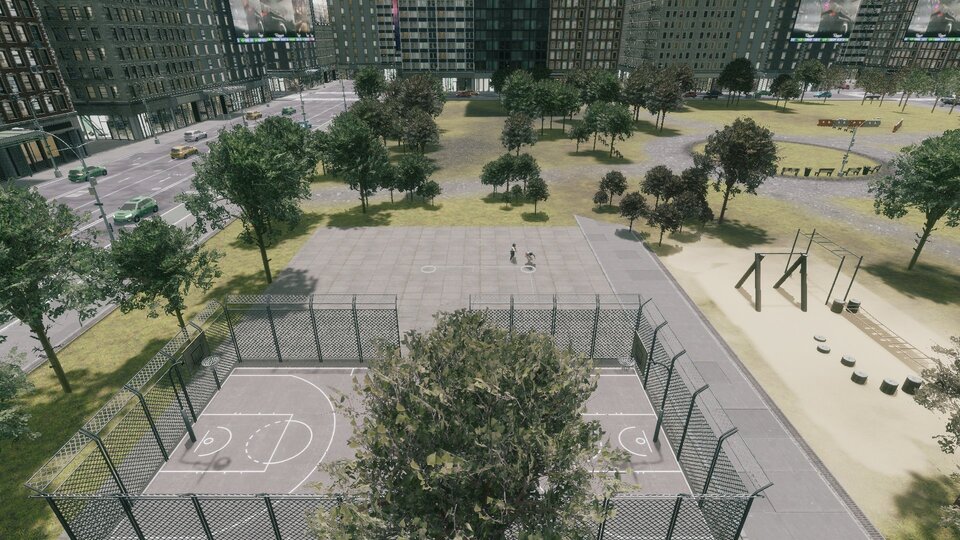} ~&~
\includegraphics[width=.32\linewidth]{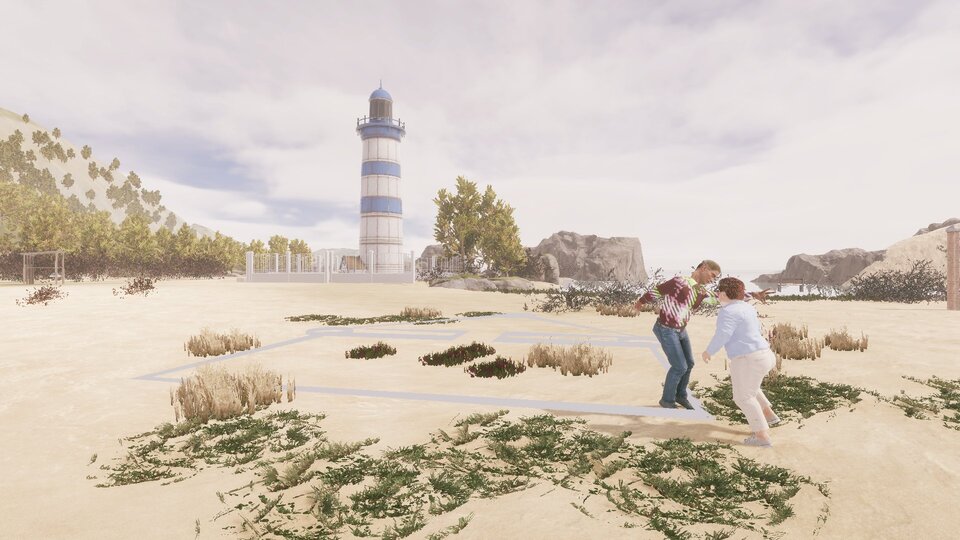} ~&~
\includegraphics[width=.32\linewidth]{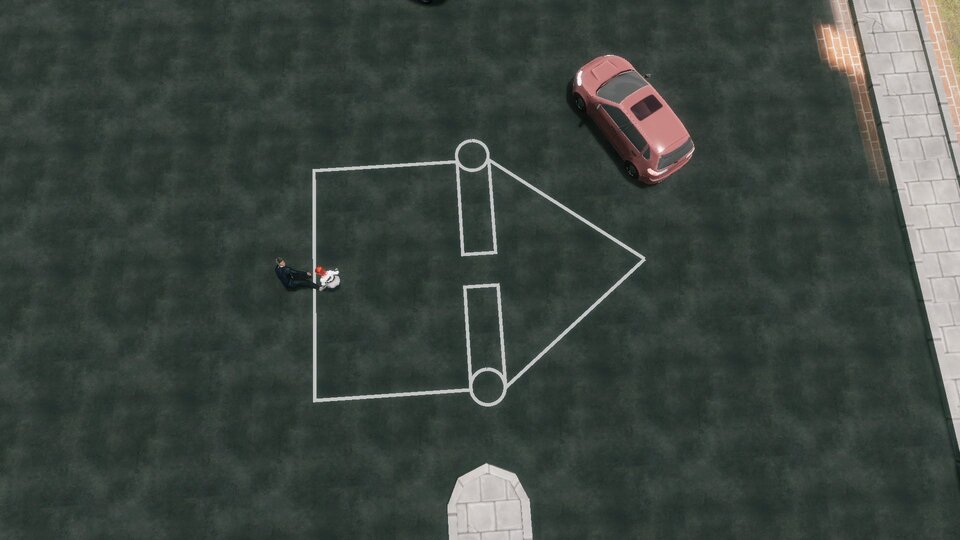} \\
CCTV side & UGV & UAV low-alt \\
\end{tabular}
}

\caption{{\bf More example images from SynPlay} are shown for all six Korean traditional games, each with various camera viewpoints.}
\label{fig:synplay_more_samples}
\end{figure*}

Fig~\ref{fig:synplay_more_samples} includes additional sample images from the SynPlay dataset. Various human appearances depending on human motion differently taken according to the game scenario, and camera viewpoints are observed. Various human appearances are observed that change depending on human motions taken differently according to the game scenario, and different camera viewpoints. In addition, various characters and backgrounds used for creating SynPlay are also visible.
\section{Benchmark Aerial View Dataset}

\begin{figure*}[h]
\centerline{
\setlength{\tabcolsep}{0.5pt}
\begin{tabular}{cccc}
\multicolumn{3}{c}{}\vspace{0.2cm}\\
\includegraphics[width=.32\linewidth]{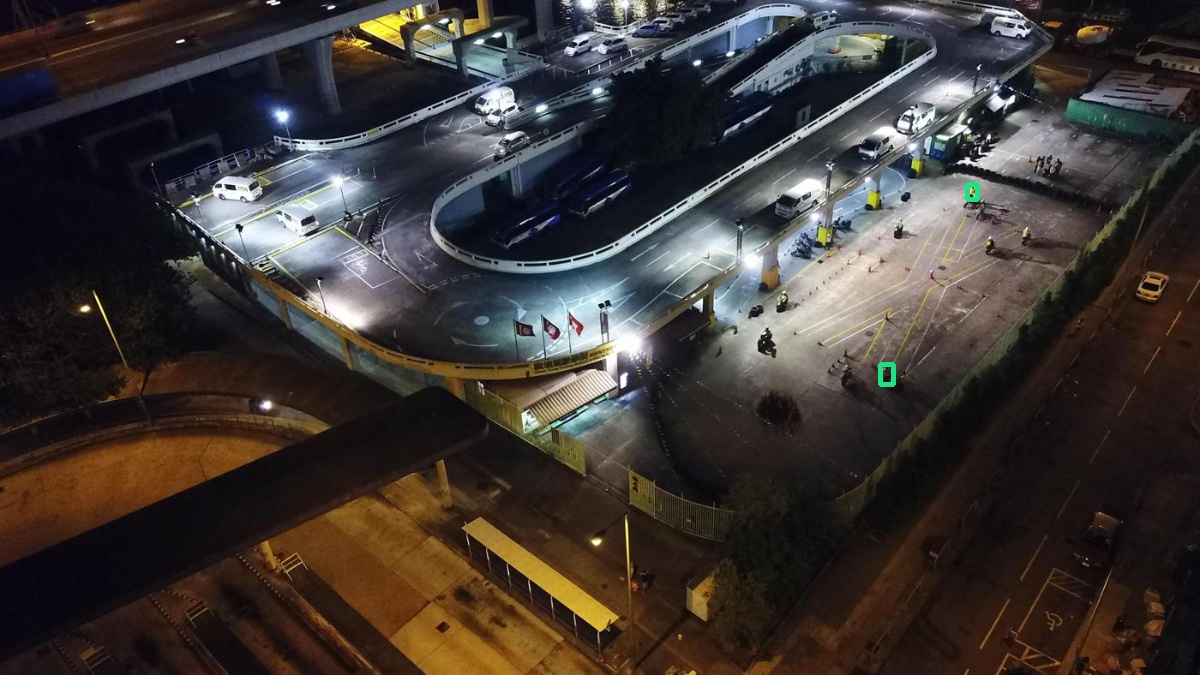} ~&~
\includegraphics[width=.32\linewidth]{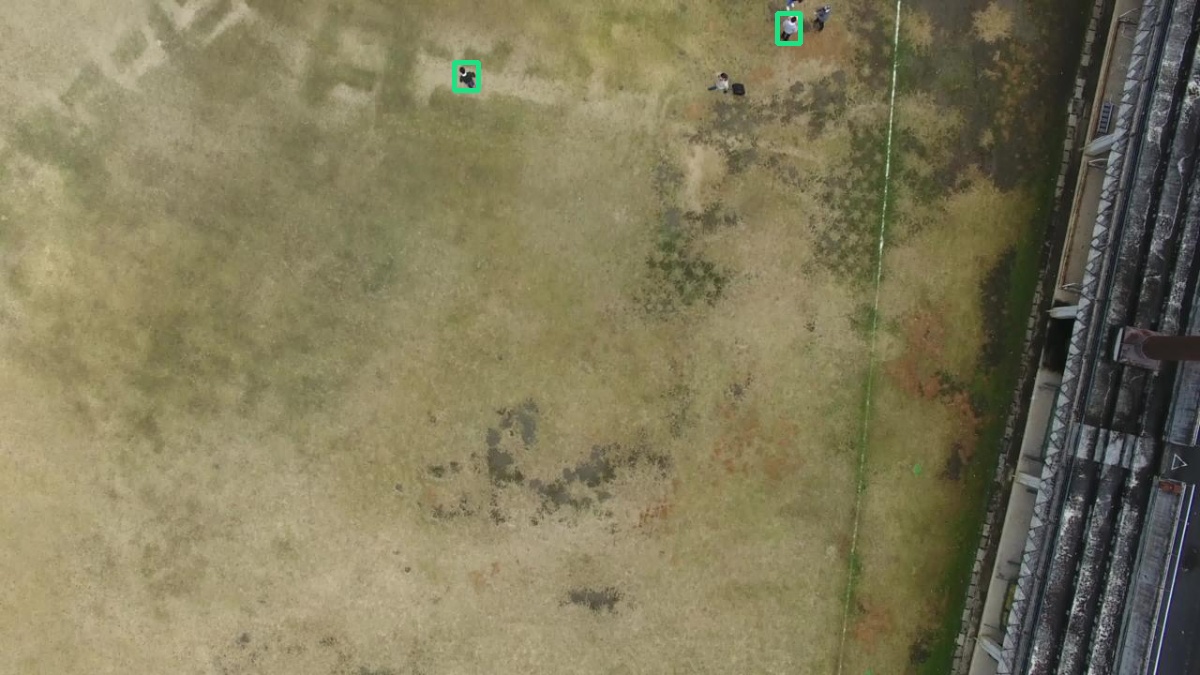} ~&~
\includegraphics[width=.32\linewidth]{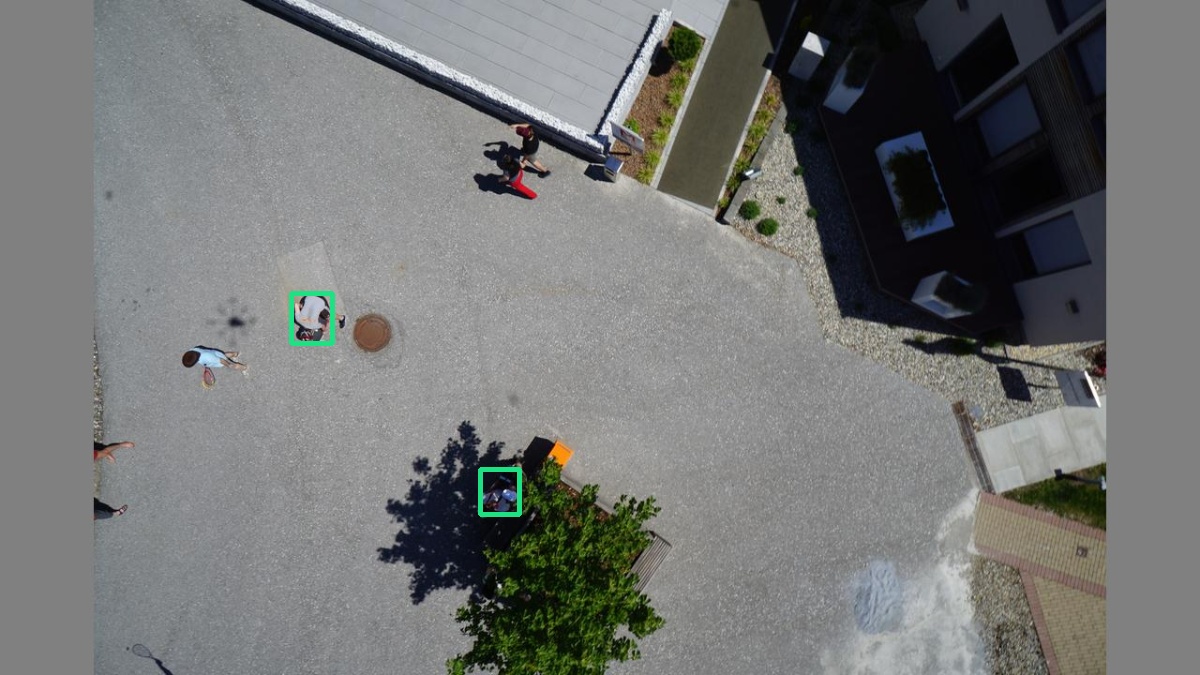} \\
\includegraphics[width=.32\linewidth]{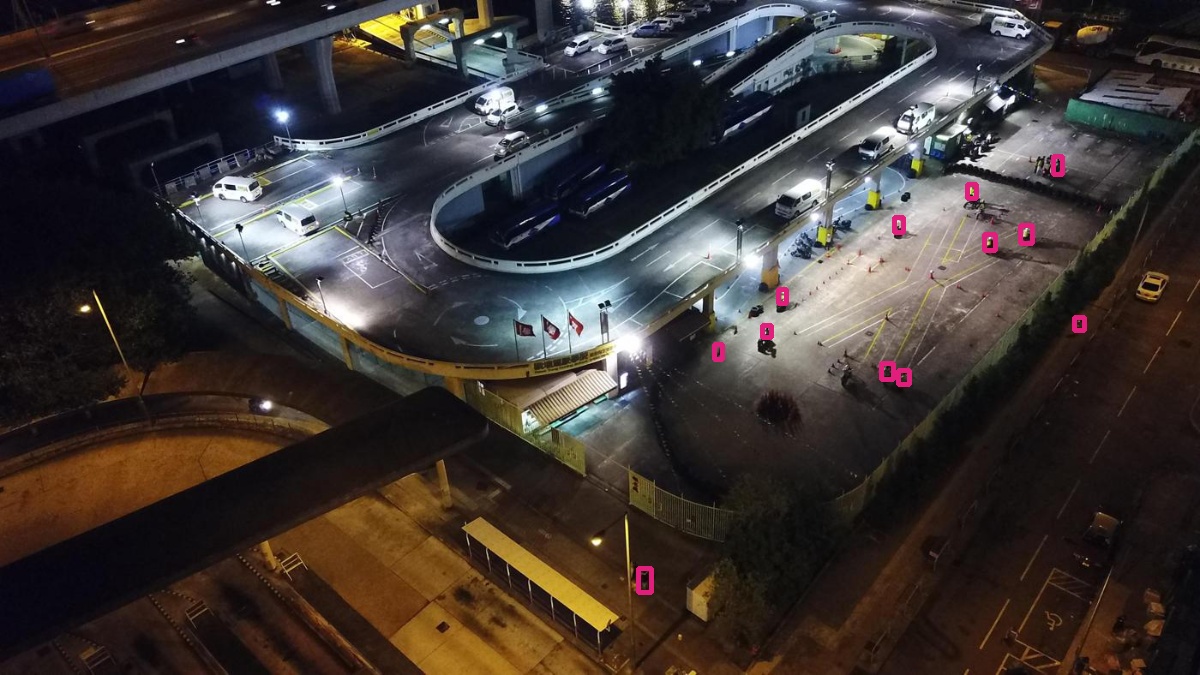} ~&~
\includegraphics[width=.32\linewidth]{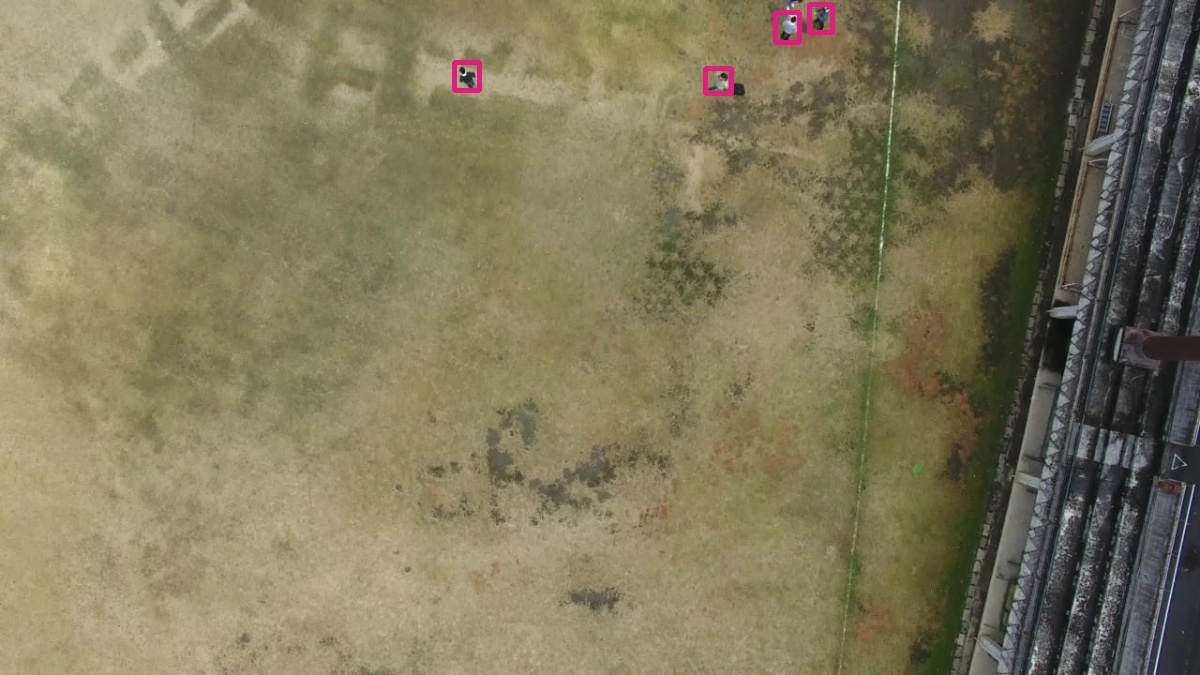} ~&~
\includegraphics[width=.32\linewidth]{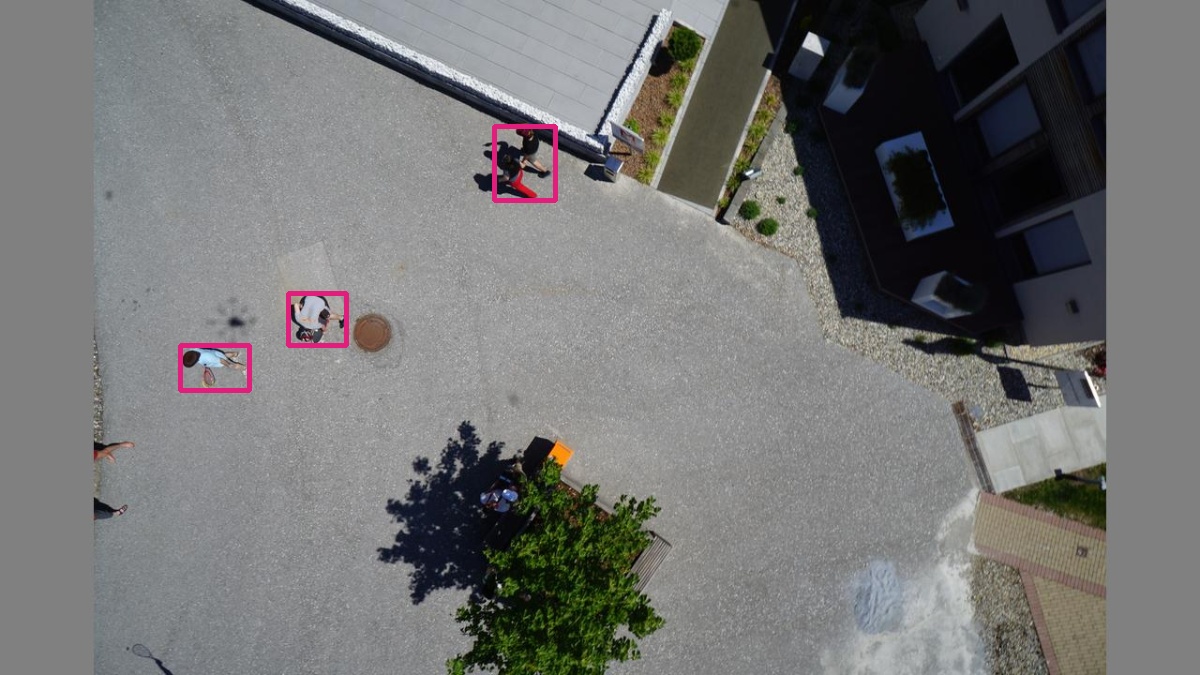} \\

\includegraphics[width=.32\linewidth]{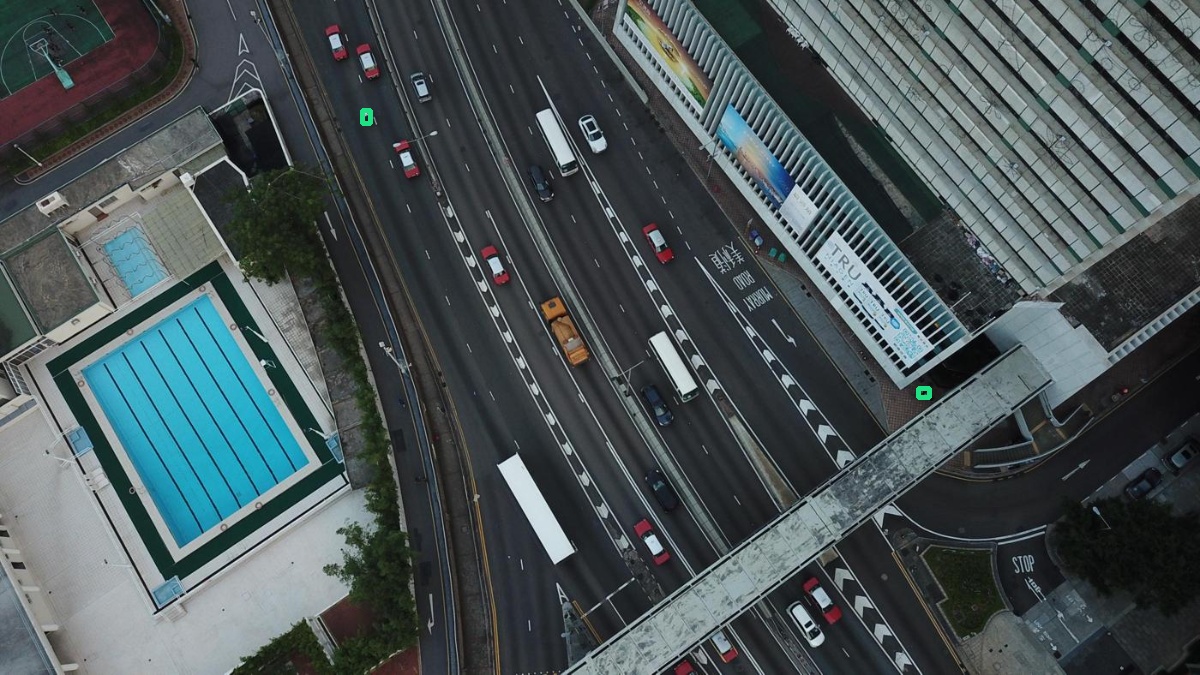} ~&~
\includegraphics[width=.32\linewidth]{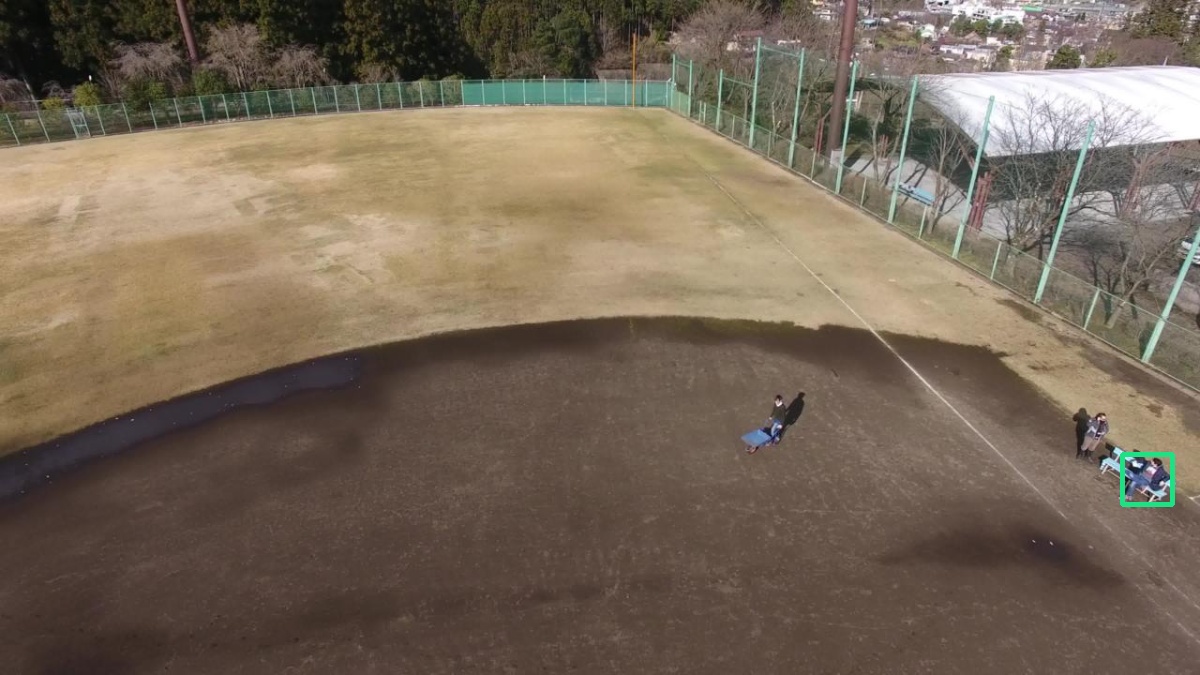} ~&~
\includegraphics[width=.32\linewidth]{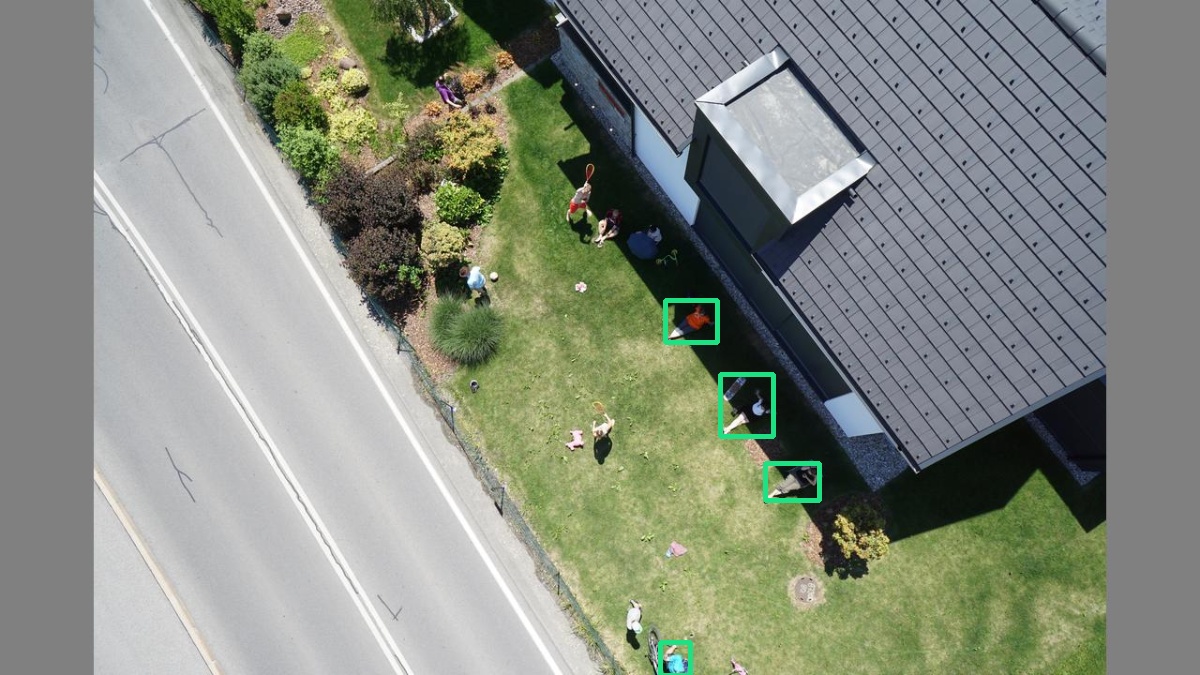} \\
\includegraphics[width=.32\linewidth]{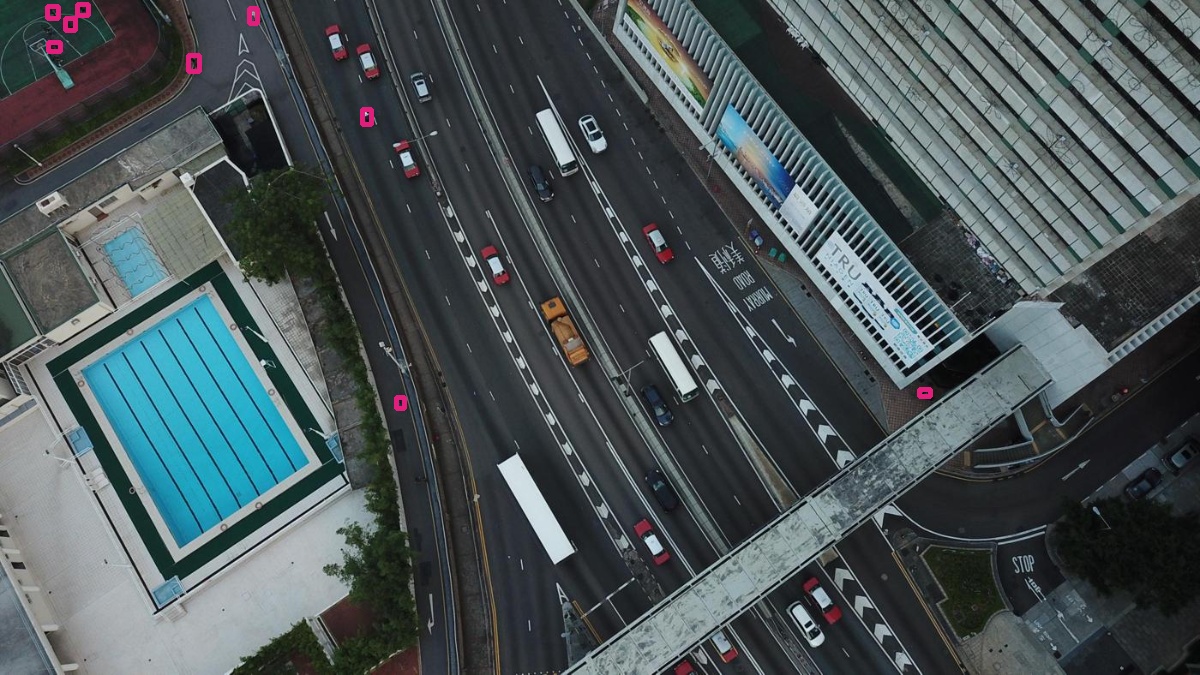} ~&~
\includegraphics[width=.32\linewidth]{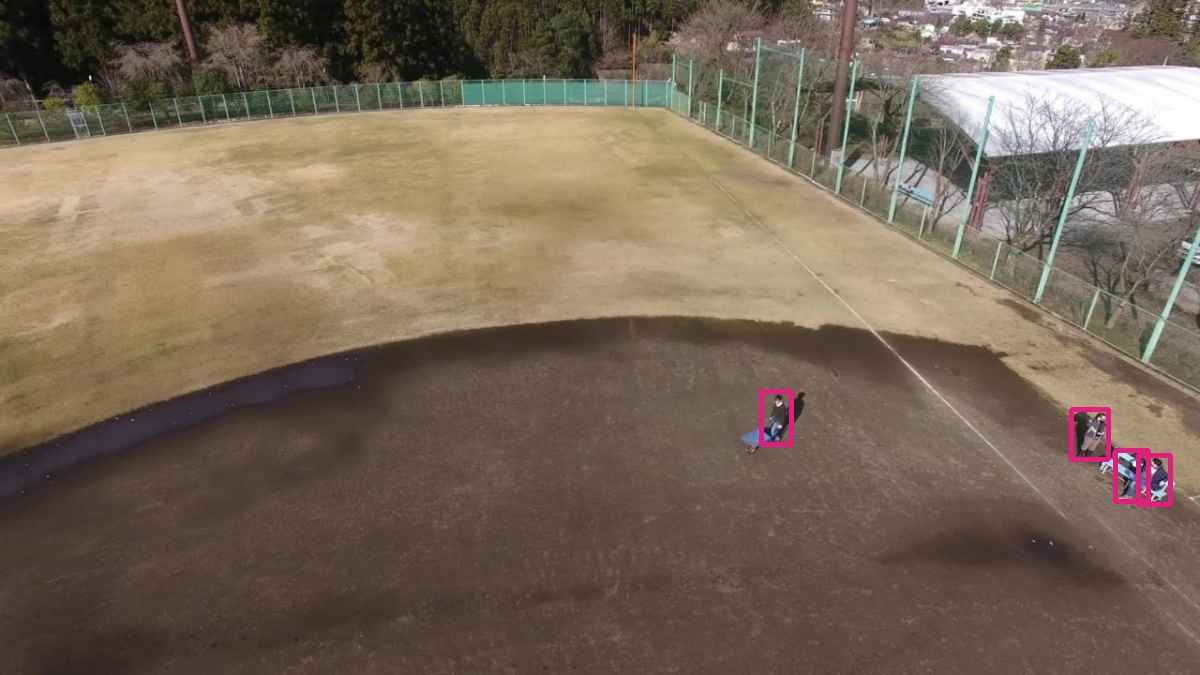} ~&~
\includegraphics[width=.32\linewidth]{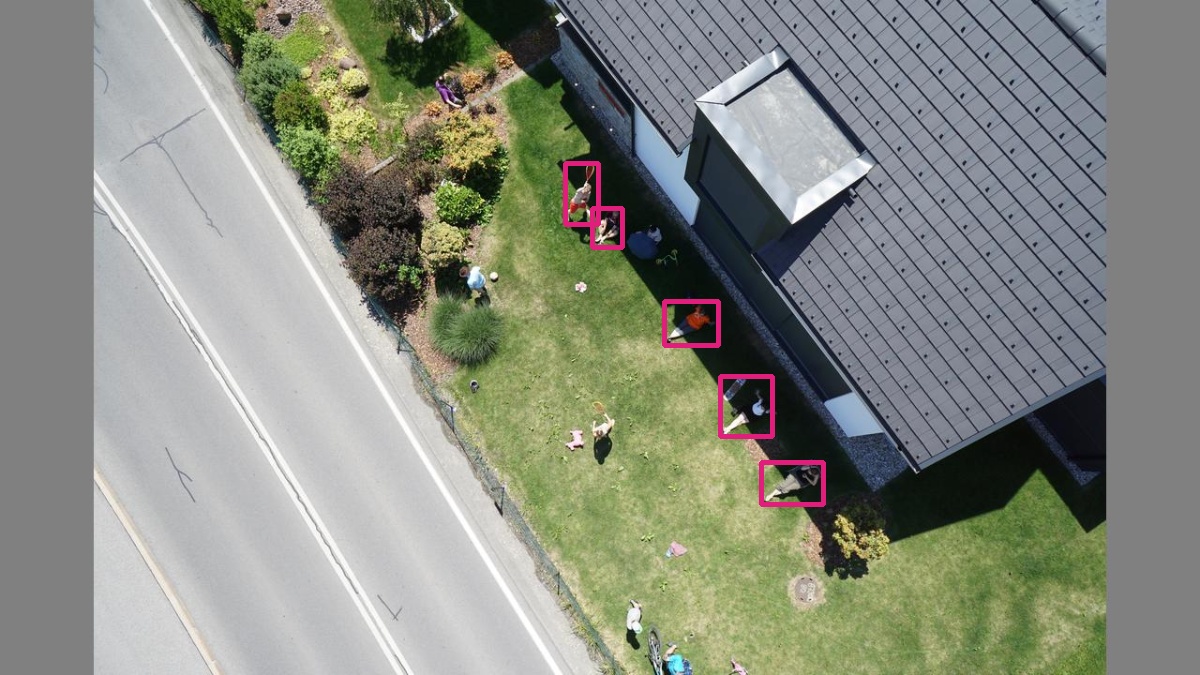} \\

VisDrone~\citep{PZhuTPAMI2022} & Okutama-action~\citep{MBarekatainCVPRW2017} & SemanticDrone~\citep{ICGlink} \\
\end{tabular}
}

\caption{{\bf Qualitative detection samples for baselines vs. SynPlay+real models.} Baseline models are trained on \textbf{real} dataset. \textcolor{teal}{$\square$} (green bounding boxes) indicate baseline (\textbf{real only}) human detection outputs while \textcolor{magenta}{$\square$} (magenta bounding boxes) indicate the ones acquired by the models trained on corresponding \textbf{SynPlay+real} dataset. Note that unlike conventional ground-level benchmarks such as MS COCO~\cite{TLinECCV2014}, aerial-view perception involves \textbf{small-scale human instances}, \textbf{extreme viewpoint variations}, and unique appearance challenges due to camera altitude and perspective. This highlights the need for specialized synthetic datasets like SynPlay that are tailored for long-range aerial human analysis.}
\label{fig:benchmark_qualitative_results}

\end{figure*}

Fig.\ref{fig:benchmark_qualitative_results} presents qualitative detection results acquired on representative aerial-view human datasets, highlighting the fundamental differences between aerial and ground-view perception. Unlike conventional datasets such as MS COCO~\cite{TLinECCV2014}, aerial human imagery involves small-scale human instances, often spanning only tens of pixels, and is captured from extreme and diverse viewpoints, including nadir, oblique, and off-nadir perspectives.

These factors create distinct challenges: while detailed appearance cues like facial features or textures become less informative at such scales, \textit{motion patterns, postures, and interaction dynamics} remain critical. This requires datasets that concurrently capture both motion diversity and viewpoint variation.

SynPlay directly addresses these challenges by combining \textit{multi-perspective captures} with \textit{rule-guided behavioral diversity}, enabling robust human identification in aerial scenarios where existing datasets fall short as shown in the qualitative comparisons.

\section{Limitations and Future Directions}

A significant number of human instances in SynPlay appear at very low resolutions, which is an inherent challenge in any aerial-view dataset. However, as shown in the bounding box size histogram included in the supplementary material, SynPlay also contains many high-resolution human instances, with approximately 10,000 examples having bounding box areas greater than 10,000 pixels. We encourage selectively using these instances depending on the requirements of specific tasks.

SynPlay was developed to provide rich visual representations of human appearance for tasks focused on localizing people in complex scenes. While it centers on the human domain, we recognize the value of incorporating features from a broader set of object categories. Expanding SynPlay to include a wider array of objects could further enhance its utility for training and evaluating general-purpose models.

Future directions for the community include improving photorealism, simulating sensor artifacts such as rolling shutter and motion blur, and integrating synthetic and real data through hybrid training protocols. Adjusting scenario priors to support more diverse interactions and cultural settings, expanding metadata for occlusion and weather conditions, and incorporating additional sensing modalities like thermal imagery may also support greater robustness. More detailed behavioral annotations could further enable research in tracking, forecasting, and social interaction understanding. These extensions build on SynPlay’s strengths in controllability, scale, and behavioral diversity while helping shape the next generation of aerial perception benchmarks.\smallskip



{
    \small
    \bibliographystyle{ieeenat_fullname}
    \bibliography{main}
}

\end{document}